\documentclass[b5paper,10pt]{book}

\usepackage{phdthesis}
\usepackage[
  paperwidth=17cm
  ,paperheight=24cm
  ,inner=2.6cm
  ,outer=2.3cm
  ,top=2.3cm
  ,bottom=2.3cm
]{geometry}

\usepackage{graphicx,times}
\usepackage[hyphens]{url}
\usepackage{tabularx,amsmath}
\usepackage{multirow}
\usepackage{booktabs}
\usepackage{amssymb}
\usepackage[comma,compress,sectionbib]{natbib}
\usepackage{wrapfig}
\usepackage{enumerate}
\usepackage{supertabular}
\usepackage{multicol}
\usepackage{setspace}
\usepackage{import}
\usepackage{booktabs} 
\usepackage{graphicx}
\usepackage{url}
\usepackage{times}
\usepackage{microtype}
\usepackage{amssymb}
\usepackage{multirow}
\usepackage{color}
\usepackage[usenames,dvipsnames,table]{xcolor}
\usepackage[normalem]{ulem}
\usepackage[english]{babel}
\usepackage{statex}
\usepackage[noend]{algorithmic}
\usepackage{algorithm}
\usepackage{paralist}
\usepackage{flushend}
\usepackage{enumerate}
\usepackage{enumitem}
\usepackage{setspace}
\usepackage{textcomp}
\usepackage{listings}
\usepackage{keyval}
\usepackage{rotating}
\usepackage{acronym}

\usepackage[utf8]{inputenc}
\usepackage{CJKutf8}
\usepackage{subcaption}
\usepackage{mathabx}
\usepackage{skak}
\usepackage{makecell}
\usepackage{tablefootnote}
\usepackage{mathtools}
\usepackage{float}
\usepackage{rotating}
\usepackage{amsthm}

\theoremstyle{definition}
\newtheorem{definition}{Definition}[section]

\DeclarePairedDelimiter\floor{\lfloor}{\rfloor}
\renewcommand{\arraystretch}{1.3} 
\newcommand{\argmax}{\operatornamewithlimits{arg\,max}}

\DeclareMathOperator{\length}{\operatorname{length}}
\DeclareMathOperator{\softmax}{\operatorname{softmax}}
\DeclareMathOperator{\concat}{\operatorname{concat}}
\newcommand{\zh}[1]{\begin{CJK}{UTF8}{gbsn}#1\end{CJK}}

\usepackage[backref=page]{hyperref}
\hypersetup{%
pdfborder = {0 0 0},
pdftitle = {An Empirical Analysis of Phrase-based and Neural Machine Translation},
pdfsubject = {PhD Thesis},
pdfkeywords = {Machine Translation, Phrase-Based Machine Translation, Neural Machine Translation, Attention Models, Reordering Models, Hidden State Analysis},
pdfauthor = {Hamidreza Ghader}
}
\usepackage{bookmark}

\renewcommand*{\backref}[1]{} 
\renewcommand*{\backrefalt}[4]{%
\ifcase #1
\or (Cited on page~#2.)  %
\else 
(Cited on pages~#2.)  
\fi
}

\let\svthefootnote\thefootnote
\newcommand\blankfootnote[1]{%
  \let\thefootnote\relax\footnotetext{#1}%
  \let\thefootnote\svthefootnote%
}
\let\svfootnote\footnote
\renewcommand\footnote[2][?]{%
  \if\relax#1\relax%
    \blankfootnote{#2}%
  \else%
    \if?#1\svfootnote{#2}\else\svfootnote[#1]{#2}\fi%
  \fi
}

\usepackage{threeparttable}

\begin{document}

\frontmatter

{\pagestyle{empty}
\newcommand{\printtitle}{%
{\Huge\bf An Empirical Analysis of Phrase-based and Neural Machine Translation \\[0.8cm]
}}

\begin{titlepage}
\par\vskip 2cm
\begin{center}
\printtitle
\vfill
{\LARGE\bf Hamidreza Ghader}
\vskip 2cm
\end{center}
\end{titlepage}

\mbox{}\newpage
\setcounter{page}{1}

\clearpage
\par\vskip 2cm
\begin{center}
\printtitle
\par\vspace {4cm}
{\large \sc Academisch Proefschrift}
\par\vspace {1cm}
{\large ter verkrijging van de graad van doctor aan de \\
Universiteit van Amsterdam\\
op gezag van de Rector Magnificus\\
prof.\ dr.\ ir.\ K.I.J. Maex\\
ten overstaan van een door het College voor Promoties ingestelde \\
commissie, in het openbaar te verdedigen in \\
de Agnietenkapel\\
op donderdag 29 oktober 2020, te 16:00 uur \\ } 
\par\vspace {1cm} {\large door}
\par \vspace {1cm}
{\Large Hamidreza Ghader}
\par\vspace {1cm}
{\large geboren te Karaj} 
\end{center}

\clearpage
\noindent%
\textbf{Promotiecommissie} \\\\
\begin{tabular}{@{}l l l}
Promotor: \\
& Dr.\ C.\ Monz & Universiteit van Amsterdam \\  
Co-promotor: \\
& Prof.\ dr.\ M.\ de Rijke & Universiteit van Amsterdam \\  
Overige leden: \\
& Dr. G. Chrupala & Tilburg University \\
& Prof. dr. J. van Genabith & German Center for AI (DFKI) \\
& Prof. dr. T. Gevers & Universiteit van Amsterdam \\
& Prof. dr. P. Groth & Universiteit van Amsterdam \\
& Dr. E. Shutova & Universiteit van Amsterdam \\
\end{tabular}

\bigskip\noindent%
Faculteit der Natuurwetenschappen, Wiskunde en Informatica\\

\vfill

\noindent
The research was supported by the Netherlands Organization for Scientific Research (NWO) under project numbers  639.022.213 and 612.001.218\\
\bigskip

\noindent
Copyright \copyright~2020 Hamidreza Ghader, Amsterdam, The Netherlands\\
Cover by Maryam Kamranfar. Photo by Peter Hermes Furian.\\
Printed by Ipskamp Printing\\
ISBN: 978-94-6421-039-2\\

\clearpage
}

{\pagestyle{empty}

\mbox{}
\vspace{2in}
\begin{center}
{\em To my parents \\ whose dedication was the most inspiring}
\end{center}

\clearpage
}

{
\newpage
\pagestyle{empty} 
\clearpage
\mbox{}\newpage
}
{\pagestyle{empty}

{
\begin{center}

\noindent
\textbf{Acknowledgements} \\ \vskip .5cm
\end{center}
}

\noindent 

 \vskip .3cm

\noindent It was amazing and exciting to do a PhD. However, to be honest, it was the most stressful. I have never faced the number of new things that I faced during this time, and I have never walked this far out of my comfort zone. As a result, I have learned more than any stage in my life. None of these was possible without the support and the love I have received from the people around me. 

Foremost, I would like to thank Christof for trusting me not only when he gave me this opportunity to start as his PhD student, but also during these years. I remember you asked me in one of the interviews that what I would do if I burn out during my PhD. That actually happened. I went through a serious burnout, but I managed to overcome it just as I explained in response to your question. However, this was not possible without you trusting me and without you being so patient. I am so grateful, and I feel indebted to you for all the support. Thanks for all the discussions, critical feedbacks and useful pieces of advice. 

Next, I would like to thank Maarten for his support and for creating such a great environment at ILPS for the researchers to flourish. Thanks for your leadership, for your insightful bits of advice, all the feedback on my thesis and the presentations.

I am thankful to my colleagues at SMT group for all the pleasant moments we had together. Arianna, Ivan, Katya, Ke, Marlies, Marzieh, Praveen thanks for the supportive comments at times of failure, sharing happiness at times of success, interesting discussions, direct feedbacks, and proofreadings of the papers. I enjoyed every moment of working with you and I learned a lot from you. 

I am happy to have Marlies and Bob as my paranymphs. Thank you for proofreading my thesis, answering my questions and more importantly accepting to fight on my side. I would like to thank my other friends who also helped with proofreading the thesis. Cristina, Livia and Maartje, thank you for helping me with that.

Ekaterina, Grzegorz, Josef, Paul and Theo, thanks for accepting to be my committee members, and for your valuable time that you have spent on it. 

I enjoyed being part of ILPS thank to all members of this amazing group. You all contributed to making these years of doing the PhD such an incredible experience to me. Thank you, Adith, Aleksandr, Alexey, Ali A, Ali V, Amir, Ana, Anna, Anne, Arezoo, Arianna, Artem, Bob, Boris, Chang, Christof, Christophe, Chuan, Cristina, Daan, Dan, David, David, Dilek, Eva, Evangelos, Evgeny, Fei, Harrie, Hendra, Hendrik, Hendrike, Hinda, Hosein, Isaac, Ilya, Ivan, Jiahuan, Jie, Julia, Julien, Kaspar, Katya, Ke, Lars, Maarten, Maarten, Maartje, Mahsa, Manos, Marc, Marlies, Mariya, Marzieh, Maurits, Mozhdeh, Mostafa, Nikos, Olivier, Pengjie, Petra, Praveen, Richard, Ridho, Rolf, Shangsong, Shaojie, Spyretta, Svitlana, Thorsten, Tobias, Tom, Trond, Vera, Xiaohui, Xiaojuan, Xinyi, Yaser, Yangjun, Yifan, Zhaochun and Ziming. Thanks to Ana, Maartje, Marlies and Maurits once more for being my nicest officemates. A special thank to Anne, Ivan and Marlies for their wise bits of advice and supportive words at the desperate times of failure. And last, but not least, a big thank to Isaac for always asking questions, even nonsense ones.

Thanks to Maryam for the design of the cover, which I really like. 

I would also like to thank my friends who made life in the Netherlands easier for me by being there at the times of fun and failure. Thanks to Ali, Ali, Amin, Amirhosein, Aylar, Azad, Azadeh, Behrouz, Bob, Cristina, Danial, Fahimeh, Fatemeh, Hadi, Hester, Hoda, Hoda, Hojat, Hoora, Hosein, Irene, Jurre, Keyvan, Livia, Maartje, Mahdi, Mahdieh, Marlies, Maryam, Marzieh, Masoud, Mehran, Mehran, Mohammad, Mohammad Hosein, Mojtaba, Mostafa, Narges, Naser, Nasrin, Parisa, Samira, Samira, Samy, Sara, Shayan, Shima, Tim, Tom, Vahid, Vivianne, Zahra, Zoheir. 

I did an internship at the National Research Institute of Japan during which I enjoyed working with Satoshi sensei, Anna, Aron, Ben, Dianna, Hassan, Jerome, Kentaro, Koki, Thomas, Tiphaine. 

I thank my siblings, Mobina and Omid for encouraging me and supporting me to take this path of doing a PhD, and three of my cousins, Mansour, Masoud and Mona who are as close as my siblings to me and have always supported me and inspired me with their kindness and their dedication. Special thanks to Masoud for being my role model.  

I would like to express my utmost gratitude to my mom and dad. This was impossible without you, your love and unconditional support. You are the most hard-working people I have ever seen. This has always been inspiring to me and was a source of motivation to me. I know and deeply appreciate all the sacrifices you have made to make this dream of mine comes true. 

My deepest gratitude and appreciation goes to my beautiful wife, Marzieh, who is my closest friend before anything else. Thank you so much for all the support and love you gave to me during these years. You played the role of a colleague at work and the best friend at life. No one could help me as you did during the hard times of failure and disappointment. You were always there by my side with your deepest feelings at life and the smartest ideas at work. Thank you for your presence during all we have been through.

\bigskip

\null\hfill Hamidreza\\
\null\hfill September 2020

\clearpage
}
\tableofcontents

\acrodef{IR}{information retrieval}


\setlength{\parskip}{0pt}
\mainmatter

\chapter{Introduction}
\label{chapter:introduction}


Until recently, phrase-based machine translation was the state-of-the-art for more than a decade~\citep{och-etal-1999-improved, och-ney-2002-discriminative, och-2003-minimum, Koehn:2003:SPT:1073445.1073462, Tillmann:2004:UOM:1613984.1614010, chiang2007hierarchical, Galley:2008:SEH:1613715.1613824, cherry-foster-2012-batch}. Phrase-based machine translation models have recently been challenged and outperformed by several neural machine translation (NMT) models~\citep{sutskever2014sequence,bahdanau-EtAl:2015:ICLR,luong-EtAl:2015:ACL-IJCNLP,jean-EtAl:2015:ACL-IJCNLP,wu2016google,NIPS2017_7181}. Despite their differences, both phrase-based and neural machine translation models are sophisticated statistical approaches that use a large amount of bilingual data to learn translation models.

Phrase-based models are a combination of different models, each responsible for capturing a specific aspect of the translation task. These models mainly consist of a \textit{translation model}, a \textit{phrase reordering model} and a \textit{language model} \citep{Koehn:2003:SPT:1073445.1073462}. This composition makes phrase-based models more modular than neural machine translation models. However, the phrase reordering model in particular is complex. It is responsible for the difficult task of translating phrases in the right order and is still not fully understood. To give an example, it is not clear which words inside a phrase-pair are the most important features for a phrase reordering model. Interpretability of the phrase reordering model is important for two reasons. First, it helps natural language processing (NLP) engineers and practitioners to better adapt the model to their specific use cases. Second, it can benefit our understanding of how the even more complex neural machine translation systems handle similar linguistic phenomena. 



Neural machine translation models are end-to-end models that do not split the responsibilities between multiple models as is the case for phrase-based machine translation. This end-to-end nature makes neural machine translation models less interpretable than phrase-based models~\citep{DBLP:phd/ndltd/Belinkov18, stahlberg-etal-2018-operation}. 

Both phrase-based and neural machine translation systems are complex systems that aim to capture complex linguistic phenomena in natural language. In this thesis, we study how we can contribute to the interpretability of both machine translation frameworks. For this study, we follow an empirical probing approach. To this end, we choose three important aspects of these models and investigate their behavior for capturing different syntactic phenomena. We study phrase reordering models in phrase-based systems and attention modeling and hidden state representations in neural machine translation systems.

The two most successful phrase reordering models in phrase-based machine translation are the lexicalized reordering model (LRM)~\citep{Tillmann:2004:UOM:1613984.1614010,koehn05iwslt} and the hierarchical reordering model (HRM)~\citep{Galley:2008:SEH:1613715.1613824}. These models use the full lexical form of phrase-pairs to decide on the correct reordering of the phrase-pairs. However, it is not clear whether it is necessary to use the full lexical form to predict the reordering distribution with reasonable accuracy. LRM and HRM both suffer from the problem of learning unreliable distributions for infrequent phrase-pairs. This problem could be mitigated if it were possible to estimate the reordering distributions using some generalization of the phrase-pairs.

In Chapter~\ref{chapter:research-01}, we investigate the influence of the words in a phrase on the reordering behavior of the phrase. We reduce the problem of infrequent phrase-pairs by eliminating or generalizing less essential words. 

In the next chapters, we shift our study towards neural machine translation models, since these models are shown to better model different aspects of translation including lexical translation and word reordering~\citep{bentivogli-etal-2016-neural}.


Attention-based neural machine translation models~\citep{bahdanau-EtAl:2015:ICLR, DBLP_journals_corr_LuongPM15} have become popular due to their ability to use the most relevant parts of the source sentence for each translation step. In the next part of our study, after having studied phrase reordering models, we investigate what the relevant parts are that the attention model attends to for different linguistic phenomena. Additionally, we compare the attention model with the traditional alignment model to find the differences and similarities between these two models. Next, we examine the distributional behavior of an attention model for different syntactic phenomena within our empirical probing methodology. Understanding attention behavior is important for the interpretability of neural machine translation models since it defines the importance of source hidden states for generating a target word.

We expand our study of interpretability to the hidden states of the encoder of neural machine translation models. These are the hidden states encoding the source side information that the attention models attend to while generating target words. Previous studies use an extrinsic approach of feeding the hidden state representations into different classifiers to find out what information is encoded in those representations~\citep{shi-padhi-knight:2016:EMNLP2016,belinkov2017neural}. Other work has also used such extrinsic approaches to compare different neural sequence-to-sequence architectures in terms of capturing syntactic structure and lexical semantics by the encoder hidden state representations~\citep{D18_Recurrent, D18-1458_Self-Attention}. However, to the best of our knowledge, there is no work that uses an intrinsic approach to study the information captured by the hidden state representations and investigates the difference of these representations with the underlying word embedding representations. The advantage of adopting an intrinsic approach compared to an extrinsic approach is that it enables us to understand \emph{what} information is captured by the hidden states, as well as \emph{how} this information is captured.

In Chapter~\ref{chapter:research-03}, we analyse the hidden state representations of different sequence-to-sequence models from a nearest neighbor perspective. We investigate the information captured by the hidden state representations that goes beyond what is transferred from word embeddings. Additionally, we compare two common neural machine translation architectures with respect to the extent to which they capture syntactic and lexical semantic information in their hidden states. This comparison is especially important to our study of interpretability as contrasting two different, yet closely related models allows us to focus our study on the aspects in which both models differ from each other.


\section{Research Outline and Questions}
\label{section:introduction:rqs}

\acrodef{rq:reorderingMain}[\ref{rq:reorderingMain}]{\textit{To what extent do the orderings of phrases in a phrase-based machine translation model depend on the full lexical forms of the phrases?}}
\acrodef{rq:reorderingSub1}[\ref{rq:reorderingSub1}]{\textit{How does the importance of the words of a phrase change in defining the ordering of the phrase?}}
\acrodef{rq:reorderingSub2}[\ref{rq:reorderingSub2}]{\textit{How helpful is it to estimate the reordering distribution of phrases by backing off to shorter forms of the phrases and removing less influential words?}}
\acrodef{rq:reorderingSub3}[\ref{rq:reorderingSub3}]{\textit{How accurately can the orderings of phrases be predicted by using class-based generalizations of the words in phrases when keeping the most influential words in their original form?}}
\acrodef{rq:attnMain}[\ref{rq:attnMain}]{\textit{What are the important words on the source side that attention-based neural models attend to while translating different syntactic phenomena?}}
\acrodef{rq:attnSub1}[\ref{rq:attnSub1}]{\textit{To what extent does an attention model agree with the traditional alignment model in capturing translation equivalent words?}}
\acrodef{rq:attnSub2}[\ref{rq:attnSub2}]{\textit{Are the differences between attention and traditional alignments due to errors in attention models or do attention models capture additional information compared to traditional alignment models?}}
\acrodef{rq:attnSub3}[\ref{rq:attnSub3}]{\textit{How does the distribution of the attention model change for different syntactic phenomena?}}
\acrodef{rq:attnSub4}[\ref{rq:attnSub4}]{\textit{What types of words are attended to by the attention model while translating different syntactic phenomena?}}
\acrodef{rq:hiddenStateMain}[\ref{rq:hiddenStateMain}]{\textit{What information is captured by the hidden states of the encoder of a neural machine translation model?}}
\acrodef{rq:hiddenStateSub1}[\ref{rq:hiddenStateSub1}]{\textit{How does the captured information in the hidden states differ from the information captured by the word embeddings?}}
\acrodef{rq:hiddenStateSub2}[\ref{rq:hiddenStateSub2}]{\textit{To what extent do the hidden states of the encoder capture different types of information, such as syntactic or semantic information?}}
\acrodef{rq:hiddenStateSub3}[\ref{rq:hiddenStateSub3}]{\textit{How different is the captured information in the encoder hidden states for different NMT architectures?}}

In this thesis, we study phrase-based and neural machine translation models and shed light on how different syntactic and semantic phenomena in natural language are captured by these models. This research is an attempt to increase the interpretability of the complex machine translation systems. Here, we study phrase reordering models in phrase-based machine translation, attention models in neural machine translation and the internal hidden states also in neural machine translation models. We structure our studies by asking three main research questions and further divide them to subquestions. We ask the following research questions:

\begin{enumerate}[label=\textbf{RQ\arabic*},ref={RQ\arabic*}, wide, labelwidth=!, labelindent=0pt]
\item \acl{rq:reorderingMain}\label{rq:reorderingMain}
\end{enumerate}

\noindent LRMs and HRMs both use the relative frequency of the orientations for phrases in the training corpus to estimate the distributions over orientations conditioned on phrase-pairs. In the case of infrequent phrase-pairs, both models suffer from the problem of insufficient observations to allow for a reliable estimate of the corresponding distributions. Originally, these models rely on the full lexical forms of the phrases when they count the orientations given the phrases. \ref{rq:reorderingMain} investigates whether it is necessary to always use the full lexical form of the phrases when estimating the corresponding distributions. 

In the subquestions below, we elaborate more on the primary research question. We approach the problem by first asking which words from the inside of a phrase have most influence on defining the reordering behavior of the phrase. This may result in ignoring less important words, allowing us to use shorter phrase-pairs that decrease the chance of being rarely observed. To this end, we ask the following subquestion:
 
    \begin{enumerate}[label=\textbf{RQ\arabic*.1}, ref={RQ\arabic*.1}, wide, labelwidth=!, labelindent=0pt]
    \item \acl{rq:reorderingSub1}\label{rq:reorderingSub1}
    \end{enumerate}

    
\noindent We answer this subquestion by experimenting with different patterns for backing-off or marginalizing to shorter sub-phrase-pairs.         

In our backing-off experiments, we assume that the importance of internal words of a phrase for reordering changes with their proximity to the borders of the phrase. In these cases, we consider border words as the most important words for defining reordering following~\citet{nagata2006clustered} and~\citet{cherry:2013:NAACL-HLT}.

In our marginalizing experiments, we consider \textit{exposed heads}~\citep{chelba2000structured, li2012head} as the important words for reordering, where exposed heads are defined by dependency parse relations. Exposed heads are words from inside a sequence that are in dependency relations with words outside of the sequence and are dominated by those words \citep{li2012head}.

Next, we investigate to what extent we can estimate the reordering distribution of phrases by removing less important words and use only part of the phrase for estimation, analogous to back-off smoothing in n-gram language models. To this end, we ask the following subquestion:

    \begin{enumerate}[label=\textbf{RQ\arabic*.2}, ref={RQ\arabic*.2}, wide, labelwidth=!, labelindent=0pt]
    \item \acl{rq:reorderingSub2}\label{rq:reorderingSub2}
        \end{enumerate}

\noindent To answer this question, we use different back-off strategies to estimate the reordering distributions of phrase-pairs. We use the resulting reordering model in a phrase-based machine translation system and compare the results with systems based on commonly used reordering models.

    Furthermore, we investigate to what extent we can better estimate the distribution of the infrequent phrases by applying class-based generalization for less influential words (\ref{rq:reorderingSub3}):

	\begin{enumerate}[label=\textbf{RQ\arabic*.3}, ref={RQ\arabic*.3}, wide, labelwidth=!, labelindent=0pt]
    \item \acl{rq:reorderingSub3}\label{rq:reorderingSub3}
    \end{enumerate}
    
\noindent We answer this question by following the same approach used to answer subquestion~\ref{rq:reorderingSub2}. We create multiple phrase reordering models using different phrase generalization patterns and compare the results of using them in the translation system with other reordering models.
    
In summary, we answer the questions in Chapter~\ref{chapter:research-01} by (i) investigating two backing off strategies to use sub-phrase-pairs of the original phrase-pairs to smooth the orientation distribution of the original phrase-pairs, and (ii) experimenting with a phrase generalization approach that uses the linguistic notion of exposed heads~\citep{chelba2000structured, li2012head} to define which words to generalize and which words to keep in their original forms.

In Chapter~\ref{chapter:research-02}, we continue investigating the importance of words, but this time from a neural machine translation perspective. As mentioned above, attention-based neural machine translation models~\citep{bahdanau-EtAl:2015:ICLR, DBLP_journals_corr_LuongPM15} have achieved popularity due to their capability to use the most relevant parts of a source sentence for each translation step. However, it is unclear what exactly the definition of relevance should be. So the following research question aims to shed light on the criteria of relevance.

\begin{enumerate}[label=\textbf{RQ\arabic*},ref={RQ\arabic*}, wide, labelwidth=!, labelindent=0pt, resume]    
\item \acl{rq:attnMain}\label{rq:attnMain} 
\end{enumerate}

\noindent Earlier machine translation research stipulates that attention models in neural machine translation are similar to traditional alignment models in phrase-based machine translation~\citep{alkhouli-EtAl:2016:WMT, cohn-EtAl:2016:N16-1, liu-EtAl:2016:COLING, chen2016guided}. As a result, there have been some attempts to train attention models with explicit signals from existing traditional alignment models~\citep{alkhouli-EtAl:2016:WMT, liu-EtAl:2016:COLING, chen2016guided}. However, these attempts yield improvements in some domains and no improvements or even drops in performance in other domains. To answer \ref{rq:attnMain}, we subdivide it into the following subquestions: 

    \begin{enumerate}[label=\textbf{RQ\arabic*.1} , ref={RQ\arabic*.1}, wide, labelwidth=!, labelindent=0pt, start=2]
    \item \acl{rq:attnSub1}\label{rq:attnSub1}
    \end{enumerate}
    
\noindent In~\ref{rq:attnSub1}, we question the similarity of traditional alignments in phrase-based machine translation with attention models in neural machine translation. Of course, there are some similarities between these models~\citep{cohn-EtAl:2016:N16-1}, but we aim to test the extent of this similarity empirically.

We answer~\ref{rq:attnSub1} by comparing attention and alignment models using both pre-existing and new measures introduced in our research.

Attention models and traditional alignment models are essentially different, since attention models learn distributions resulting in a soft alignment, whereas traditional alignments are hard alignments models. However, the question is what advantages are brought about by this difference. So we ask the following subquestion:

    \begin{enumerate}[label=\textbf{RQ\arabic*.2} , ref={RQ\arabic*.2}, wide, labelwidth=!, labelindent=0pt, start=2]
    \item \acl{rq:attnSub2}\label{rq:attnSub2}
    \end{enumerate}
    
\noindent We answer \ref{rq:attnSub2} by comparing the distributional behavior of attention and the alignment model and their correlation with translation quality. 

We observe that attention is not as static as the alignment model. This motivates us to study this phenomena more closely as it could be an advantage of attention model over the alignment model. So we ask the following research subquestion:

    \begin{enumerate}[label=\textbf{RQ\arabic*.3} , ref={RQ\arabic*.3}, wide, labelwidth=!, labelindent=0pt, start=2]
    \item \acl{rq:attnSub3}\label{rq:attnSub3}
    \end{enumerate}

\noindent \ref{rq:attnSub3} requires a more in-depth investigation of the attention model by analyzing its distributional behavior for different syntactic phenomena and its impact on translation quality. We define an entropy-based metric to measure how focused or dispersed attention weights are for different syntactic phenomena. We also investigate the correlation of this metric with translation quality and alignment quality to further analyze the distributional behavior of the attention model.
  
      \begin{enumerate}[label=\textbf{RQ\arabic*.4} , ref={RQ\arabic*.4}, wide, labelwidth=!, labelindent=0pt, start=2]
    \item \acl{rq:attnSub4}\label{rq:attnSub4}
    \end{enumerate}
      
\noindent \ref{rq:attnSub4} investigates what word types the attention model attends to on the source side while generating different syntactic types on the target side. For example, what types of words are attended to on the source side when the model generates a noun on the target side? Such an analysis can help explain the differences between the attention model and the alignment model if there are any.

In Chapter~\ref{chapter:research-02}, we answer these questions by defining metrics to compare attention models to traditional alignment models. We empirically show that these models are different and attention models capture more information than alignment models. We study the behavior of the attention model for different syntactic phenomena and provide an in-depth analysis of what information the attention model attends to during translation.

\noindent Neural machine translation models follow a general encoder-decoder architecture that encodes the source sentence into distributed representations and then decodes these representations into a sentence in the target language. At each target word generation step, the attention model computes a weighted sum over the encoder hidden representations. This weighted sum is used by the decoder to generate the next word in the target sentence. So far, we asked questions about the behavior of the attention model and the relevant words that the attention model attends to while generating different syntactic phenomena. That means we have assumed that the corresponding hidden state of a source word is representative of that word. However, it is not precisely clear what linguistic information from the source words is encoded in these hidden states. Assuming the corresponding hidden states of a source word to be representative of the source word means that the lexical information encoded in the word embeddings should be transferred to corresponding hidden states. To investigate what information is captured by the hidden states, we ask the following research question:

\begin{enumerate}[label=\textbf{RQ\arabic*},ref={RQ\arabic*}, wide, labelwidth=!, labelindent=0pt, resume ]
\item \acl{rq:hiddenStateMain}\label{rq:hiddenStateMain}
\end{enumerate}

\noindent \ref{rq:hiddenStateMain} generally asks about the information encoded in the hidden states. In the subquestions, we break this investigation into more fine-grained steps by questioning the similarity and difference to the underlying word embeddings.

\begin{enumerate}[label=\textbf{RQ\arabic*.1}, ref={RQ\arabic*.1}, wide, labelwidth=!, labelindent=0pt,  start=3]
    \item \acl{rq:hiddenStateSub1}\label{rq:hiddenStateSub1}
    \end{enumerate}

\noindent In~\ref{rq:hiddenStateSub1}, we question the similarity and the difference of the hidden states with the corresponding word embeddings. Word embeddings are distributed representations of words, encoding the general lexical information about the corresponding words. Hence, if no similarity existed between the hidden states and the corresponding word embeddings, then taking the hidden states as representative of the corresponding source word would not be a correct assumption. 

We answer this question by comparing the lists of the nearest neighbors of hidden states with the nearest neighbors of the corresponding word embeddings. We define a quantitative measure for this comparison. This is the first measure used by our intrinsic study.
        
In the next step, we ask to what extent the difference between the hidden states and the word embeddings is due to the syntactic and lexical semantic information captured by the hidden states: 
         
\begin{enumerate}[label=\textbf{RQ\arabic*.2}, ref={RQ\arabic*.2}, wide, labelwidth=!, labelindent=0pt,  start=3]
    \item \acl{rq:hiddenStateSub2}\label{rq:hiddenStateSub2}
    \end{enumerate}
    
\noindent Earlier work has fed the hidden state representations into diagnostic classifiers to reveal the syntactic and semantic information captured by the hidden states~\citep{shi-padhi-knight:2016:EMNLP2016,belinkov2017neural}. However, we take a more intrinsic approach to answer research question~\ref{rq:hiddenStateSub2}. We investigate the nearest neighbors of the hidden states in terms of the captured syntactic structure and the coverage of WordNet~\citep{wordnet_book, Miller:1995:WLD:219717.219748} connections to answer this question. We define two intrinsic measures to quantify the captured information.

Once we have defined our intrinsic measures to interpret the hidden state representation of neural machine translation models, we are able to compare different neural machine translation architectures considering the information they encode from the source side. So, to study how the captured information differs in different NMT architectures we ask the following subquestion:

 \begin{enumerate}[label=\textbf{RQ\arabic*.3}, ref={RQ\arabic*.3}, wide, labelwidth=!, labelindent=0pt,  start=3]
    \item \acl{rq:hiddenStateSub3}\label{rq:hiddenStateSub3}
    \end{enumerate}
    
\noindent \ref{rq:hiddenStateSub3} investigates whether different NMT architectures capture the same information to the same extent. In particular, we investigate how the hidden states in a recurrent neural model~\citep{bahdanau-EtAl:2015:ICLR,luong-EtAl:2015:ACL-IJCNLP} are different from the hidden states in a transformer model~\citep{NIPS2017_7181}.
  
In Chapter~\ref{chapter:research-03}, we answer these questions by looking into the nearest neighbor list of the hidden states. We compare these lists to the lists of the nearest neighbors of the corresponding word embeddings to find differences. We also look up the nearest neighbors in the WordNet connections of the corresponding source word of a hidden state to define a metric to measure lexical semantics. We use this metric to compare different models in terms of their capability to capture lexical semantics. We also compare different models by computing the similarity of the local syntactic structures of the nearest neighbors of a hidden state to the local syntactic structure of the corresponding word of the hidden state. We show that transformer models are better in capturing lexical semantics, whereas recurrent models are superior in capturing syntactic structure. Our observation is in line with the observed results of the extrinsic comparison of these models by~\cite{D18_Recurrent} and~\cite{D18-1458_Self-Attention}.


\section{Main Contributions}
\label{section:introduction:contributions}

In this thesis, we mainly contribute to the interpretability of phrase-based and neural machine translation models. We provide algorithmic contributions as well as metrics to study the behavior of neural models. We also make empirical contributions by providing a more detailed analysis of the models based on our metrics.
\subsection{Algorithmic Contributions}

\begin{enumerate}
\item In Chapter~\ref{chapter:research-01}, we propose two algorithms to use shortened forms of a phrase-pair to smooth the orientation distribution of the original phrase-pair following back-off smoothing in n-gram language modelling~\citep{chen1999empirical}. These algorithms are applicable to lexicalized~\citep{Tillmann:2004:UOM:1613984.1614010,koehn05iwslt} and hierarchical reordering models~\citep{Galley:2008:SEH:1613715.1613824} yielding slight improvements for these models. Our algorithms use linear interpolation and recursive MAP~(Maximum a Posteriori) smoothing~\citep{cherry:2013:NAACL-HLT, chen-foster-kuhn:2013:NAACL-HLT} to combine the shortened forms of the phrase-pairs. 

\item In Chapter~\ref{chapter:research-01}, we also propose four methods to use generalized forms of a phrase-pair to smooth the orientation distribution of the original phrase-pair. These generalized forms are produced by keeping essential words and marginalizing other words. Our proposed methods are applicable to both LRM and HRM models and improve the reordering performance of these models.
\item In Chapter~\ref{chapter:research-02}, we propose an approach to compare attention with traditional alignment. We define various metrics to measure the agreement between attention and alignment and to investigate the relationship of attention and alignment agreement with translation quality.

\item In Chapter~\ref{chapter:research-03}, we propose an intrinsic approach to study the syntactic and lexical semantic information captured by the hidden state representations based on their nearest neighbors.

\item In Chapter~\ref{chapter:research-03}, we define metrics that provide interpretable representations of the information captured by the hidden states in NMT systems, highlighting the differences between hidden state representations and word embeddings.

\end{enumerate}

\subsection{Empirical Contributions}

\begin{enumerate}

\item In Chapter~\ref{chapter:research-01}, for the phrase reordering models, we show how keeping essential words of a phrase-pair and marginalizing out the other words lead to generalized forms of the phrase-pair that improves the performance of the reordering model when being used to smooth the reordering distribution. 


\item In Chapter~\ref{chapter:research-01}, we provide an in-depth analysis showing that orientation distributions conditioned on long phrase-pairs typically depend on a few words within phrase-pairs and not the whole lexicalized form. As a result, using generalized forms of the phrase-pairs to smooth the original reordering distribution leads to performance improvements of the reordering models.

\item In Chapter~\ref{chapter:research-02}, we provide a detailed comparison of an attention model in neural machine translation against a word alignment model. We show that the attention and the alignment models have different behaviors and attention is not necessarily an alignment model.  

\item In Chapter~\ref{chapter:research-02}, we also show that while different attention mechanisms can lead to different degrees of compliance with respect to word alignments, a full compliance is not always helpful for word prediction.

\item In Chapter~\ref{chapter:research-02}, we additionally show that attention follows different patterns depending on the type of word being generated. We contribute to the interpretability of the attention model by providing a detailed analysis of the attention model. Our analysis explains the mixed observations of earlier works that have used alignments as an explicit signal to train attention models~\citep{chen2016guided, alkhouli-EtAl:2016:WMT, liu-EtAl:2016:COLING}.

\item  In Chapter~\ref{chapter:research-02}, we provide evidence showing that the difference between attention and alignment is due to the capability of the attention model to attend to the context words influencing the current word translation. We show that the attention model attends not only to the translation equivalent of the word being generated, but also other source words that may have some effect on the form of the target word. For example, while generating a verb on the target side, the attention model often attends to the preposition, subject or object of the translation equivalent of the verb on the source side.



\item In Chapter~\ref{chapter:research-03}, we use our proposed intrinsic approach for the analysis of syntactic and lexical semantic information captured by the hidden states to compare transformer and recurrent models in terms of their capabilities to capture this kind of information. We show that transformer models are superior in capturing lexical semantic information, whereas recurrent neural models are better in capturing syntactic structures.

\item In Chapter~\ref{chapter:research-03}, we additionally provide analyses of the behavior of the hidden states for each directional layer and the concatenation of the states from the directional layers in a recurrent neural model. We show that the reverse recurrent layer captures more lexical semantic information. In contrast, the forward recurrent layer captures more extended context information.

\end{enumerate}


\section{Thesis Overview}
\label{section:introduction:overview}

The rest of this thesis is organized as follows. Chapter~\ref{chapter:background} discusses background and related work. The main research is presented in the following three chapters (Chapters~\ref{chapter:research-01}--\ref{chapter:research-03}). Chapter~\ref{chapter:conclusions} summarizes the main findings of this thesis. Below, we provide a high-level summary of the content of each chapter.

\begin{itemize}
\item \textbf{Chapter~\ref{chapter:background} (Background)} gives an introduction to phrase-based and neural machine translation, which are the main machine translation approaches of recent years and are used in the research presented in this thesis. We discuss phrase reordering models including the lexicalized reordering model (LRM)~\citep{Tillmann:2004:UOM:1613984.1614010,koehn05iwslt} and hierarchical reordering model (HRM)~\citep{Galley:2008:SEH:1613715.1613824}, which are the focus of the research presented in Chapter~\ref{chapter:research-01}. Next, we provide an introduction to neural machine translation and attention models in neural machine translation, which are the focus of Chapter~\ref{chapter:research-02}. We also briefly talk about alignment models to be referred back to in Chapter~\ref{chapter:research-02}, where we show its contrast with attention models. Subsequently, we discuss hidden state representations and word embeddings in neural machine translation. We also discuss the transformer neural machine translation model, since we compare this model with recurrent machine translation model using their hidden representations in Chapter~\ref{chapter:research-03}.

\item \textbf{Chapter~\ref{chapter:research-01} (The Impact of Internal Words on Phrase Reorderings)} introduces the problem of reliable estimation of the phrase reordering distribution for infrequent phrase-pairs. We briefly discuss previous methods proposed to improve estimation. We answer~\ref{rq:reorderingMain} by experimenting with different ways of shortening phrase-pairs following the idea of backing off in n-gram language models. We experiment with keeping essential words of the phrase-pairs and removing or using generalized forms of the remaining words. We show that the latter approach achieves improvements over strong models based on LRM and HRM.

\item \textbf{Chapter~\ref{chapter:research-02} (What does Attention in Neural Machine Translation Pay Attention to?)} provides a detailed analysis of the attention model in neural machine translation. We answer~\ref{rq:attnMain} by comparing the attention and alignment models. We define metrics to analyze the behavior of the attention model for different syntactic phenomena. We use part of speech (POS) tags to provide a more fine-grained analysis of the attention behavior. In this chapter, we show that the discrepancies between attention and alignment are not due to errors in attention for several specific syntactic phenomena. We also show what word types are being attended to when generating specific syntactic word forms and discuss why these attentions are intuitive.

\item \textbf{Chapter~\ref{chapter:research-03} (Interpreting Hidden Encoder States)} takes the research presented in Chapter~\ref{chapter:research-02} further to study the information captured by the hidden states of the encoder side of a neural machine translation model. 
We compare hidden states and word embeddings to reveal the information captured by the hidden states on top of the information captured by embeddings. We investigate what share of the nearest neighbors list of the hidden states are covered by the direct connections of the corresponding source words in WordNet to measure how much of the capacity of the hidden states is capturing lexical semantics. We also look into the similarities of the syntactic structures captured by the nearest neighbors and the hidden states. We use these intrinsic methods to compare recurrent and transformer architectures in terms of capturing syntactic and lexical semantic information. This chapter answers~\ref{rq:hiddenStateMain}.

\item \textbf{Chapter~\ref{chapter:conclusions} (Conclusion)} concludes this thesis by reviewing the research questions and our answers as the result of this research. We also provide a brief discussion of how this thesis has contributed to the interpretability of phrase-based and neural machine translation models. Finally, we discuss problems that remain open for further studies.

\end{itemize}


\section{Origins}
\label{section:introduction:origins}

The research presented in the Chapters \ref{chapter:research-01}--\ref{chapter:research-03} is based on the following peer reviewed publications. 

\begin{itemize}
\item \textbf{Chapter~\ref{chapter:research-01}} is based on~\cite{ghader2016words}, \textit{Which Words Matter in Defining Phrase Reorderings in Statistical Machine Translation?},  published in the Proceedings of the Twelfth Conference of the Association for Machine Translation in the Americas (AMTA 2016), Volume 1: MT Researchers' Track, pages 149--162, Austin, TX, USA, Association for Machine Translation in the Americas.

The back-off model was proposed by Monz and the generalization model was proposed by Ghader. Experiments and analysis were performed by Ghader. Both authors contributed to the writing of the article. Ghader did most of the writing.

\item \textbf{Chapter~\ref{chapter:research-02}} is based on~\cite{ghader2017does}, \textit{What does Attention in Neural Machine Translation Pay Attention to?}, published in the Proceedings of the Eighth International Joint Conference on Natural Language Processing (IJCNLP2017), Volume 1: Long Papers, pages 30--39, Taipei, Taiwan, Asian Federation of Natural Language Processing.

Ghader proposed the analysis of attention and carried out experiments and analysis. The article was mostly written by Ghader. Monz contributed to the article and discussion.

\item \textbf{Chapter~\ref{chapter:research-03}} is based on~\cite{ghader2019}, \textit{An Intrinsic Nearest Neighbor Analysis of Neural Machine Translation Architectures}, published in the Proceedings of Machine Translation Summit XVII, Volume 1: Research Track, pages 107--117, Dublin, Ireland, European Association for Machine Translation.

Monz suggested the analysis of hidden states. Experiments and analysis were performed by Ghader. Both authors contributed to the article. Ghader did most of the writing.

\end{itemize}
\graphicspath{ {02-background/images/} }

\chapter{Background}
\label{chapter:background}

In this chapter, we introduce the two main machine translation paradigms that are used in this thesis. Phrase-based machine translation is used in Chapter~\ref{chapter:research-01} and two different types of neural machine translation architecture are used in Chapters~\ref{chapter:research-02} and \ref{chapter:research-03}. We discuss the components of the models in detail, focusing on the components that are relevant to this thesis.

\section{Phrase-Based Machine Translation}

In phrase-based machine translation~\citep{koehn05iwslt, Koehn:2003:SPT:1073445.1073462, koehn2007moses}, continuous sequences of words (phrases) play the role of the translation unit. These translation units come in pairs of source and target language phrases. The phrase-pairs are learned from bilingual parallel data using unsupervised learning algorithms. In the next sections, we explain how these translation units are learned from data and are used to translate from the source language into the target language.

\subsection{Word Alignment}
\label{subsec:word_alignment}

The translation units (phrase-pairs) for phrase-based machine translation are learned from a parallel corpus consisting of a large body of text in a source and a target language which is parallel at the sentence level. This means that each sentence in the target language is the translation of the corresponding sentence in the source language~\citep{Koehn:2003:SPT:1073445.1073462}. Parallel corpora are typically comprised of hundreds of thousands or millions of sentence pairs. 

\begin{figure}[thb]
\centering
\includegraphics[scale=0.50, width=0.60\textwidth]{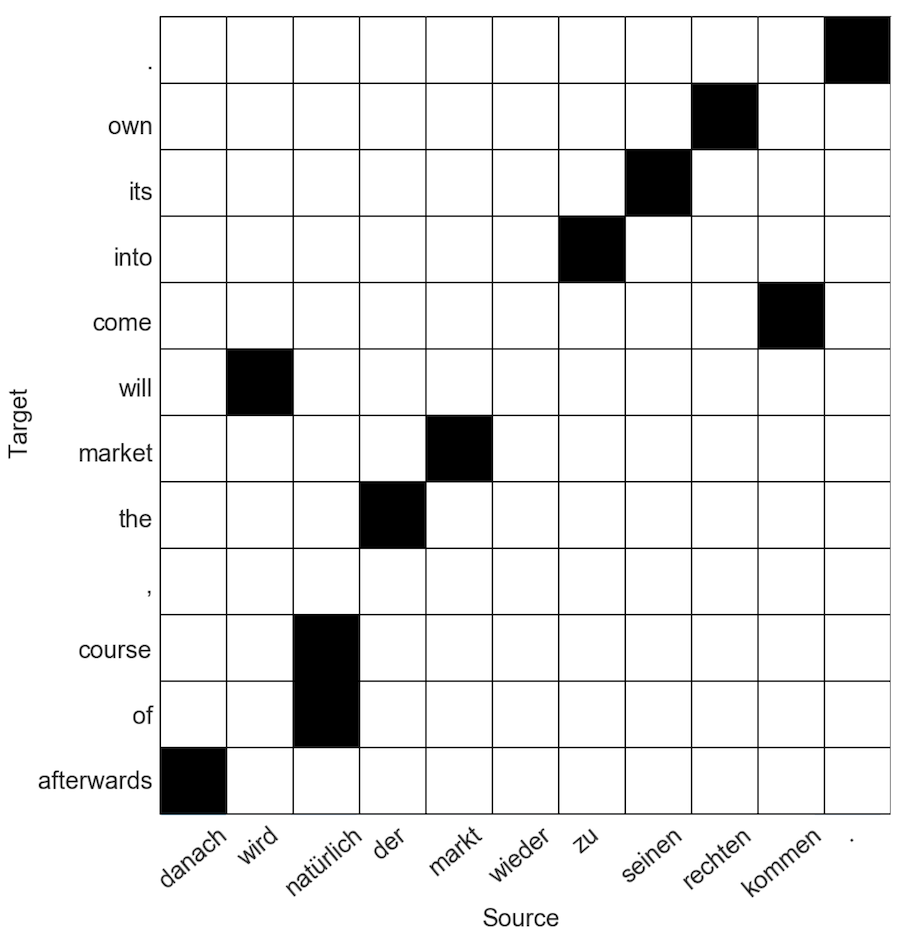}
\caption{An example of a word alignment matrix. As shown, words from the source and target can remain unaligned or can have more than one aligned words on the other side.}
\label{fig:alignment_matrix}
\end{figure}

In order to learn the phrase-pairs from the parallel corpus, the parallel corpus should be word aligned first. This means that words from the source side are aligned with their corresponding word or words from the target side. It can also happen that a word from one side has no corresponding word on the other side to be aligned with. For example, in Figure \ref{fig:alignment_matrix}, the word ``wieder" on the German side and the comma on the english side have no corresponding word on the other side and hence are left unaligned. It can also happen that one word from the target side is aligned with multiple words from the source side. Therefore, commonly used word alignments are many-to-many~\citep{Koehn:2003:SPT:1073445.1073462}.

Figure \ref{fig:alignment_matrix} shows an example of a word alignment matrix. Word alignments were first introduced as part of the IBM models \citep{brown-etal-1993-mathematics}. IBM models produce one-to-many word alignments. To produce a many-to-many alignment, two alignments are produced by running IBM models on the source-to-target and the inverse direction. Then, the two alignments are merged into one many-to-many alignment in a process called symmetrization. The extreme cases of symmetrization are the union and intersection of the two alignments. However, it has been shown that heuristics that explore the space between the intersection and the union of the two alignments and expand the intersection with the most reliable alignments from the union result in alignments of a better quality \citep{Koehn:2003:SPT:1073445.1073462}. One of the commonly used heuristics is called \emph{grow-diag-final-and} \citep{koehn05iwslt}. Starting from the intersection of the two alignments, the heuristic adds neighboring alignment points that are part of the union but not the intersection. 

The quality of an automatic word alignment tool is commonly measured by comparing its output on a test set with a ground truth alignment done by human experts~\citep{DBLP:conf/acl/OchN00, och2003systematic,fraser-marcu-2007-squibs}. Human experts have to follow some guidelines to solve the ambiguities that come up during the annotation process. One of these guidelines is to use \emph{sure alignment points} and \emph{possible alignment points}~\citep{DBLP:conf/acl/OchN00, 10.5555/1734086}. Possible alignments are alignment points where human annotators are not in agreement. For example, function words that do not have a clear equivalent in the other language or two equivalent idiomatic expressions that are not a word-level translation of one another are aligned using \emph{possible alignments}. A commonly used measure of alignment quality is alignment error rate (AER). AER is computed using the following formula:
\begin{equation}
\text{AER}(S,P;A)= 1 -\frac{\left|A\cap S\right| + \left|A \cap P\right|}{\left|A\right| + \left|S\right|}.
\end{equation}
\noindent Here, $A$ is the set of alignment points that is being evaluated. $S$ and $P$ are the sets of sure alignment points and possible alignment points from a gold standard alignment, respectively.

GIZA++~\citep{och2003systematic} and fast\_align~\citep{dyer-chahuneau-smith:2013:NAACL-HLT} are two commonly used tools to compute automatic word alignments. Throughout this thesis, we use GIZA++, which produces alignments for each direction using the IBM models and symmetrizes the alignments by performing grow-diag-final-and heuristic.

Next, we discuss the most basic model of a phrase-based machine translation system. We explain how the model works and how it is created using the output of a word alignment system.

\subsection{Translation Models}
\label{subsec:trans_model}

The translation model can be considered the core model of a phrase-based machine translation system. In a word-based model the translation units are pairs of words, but in phrase-based models each item of the translation model or phrase table can be either a pair of words or a pair of phrases in the source and target languages.
 
Each item in a phrase table has four corresponding scores. Two of these are conditional phrase probability scores and the other two are lexical weightings. Let $\bar{f}$ and $\bar{e}$ stand for a source and a target phrase, respectively, then the phrase translation probabilities are $p(\bar{e}\mid\bar{f})$ and $p(\bar{f}\mid\bar{e})$. Phrase translation probabilities are computed using the counts of the phrase-pairs in the bilingual training corpus:
\begin{equation}
p(\bar{f}\mid\bar{e}) = \frac{C(\bar{e},\bar{f})}{\sum\nolimits_{\bar{f_{i}}}{C(\bar{e},\bar{f_{i}})}}.
\end{equation}
\noindent Here, $C(\bar{e},\bar{f})$ is the number of times that phrases $\bar{e}$ and $\bar{f}$ are extracted together as a phrase-pair from the bilingual training corpus. 

The noisy channel model~\citep{shanon1056798} forms the basis for phrase-based machine translation:
\begin{equation}
\begin{aligned}
\argmax_{e}\ p(e\mid f) &= \argmax_{e}\ \frac{p(f\mid e)p(e)}{p(f)}\\
& =\argmax_{e}\ p(f\mid e)p(e).
\end{aligned}
\end{equation}
\noindent Here, $f$ is source sentence, $e$ is target sentence and $p(f\mid e)$ is the likelihood that $f$ is a translation of $e$. In phrase-based machine translation, the likelihood $p(f\mid e)$ is estimated by breaking $f$ and $e$ into phrases and using the phrase translation probabilities.

Based on the noisy channel definition of translation probability estimation, the probability of source phrases conditioned on target phrases, $p(\bar{f}\mid\bar{e})$, are needed for translation probability estimation. However, if a target phrase is infrequent then the probability will not be a reliable estimate. To improve estimation in such a situations, it can be helpful to use the inverse conditional probability as well~\citep{koehn05iwslt}.

As mentioned before, the other two scores of each item in a phrase table are the lexical weighting scores. These scores are meant to help estimate phrase translation probabilities if a phrase-pair is infrequent in the training corpus. The lexical weightings are also computed in two directions as the conditional phrase probabilities. Lexical weighting is basically a back-off smoothing method to improve phrase probability estimations. In this case, we back off to the probabilities of the words in a phrase, since they are more frequent than the phrase itself and the probability estimates are more reliable. Given the alignment of the words in a phrase-pair, we compute the lexical weighting of a phrase-pair as follows:
\begin{equation}
p_l(\bar{e}\mid\bar{f}, a) = \prod_{i=1}^{\length(\bar{e})}{\frac{1}{\left| \{j\mid(i,j) \in a \}\right|}\sum_{\forall (i,j) \in a}{w(e_i\mid f_j)}},
\end{equation}
where $a$ refers to the alignment and $w(e_i\mid f_j)$ is the probability of generating word $e_i$ given its aligned word $f_j$ on the source side \citep{Koehn:2003:SPT:1073445.1073462}.

\subsubsection{Phrase Extraction}

Next, we describe the standard phrase extraction algorithm~\citep{och-etal-1999-improved, Koehn:2003:SPT:1073445.1073462}, which we use in Chapter~\ref{chapter:research-01}. As mentioned in Section~\ref{subsec:word_alignment}, phrases in a phrase table are extracted from word alignments and have to be consistent with the word alignment. Consistency with the alignment is ensured by following three conditions:

\begin{enumerate}
\item each phrase-pair includes at least one alignment link;
\item phrases are continuous sequences of words without any gaps; and
\item no word in any of the phrases of a phrase-pair has alignment links to words outside of the phrase-pair. 
\end{enumerate}

\noindent The phrase extraction algorithm loops over target side sequences and matches all source phrases that satisfy the consistency restrictions. If the matched source phrase has unaligned words at its borders, it can be expanded by those unaligned words and adds the new phrase-pairs to the extracted phrase-pairs set. There is a limit to the length of extracted phrase-pairs. This limit is typically set to 7 for each of the phrases in a phrase-pair~\citep{Koehn:2003:SPT:1073445.1073462}. Therefore, the phrase extraction algorithm continues to expand a phrase-pair until either the consistency condition is violated or the length of one of the source or target phrases exceeds the limit.

\begin{figure}[thb]
\centering
\includegraphics[scale=1, width=1\textwidth]{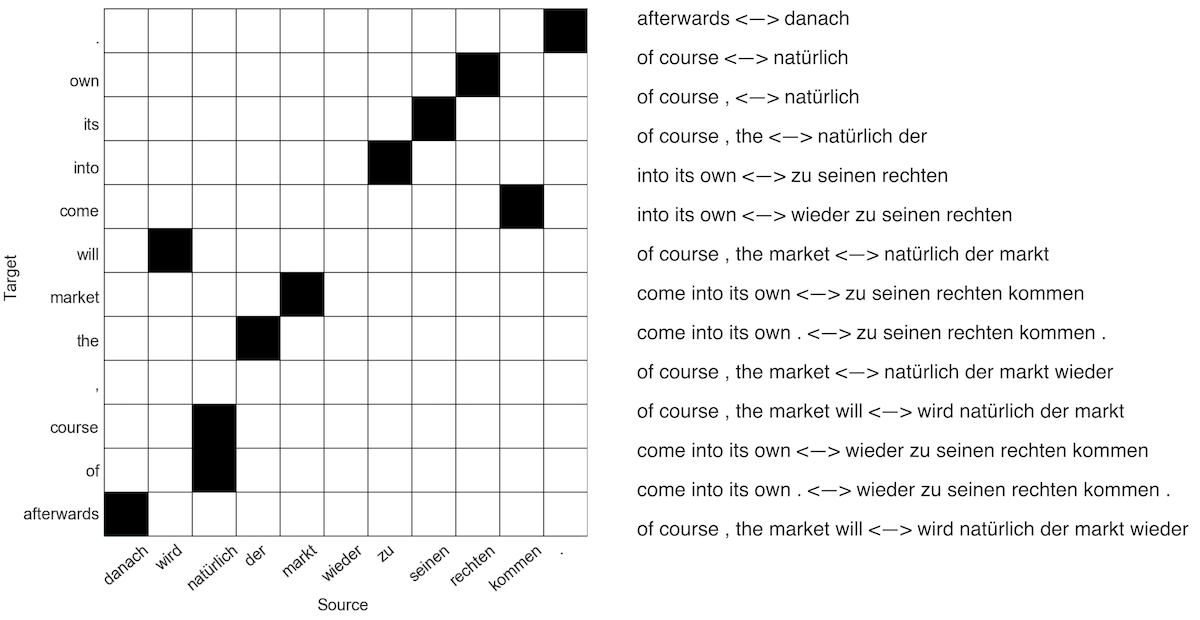}
\caption{Extracted phrase-pairs by the phrase-pair extraction algorithm from the alignment in Figure~\ref{fig:alignment_matrix}.}
\label{fig:phrase_pairs}
\end{figure}

Figure~\ref{fig:phrase_pairs} shows extracted phrase-pairs by the phrase extraction algorithm from the alignment given in Figure~\ref{fig:alignment_matrix}.

\subsection{Log-Linear Model and Parameter Estimation}
The feature functions of a phrase-based machine translation are combined in a log-linear model with each feature function having its own weight~\citep{och-ney-2002-discriminative}. The formal definition of the model is as follows:
\begin{equation}
p(e \mid a, f) = \exp \sum_{i=1}^{n}\lambda_{i}h_i(e, f, a), 
\end{equation} 
\noindent where $e$ is a target sentence, $f$ is a source sentence and $a$ is a latent variable representing the phrasal alignment between $f$ and $e$; $h_i(e, f, a)$ are the feature functions and $\lambda_i$ are their corresponding weights in the log-linear model. The feature functions are the log-values of the features including phrase probability, inverse phrase probability, and lexical weightings, as introduced in Section~\ref{subsec:trans_model}, as well as the language model and reordering model probabilities that will be discussed in Sections~\ref{subsec:n-gramLM} and \ref{subsec:Reomodel}, respectively.

The optimal values for the weight parameters of the log-linear model, i.e., the values for $\lambda_i$, are learned on a held-out data set called development set. The process of finding these optimal values is called ``tuning." The development set should be close to the target translation task in terms of vocabulary and the domain of the data. The optimal values are learned using a parameter optimization algorithm such as minimum error rate training (MERT)~\citep{och-2003-minimum} or pairwise ranking optimization (PRO)~\citep{hopkins2011tuning}. MERT performs a grid-based optimization by generating translations of the development set for different values of the parameters of the model and compares the generated translations with the corresponding reference translations. Both MERT and PRO can optimize any evaluation metric that provides a comparison of generated translation and reference translation. PRO converts the parameter optimization problem into a binary classification problem by using a pairwise ranking of translation hypotheses. It samples the parameter values from the search space and finds the best value vector by pairwise ranking. PRO can optimize systems with a large number of features whereas MERT is not scalable to handle more than a dozen parameters. For tuning the phrase-based model we use PRO, see Section~\ref{sec:ch2eval}.


\subsection{N-Gram Language Models}
\label{subsec:n-gramLM}

To ensure the fluency of the generated translations, phrase-based machine translation leverages n-gram language models~\citep{10.1162/0891201042544884}. An n-gram language model computes the probability of the next word given the history of the previously generated words. However, the length of the history is limited by $n$ which is the order of the language model. The order of the language model needs to be large enough to capture the specificity of different sequences and small enough that the learned distributions are not too sparse. An n-gram language model is trained on a large monolingual corpus in the target language. The language model is trained using the relative frequency of sequences of words with length of $n$. For example, for a language model of order 3, this is defined as:
\begin{equation}
P(w_3 \mid w_1, w_2) = \frac{C(w_1 w_2 w_3) }{ \sum\nolimits_{w_i} {C(w_1 w_2 w_i)}},
\end{equation}
where $C(w_1 w_2 w_3)$ is the count of string $``w_1 w_2 w_3"$ in the training data. To compute the probability score of a sentence, the probabilities of all words in the sentence given their $n-1$ words histories are multiplied:
\begin{equation}
p(s) = \prod_{i=1}^{\left|s\right|}{p(w_i\mid w_{i-n+1}^{i-1})}.
\end{equation}
\noindent It may happen that while scoring a sentence, sequences of words occur that have never been encountered during training. A maximum likelihood n-gram language model would assign a zero probability score to such a sequence, and hence a zero probability to the full sentence. To avoid this, the language model needs to be smoothed. There are various smoothing methods including Kneser-Ney~\citep{conficasspKneserN95,chen1999empirical}, Add-k smoothing~\citep{GaleChurch:94}, and Backoff and Interpolation \citep{JelMer80}. We briefly discuss the Backoff and Interpolation method, since we refer back to it in Chapter~\ref{chapter:research-01}. 
In the back-off method, we back off to a lower order n-gram ($n-1$) whenever the count for the n-gram in the training data is zero. We continue backing off until we reach a shorter history that has a non-zero count. The formal definition of the back-off method is:
\begin{equation}
P_b(w_i\mid w_{i-n+1}^{i-1}) =
\begin{cases}
\hat{P}(w_i\mid w_{i-n+1}^{i-1}), & \mathrm{if} \  C(w_{i-n+1}^{i}) > 0\\
\alpha_{w_{i-n+1}^{i-1}} P_b(w_i\mid w_{i-n+2}^{i-1}) & \text{otherwise}.\\
\end{cases}
\end{equation}
\noindent Here, $w_{i-n+1}^{i-1}$ is the n-gram history sequence, $C(w_{i-n+1}^{i})$ is the count of n-gram in training data, $\hat{P}(w_i\mid w_{i-n+1}^{i-1})$ is the discounted probability distribution which leaves some probability mass for unseen n-gram sequences and $\alpha_{w_{i-n+1}^{i-1}}$ is the back-off weight which specifies how much of the left-over probability mass is assigned to the current unseen sequence. The discounted probability distribution $\hat{P}(w_i\mid w_{i-n+1}^{i-1})$ and the back-off weight $\alpha_{w_{i-n+1}^{i-1}}$ can be computed by the Good-Turing method~\citep{katz1987estimation}.

Another method to use lower order n-gram probabilities to estimate the probability of unseen n-grams is interpolation. Here, the estimated probability for an n-gram sequence is a linear interpolation of the maximum likelihood probabilities for the n-gram sequences with varying values for $n$. For example, for a 3-gram sequence, the estimated probability is computed as follows:
\begin{equation}
\hat{P}(w_n \mid w_{n-2}^{n-1}) = \lambda_{1} P(w_n\mid w_{n-2}^{n-1}) + \\ \lambda_{2} P(w_n\mid w_{n-1}) + \lambda_{3} P(w_n),
\end{equation}
where the $\lambda$s sum to 1. Here, $\lambda_1$, $\lambda_2$ and $\lambda_3$ are weights that can be learned from data using expectation maximization~\citep{DEMP1977,Federico1996BayesianEM}. As can be seen, in this method, we include shorter histories, which are more frequent, to obtain more reliable estimates of the probability of the n-gram.  


\subsection{Reordering Models}
\label{subsec:Reomodel}

Different languages have different word orderings. To translate from one language to an other by using phrases as the unit of translation, a model needs to select source phrases for translation in the order that is natural in the target language. Multiple reordering models have been proposed for phrase-based machine translation. We describe three that are used in Chapter~\ref{chapter:research-01}.

\subsubsection{Linear Reordering}

The simplest reordering model for phrase-based machine translation is the linear reordering model or linear distortion model~\citep{Koehn:2003:SPT:1073445.1073462}. The linear reordering model is a simple function that penalizes long jumps in the source side while translating. The penalty depends on the distance between the last translated phrase and the current one. The distance is computed by using the position of the last word of the previously translated phrase and the first word of the current phrase. This function basically encourages translating the source phrases in the same order as they appear on the source side assuming that monotone translation is the best strategy in most cases. The formal definition of linear distortion is:

\begin{equation}
d(x) = \alpha^{\left| start_i - end_{i-1} -1\right|},
\end{equation} 
which is an exponential decay function with $0<\alpha<1$. The only parameter controlling the reordering cost in linear reordering is the distance and therefore does not depend on any lexical information. In this case, the language model controls the lexical aspect of the reordering to some extent by computing the fluency score of the resulting translations based on different reorderings.


 
\subsubsection{Lexicalized Reordering}
\label{subsub:LRM}

As described above, the linear distortion model does not leverage any lexical information to control reordering of the phrases. However, there are many situations where the reordering of a phrase has some relation with the phrase and its context (neighboring phrase). The lexicalized reordering model (LRM)~\citep{Tillmann:2004:UOM:1613984.1614010,koehn05iwslt} addresses these kinds of situation by conditioning the reordering movements on the lexical form of a phrase and its neighbors.

Since many phrase-pairs are not frequent enough to observe all possible reorderings during translation, the lexicalized reordering model defines a limited number of possible reordering moves to compensate for this sparsity. These reordering moves are referred to as ``orientations." The original lexicalized reordering model originally includes three orientations: (i) monotone (M), (ii) swap with previous phrase (S), and (iii) discontinuous jumps (D). Some variants of the lexicalized reordering model split discontinuous jumps into two types of orientations: discontinuous left (DL) and discontinuous right (DR), since it is more linguistically motivated \citep{Galley:2008:SEH:1613715.1613824}. In this thesis, we use the variant with four orientations (M, S, DL and DR).

In a lexicalized reordering model we use two probability distributions to decide about the reordering: (i) the probability of the orientations given the phrase-pair being translated and (ii) the probability of the orientations given the last translated phrase-pair. The probability of the orientations given the phrase-pair being translated is called \emph{the reordering probability with respect to the left of a phrase-pair}. Similarly, the probability of the orientations given the last translated phrase-pair is called \emph{the reordering probability with respect to the right of a phrase-pair}.

Lexicalized reordering probability distributions are learned from word aligned data. To this end, we count the number of times each phrase-pair appears with different orientations in the word-aligned training data. Then, the probability distribution is estimated using the relative frequency of the counts: 
\begin{equation}
P(o \mid \bar{f}, \bar{e}) = \frac{C(o, \bar{f}, \bar{e}) }{ \sum\nolimits_{o^\prime} {C(o^\prime, \bar{f}, \bar{e})}}.
\end{equation}
\noindent Here, $C(o, \bar{f}, \bar{e})$ refers to the number of times a phrase-pair co-occurs with orientation $o$.

\begin{figure}[thb]
\centering
\includegraphics[scale=0.5, width=0.75\textwidth]{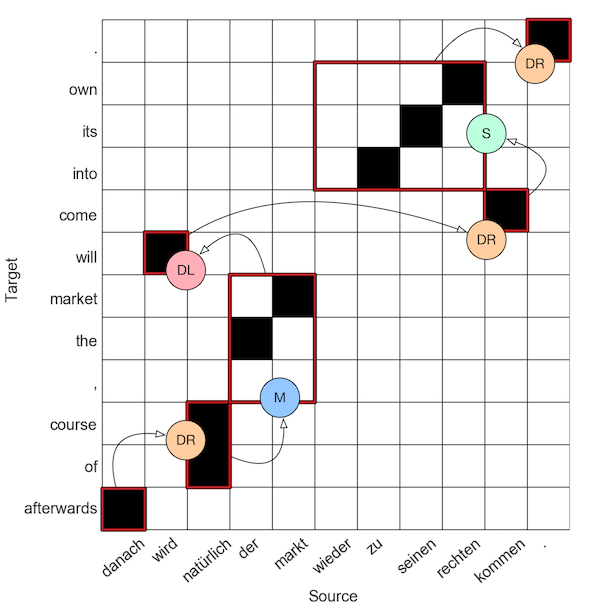}
\caption{An example of phrase-pairs with their corresponding reordering orientations for a given word alignment. The highlighted boxes are the phrase-pairs that are used for translation. Note that other sequences of phrase-pairs that are consistent with the word alignments could also be used.}
\label{fig:orientations}
\end{figure}

Figure~\ref{fig:orientations} shows an example of how orientations are identified. 


\subsubsection{Hierarchical Reordering}
\label{subsub:HRM}

The hierarchical reordering model~\citep{Galley:2008:SEH:1613715.1613824} has the same orientations as a lexicalized reordering model. It also uses the conditional probability distributions of the orientations given phrase-pairs. However, the way the orientations are counted during training of the model and the way the right orientation is identified during translation are different. 

\begin{figure}[thb]
\centering
\includegraphics[scale=0.5, width=0.75\textwidth]{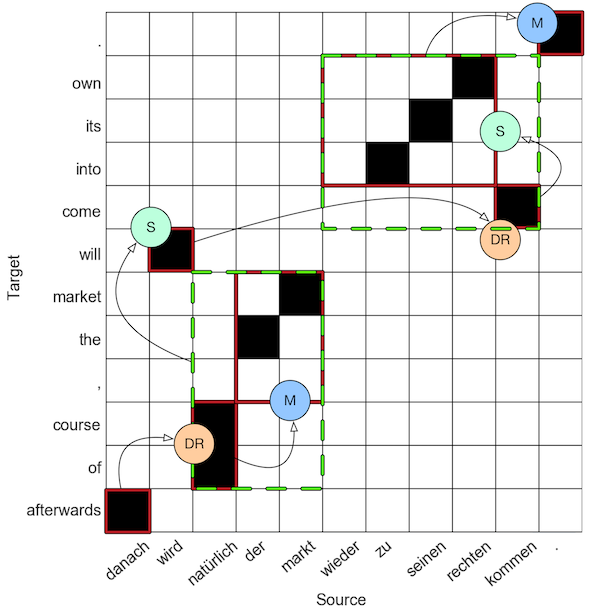}
\caption{The changes to the orientations of Figure~\ref{fig:orientations} if a hierarchical reordering model is used in place of the lexicalized reordering model. See the orientations at \emph{(wird, will)} and \emph{(., .)} phrase-pairs.}
\label{fig:hierarchical}
\end{figure}

During training, the hierarchical reordering model uses the longest possible phrase-pairs that the phrase extraction algorithm could have extracted if there was no limit for the phrase-pair length. 
This considerably reduces the number of discontinuous orientations and increases the number of swap and monotone orientations. During translation, the algorithm uses a shift-reduce parser to merge the currently translated phrase-pairs to the longest possible phrase-pairs and identifies the right orientation for the next phrase-pair with respect to the long phrase-pair formed in the neighborhood. The shift-reduce parser can be replaced by an approximation method that uses the source coverage vector to approximate the top block in the stack of the shift-reduce parser~\citep{cherry2012hierarchical}.

Figure~\ref{fig:hierarchical} shows how the orientations in Figure~\ref{fig:orientations} change if a hierarchical reordering model is used in place of a lexicalized reordering model. 

The dashed groupings are adjacent phrase-pairs grouped by the hierarchical reordering model. We have not shown all groupings that a hierarchical reordering model would create in this example, but those that result in a change in the orientations shown in Figure~\ref{fig:orientations}. 

For almost more than a decade phrase-based machine translation was the stat-of-the-art model. However, it was suffering from some of the problems including but not limited to reordering problems~\citep{cherry2012hierarchical,cherry:2013:NAACL-HLT, wuebker2013improving, chahuneau2013translating, li2014neural}. Neural machine translation showed great potential in the early stages by improving different aspects including reorderings~\citep{bentivogli-etal-2016-neural}. Next, we describe neural machine translation as we use it in Chapters~\ref{chapter:research-02} and \ref{chapter:research-03}.

\section{Neural Machine Translation}

\citet{devlin2014fast} showed that a feed-forward neural language 
model~\citep{bengio2003neural} with its conditioning history expanded to the source phrase in a machine task can outperform strong phrase-based machine translation models by a large margin.
This showed the great potential of neural language models for machine translation with additional research adapting neural language models to end-to-end machine translation~\citep{kalchbrenner-blunsom-2013-recurrent-continuous,sutskever2014sequence,bahdanau-EtAl:2015:ICLR,cho-etal-2014-learning,luong-EtAl:2015:ACL-IJCNLP, jean-EtAl:2015:ACL-IJCNLP, wu2016google, NIPS2017_7181}.

In this section, we describe neural machine translation and two commonly used model architectures.

\subsection{Neural Language Models}

Like n-gram language models, neural language models also compute the conditional probability of the next word, given a sequence of previous words. However, neural language models are more powerful when it comes to computing probabilities for unseen sequences. Therefore, there is no need for smoothing in neural language models as it is the case for n-gram language models.

\subsubsection{Feed-Forward Language Models}

\citet{bengio2003neural} made it possible to effectively learn the conditional probability of the next word given a limited history of the previous words by introducing a distributed representation of words. This distributed representation which is nowadays called \emph{word embedding} was the key concept for neural networks to open their way into language modeling and machine translation \citep{conf/interspeech/MikolovKBCK10, zou-etal-2013-bilingual, cho-etal-2014-properties, cho-etal-2014-learning, vaswani-etal-2013-decoding}. 

In the feed-forward neural language model by~\citet{bengio2003neural}, the history of the n-gram is given as the input and the probability score of each of the words from a predefined vocabulary appearing as the next word is returned as the output. The input words are encoded in one-hot vectors and are fed to a dictionary-like lookup layer that returns a high dimensional real-number vector representation for each word. The same high dimensional vector is returned for a word independent of the position of the word in a sentence:
\begin{equation}
C(w_i) = C w_i.
\end{equation}
\noindent Here, $C$ is the word embeddings weight matrix and $C(w_i)$ is the word embedding of word $w$ which is shown by its index $w_i$.

The output of this layer is then fed to a hidden layer with a non-linear activation function such as the hyperbolic tangent or $\tanh$. 
\begin{equation}
h = \tanh(b_h + \sum_{i}H_i C(w_i)).
\end{equation}
\noindent Here, $H_i$ are the weights of the $i$-th unit in the hidden layer and $b_h$ is the bias of the hidden layer.

Finally, the output of the hidden layer is fed to a softmax function to ensure the characteristics of a probability distribution:
\begin{equation}
\begin{aligned}
s &= W h, \\
p_i &= \softmax(s_i, s) = \frac{e^{s_i}}{\sum_{j}{e^{s_j}}}.
\end{aligned}
\end{equation}
\noindent Here, $W$ is the weights of the output embedding layer. In the final layer, a softmax function converts the output to a probability distribution over the vocabulary. This formulation of the feed-forward language model is based on \citep{koehn2017neural} and eliminates the skip layer connections in the version by \citet{bengio2003neural}.

\subsubsection{Recurrent Neural Language Models}

The history that the feed-forward language model can capture is limited in the same way as n-gram language models. To expand this to longer histories, allowing one to capture longer dependencies, recurrent neural language models were introduced \citep{conf/interspeech/MikolovKBCK10}. The basic idea in these models is to feed back the state of the hidden layer in the previous time step to the current time step to keep track of the previous decisions of the network. The hidden layer plays the role of a representation for the whole history observed up to a given point in time. Since the history can become very long and the effect of earlier observations by the network vanishes in the hidden layer representation, long short-term memory models (LSTM)~\citep{Hochreiter:1997} are used for modeling sentences~\citep{Sundermeyer2012LSTMNN}. LSTM units have a memory state allowing them to store information for longer periods of time. Additionally, the gates of LSTM units make it possible to control the information in the memory state to be fully or partially changed at each time step. These properties of an LSTM unit help recurrent neural networks with LSTM units avoid the problem of vanishing gradients for longer histories~\citep{DBLP:journals/corr/LeJH15, DBLP:journals/corr/abs-1801-01078}.

\begin{figure}[thb!]
\centering
\includegraphics[scale=0.2, width=0.52\textwidth]{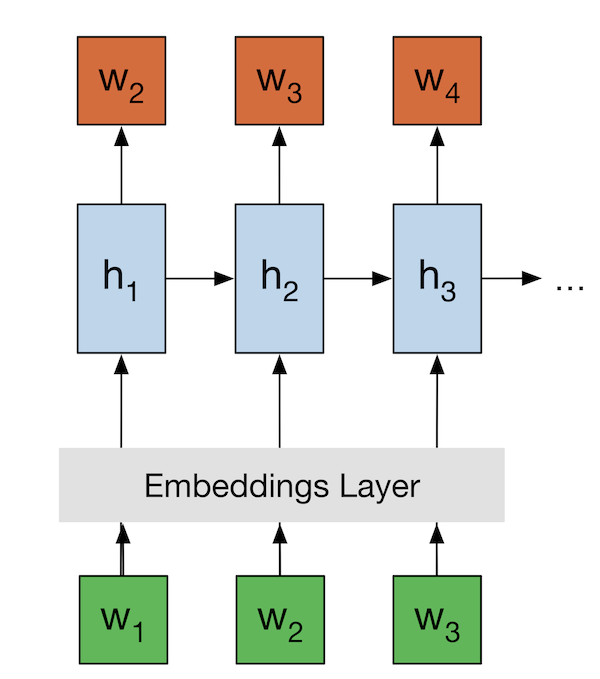}
\caption{The flow of information in a recurrent neural language model. The model reads a word from the input sequence at each time step and predicts the next word. The predicted word is not necessarily the same as the next word in the input sequence. This is reflected by the difference in the colors.}
\label{fig:rnnlm}
\end{figure}

Figure~\ref{fig:rnnlm} shows the general flow of information in a recurrent neural language model. The model digests one word at a time from the sequence and outputs the probability score for each word of a predefined vocabulary to appear as the next word of the sequence given the entire history the network has observed so far. The highest scoring word is the predicted next word.

\subsection{Word Embeddings}
 
Word embeddings are high dimensional vector representations of words based on the context in which they appear. The capability of representing words in their context by high dimensional vectors of real numbers is a key concept to the success of the neural networks in language modeling, machine translation and other NLP tasks~\citep{bengio2003neural, mikolov-etal-2013-linguistic, pennington2014glove, zou-etal-2013-bilingual}. The core idea behind word embeddings is that words that occur in similar contexts are semantically similar and therefore should have similar representations~\citep{firth1957synopsis, doi:10.1080/01690969108406936}. The word embedding layer in neural networks is typically a look-up table matrix of weights that is learned during training of the neural network for a language generation task (typically language modeling). An interesting and useful characteristic of word embeddings is that the similarity between words can be quantified by cosine distance (or similarity) between their respective embedding representations. Word embeddings also offer the possibility of adding and subtracting word vectors to achieve some sort of semantic inference~\citep{mikolov-etal-2013-linguistic}.


Word embeddings capture a general representation of all senses of a multi-sense word. Hence, the most frequent sense of a word in the training data is the dominant sense captured by an embedding representation~\citep{faruqui-etal-2016-problems, fadaee-etal-2017-learning}.

In neural machine translation, it is common to learn the word embedding during training of the machine translation model~\citep{qi-etal-2018-pre,bahdanau-EtAl:2015:ICLR,luong-EtAl:2015:ACL-IJCNLP,jean-EtAl:2015:ACL-IJCNLP,wu2016google,NIPS2017_7181}, in contrast to most other language-specific tasks that often use pre-trained word embeddings~\citep{qi-etal-2018-pre, ma-hovy-2016-end, lample-etal-2016-neural}. Nevertheless, it is also possible to use pre-trained word embeddings in neural machine translation~\citep{qi-etal-2018-pre}.

\subsection{Recurrent Neural Translation Models}

With long dependencies and long reorderings being very difficult phenomena for phrase-based machine translation to capture~\citep{Galley:2008:SEH:1613715.1613824, green-etal-2010-improved, durrani-etal-2011-joint, bisazza2015survey}, recurrent neural machine translation appeared to be able to solve these shortcomings of phrase-based machine translation~\citep{bentivogli-etal-2016-neural,bahdanau-EtAl:2015:ICLR}. Recurrent neural translation models are end-to-end models that encode source sentences as high dimensional distributed representations and then decode the representations into a translation in the target language. 
Both the encoder and decoder can be LSTM~\citep{Hochreiter:1997} or GRU~\citep{cho-etal-2014-learning} networks.  

Recurrent neural translation models are extensions to the recurrent neural language models. In recurrent language models, we predict the next word given the words generated so far in a sequence. In recurrent neural translation model, we predict one word at a time, given the translated words generated so far and the high dimensional representation of the source sentence.

Formally, the source sentence can be seen as a sequence of embedding vectors:
\begin{equation}
S = (v_1, \ldots, v_{T_s}).
\end{equation}
\noindent The encoder reads the source sentence one vector at a time into a hidden recurrent layer:
\begin{equation}
h_t = f(v_t, h_{t-1}),
\end{equation}
where $h_t$ is the output of the hidden layer at time $t$ and $f(v_t, h_{t-1})$ can be an LSTM \citep{Hochreiter:1997} or GRU~\citep{cho-etal-2014-learning} recurrent layer applied to the previous hidden state and the current input word.

Then, the sequence of the hidden states generated by the observation of the input word at each time step is used to generate a hidden state representation of the input sentence which is referred to as context vector:
\begin{equation}
\label{eq:contextVec}
c = q({h_1, \ldots, h_{T_s}})
\end{equation}

In the simplest recurrent neural machine translation model $q({h_1, \ldots, h_{T_s}})$ returns the last hidden state of the encoder as the output. This function can be more complex including various types of attention functions, which are discussed in Section~\ref{subsec:attn}.

The decoder predicts the next translation word given the context vector $c=q({h_1, \ldots, h_{T_s}})$ and all predicted target words so far. 
\begin{align}
r_{t} &= f^{\prime}(r_{t-1}, y_{t-1}, c)\\
p(y_t \mid {y_1, \ldots, y_{t-1}}, c) &= g(y_{t-1}, r_{t-1}, c).
\end{align}

\noindent Here, $f^{\prime}$ can be an LSTM or GRU recurrent layer, $r_t$ is the hidden state of the decoder recurrent layer at time $t$, $g$ is a linear transformation followed by a softmax function, and ${y_1, \ldots, y_{t-1}}$ are the target words predicted so far.
 
Some variants of encoders use bidirectional recurrent layers to generate the encoder hidden states at each time step~\citep{bahdanau-EtAl:2015:ICLR}. In these models, one recurrent layer encodes the source sentence from left to right and the other layer encodes it from right to left. Then, the output hidden states of each layer at time step $t$ are concatenated to form the output hidden state of the encoder at time step $t$. The bidirectional encoding causes the model to learn more temporal dependencies in the source sequence and has been shown to achieve performance improvements~\citep{zhou-etal-2016-deep}.
 
\subsection{Attention Models}
\label{subsec:attn}

In the case of long source sentences, the capacity of a single encoder hidden state is not enough to capture all information from the source side that is needed for the decoder to generate a correct translation~\citep{bahdanau-EtAl:2015:ICLR}. \citet{cho-etal-2014-properties} show that the performance of the vanilla recurrent neural model indeed deteriorates for long sentences as a result of the limited capacity of the output representation of the encoder. To address this problem, the context vector $c$ in Equation~\ref{eq:contextVec} is computed in a more complex way that learns to choose the required information from the source side needed for target word generation at each time step. 

These so-called attention models define the context vector $c$ to be a weighted average of encoder hidden states from all time steps:
\begin{equation}
c_i = \sum_{i=1}^{|S|}\alpha_{t,i}h_i.
\end{equation}

\noindent The weights $\alpha_{t,i}$ are computed by a softmax over the similarity of hidden states from the encoder and the current hidden state of the decoder:
\begin{align}
e_{t,i} &= h^{T}_{i} r_{t} \label{eq:hiddenStateComparison}\\
\alpha_{t,i} &= \frac{\mathrm{exp}(e_{t,i})}{\sum_{j=1}^{|S|} \mathrm{exp}(e_{t,j})} \label{eq:attentionWeight}.
\end{align}
\noindent Instead of using a single vector to generate the entire target translation as is the case in earlier neural machine translation models~\citep{sutskever2014sequence,kalchbrenner-blunsom-2013-recurrent-continuous, cho-etal-2014-properties}, the context vector changes at each time step during the generation of the target sentence and assigns more weight to the relevant parts of the source sentence.

There are more complex types of attention that keep track of the amount of attention payed to different parts of the source side at previous time steps~\citep{bahdanau-EtAl:2015:ICLR, DBLP_journals_corr_LuongPM15, feng-etal-2016-improving}. We introduce one of these attention models in more detail in Section~\ref{sec:attentionModels}.

\subsection{Transformer Model}

\citet{NIPS2017_7181} showed that attention can be used to replace the capabilities that recurrent neural networks provide for machine translation. Transformers are also encoder-decoder models in which both the encoder and decoder use fully attentional networks. 

Attention is a core component in the transformer architecture. Here, we introduce the notation used to describe attention in the transformer model. We use general notation that can describe all different attentions in a transformer architecture.

In the transformer model, each word embedding is mapped to three types of vector: query, key and value, using three different linear transformations for which the weights are learned during training.  

In the attention model, query vectors are compared to key vectors by using the dot product. The result of the dot product is scaled by the square root of the dimension size of the vectors. The result of the scaled dot product of a query and the keys is then passed through a softmax function, and the result is used as weights for the value vectors to form the output of the attention function. This is shown by the following formula:
\begin{equation}
A(Q, K, V) = \softmax\left(\frac{QK^T}{\sqrt{d_k}}\right) V.
\end{equation} 
\noindent Here, $Q$ stands for queries, $K$ stands for keys, $V$ stands for values and $d_k$ is the size of query and key vectors.

The difference between the attention described here and the attention model described in Section~\ref{subsec:attn} is the scaling factor and the linear mappings to query, key and value vectors. Otherwise, the attention model used in recurrent neural machine translation can also be described by using the current general notation.

\subsubsection{Multi-Head Attention}

Attention is the only mechanism that gathers all information from the source sentence as well as the prefix of the target sequence in the transformer architecture. This is in contrast to the usage of recurrence and attention in recurrent neural models. \citet{NIPS2017_7181} also introduce multi-head attention which allows attention to attend to information from different representation subspaces. The difference in the functionality of each head is realized by different linear projections for each head. 

\begin{figure}[thb]
\centering
\includegraphics[scale=0.2, width=0.75\textwidth]{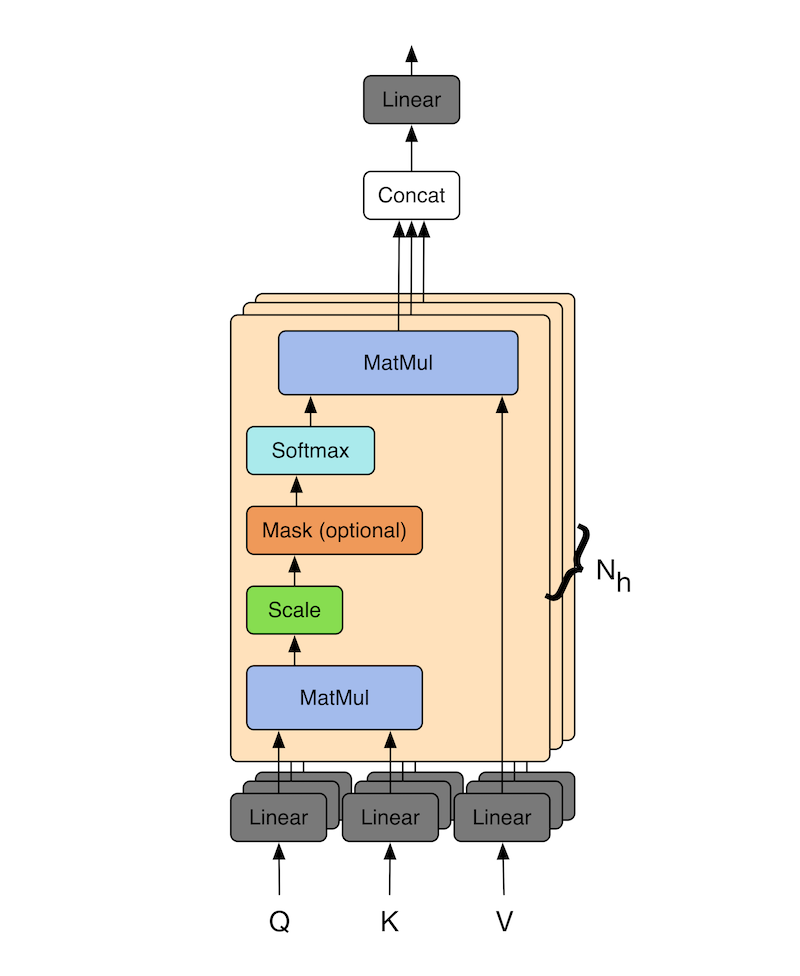}
\caption{Multi-head attention.}
\label{fig:multiAttn}
\end{figure}

Figure~\ref{fig:multiAttn} shows how multi-head attention works. Typically the number of heads being used is set to 8 following~\citet{NIPS2017_7181}. The output vectors of each head are concatenated and passed through another linear transformation to form the ultimate output vector of the attention layer. The formal definition of multi-head attention is as follows:
\begin{equation}
A_m(Q, K, V) = \concat(h_1, \ldots, h_n)W^o\ \text{where} \ h_i = A(QW_{i}^{Q}, KW_{i}^{K}, VW_{i}^{V}).
\end{equation}
\noindent Here, $W_{i}^{Q}$, $W_{i}^{K}$ and $W_{i}^{V}$ are the weights for linear transformation of queries, keys and values for $i$th attention head, respectively. $W^o$ are the weights of the output linear transformation.

\begin{figure}[thb]
\centering
\includegraphics[scale=0.2, width=0.62\textwidth]{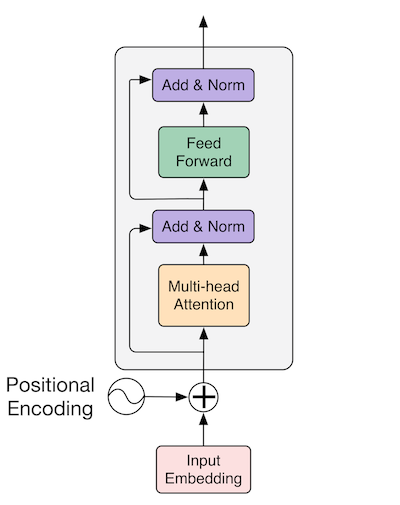}
\caption{Encoder of a transformer.}
\label{fig:encoder}
\end{figure}

\subsubsection{Encoder}

The encoder in a transformer architecture is composed of stacked network layers each of which contains a multi-head self-attention~\citep{parikh-etal-2016-decomposable} and a feed-forward sub-layer preceded by layer normalization~\citep{DBLP:journals/corr/BaKH16}. Figure~\ref{fig:encoder} shows the architecture of one of these layers. This layer is stacked $n$ times to form the encoder. 

Words from the input sequence are looked up in an embedding layer just as in a feed-forward or recurrent language model. Since the input is read one word at a time and there is no recurrence to encode and transfer the order of information, we need some type of position information for each input word. To this end, a positional encoding~\citep{CNN_positional_enc} is added to each word embedding. A positional encoding is a vector of the same dimensionality as the word embeddings to allow for vector addition. Positional encodings can be embeddings that are learned during training just as the word embeddings, or it can be a fixed vector. 



Word embeddings together with the positional encodings go into a multi-head self-attention layer. In this self-attention layer all queries, keys and values are computed from the word embeddings and positional encodings. This layer learns to attend to different parts of the source sentence depending on which source word is considered as the query. Then, the output of self-attention added with the positional encoded word embedding goes through a layer normalization component. The positional encoded word embeddings are connected by high-way or residual connections~\citep{He2016DeepRL} to the output of the self-attention layer, see Figure~\ref{fig:encoder}. The result is passed through a two-layer simple feedforward network with the following formal definition:
\begin{equation}
\label{eq:LRU}
f(x) = \max(0, xW_1 + b_1)W_2 + b_2.
\end{equation}
\noindent A ReLU activation function, the $\max(0, y)$ function in Equation~\ref{eq:LRU}, is used between the two linear transformation layers of the feed-forward layer. $W_1$ and $W_2$ are the weights of the linear layers and $b_1$, $b_2$ are the respective biases.

The size of the first layer of the feed-forward network is commonly set to 4 times the model size while the input and the output size are equal to the model size \citep{NIPS2017_7181}. For the feed-forward layer also the output and the input are added and the result goes through layer normalization. The process of going through self-attention and then a feed-forward network is repeated $n$ times using identical architectures but different weights.

\begin{figure}[thb!]
\centering
\includegraphics[scale=0.2, width=0.90\textwidth]{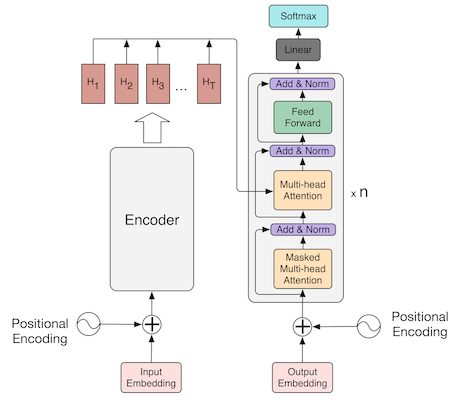}
\caption{Decoder of a transformer as part of the whole architecture.}
\label{fig:decoder}
\end{figure}

\subsubsection{Decoder}

The decoder in the transformer architecture has three main sub-layers: a multi-head self-attention, a multi-head attention which attends to hidden state outputs of the encoder, and a feed-forward network similar to the one in the encoder.

Figure~\ref{fig:decoder} shows the decoder as part of the full architecture of the transformer model. As can be seen, the architecture of a decoder layer is very similar to the encoder layer, see Figure~\ref{fig:encoder}, with an extra multi-head attention sub-layer which attends to the encoder outputs.

The first multi-head attention sub-layer in the decoder is masked for target words that occur after the current position in the target sentence. This masking is required to make sure that the model learns only to depend on the previously generated target words when predicting the next word.

The second multi-head attention performs a similar task to the attention layer in a recurrent neural machine translation model. This sub-layer compares the current state in the decoder, which is the output of the masked attention added with its input through a residual connection, to all the output hidden states from the encoder stack. Based on this comparison, it computes the attention weights. The attention weights are then used to compute a weighted average over the output hidden states of the encoder. 

The rest of the decoder layer is identical to an encoder layer. The ultimate output of the decoder stack is transformed linearly and goes through a softmax layer to be converted to a probability distribution over the entire target vocabulary.

 
\section{Evaluation}
\label{sec:ch2eval}

In this thesis, we use automated metrics to evaluate the translation quality of our systems. We use BLEU \citep{papineni2002bleu} as the most common evaluation metrics for all our systems. BLEU is a precision-based evaluation metric that matches n-grams in the translation output with (multiple) human reference translations. The BLEU metric computes precisions for different n-grams (usually up to 4-gram) and combines the precisions by multiplying them. Since precision encourages short translations, BLEU uses a \emph{brevity penalty} to compensate for this preference. 

We also use two other evaluation metrics that are more sensitive to reorderings in Chapter~\ref{chapter:research-01}. We use translation error rate (TER)~\citep{snover2006study}, which computes a word-based edit distance between a translation and corresponding reference translation inspired by the Levenshtein distance. TER has one more editing step in addition to the insertion, deletion and substitution editing steps from the Levenshtein distance. This extra edit is a block movement called jump. Using its list of editing steps, TER finds the shortest sequence of editing steps to convert a translation output into the corresponding human reference translation. The last evaluation metric we use is RIBES~\citep{isozaki2010automatic}. RIBES is a rank correlation based evaluation metric designed for translation tasks with language pairs that require long-distance reorderings. RIBES aligns words between the translation output and the human reference translation and computes the rank correlation of the aligned word positions. 

In the next chapter, we will discuss phrase reordering models and how different internal words of phrase-pairs can have a different impact on phrase reordering.

%
\graphicspath{ {04-research-01/images/} }

\chapter{The Impact of Internal Words on Phrase Reorderings}
\label{chapter:research-01}




\section{Introduction}

Structural differences between languages make the correct reordering of phrases an important and complex aspect of phrase-based machine translation systems~\citep{bisazza2015survey}. To this end, there is a lot of work on modeling reordering for phrase-based machine translation~\citep{Tillmann:2004:UOM:1613984.1614010, Galley:2008:SEH:1613715.1613824,  green-etal-2010-improved, durrani-etal-2011-joint, cherry2012hierarchical, cherry:2013:NAACL-HLT}.

The introduction of lexicalized reordering models (LRMs) \citep{Tillmann:2004:UOM:1613984.1614010} was a significant step towards better reordering, outperforming the distance-based reordering model~\citep{koehn05iwslt}, 
by modeling the orientation of the current phrase-pair with respect to the previously translated phrase; see Section~\ref{subsub:LRM} for a detailed introduction. LRMs score the order in which phrases are translated by using a distribution over orientations conditioned on phrase-pairs. Typically, the set of orientations consists of: \textit{monotone} (M), \textit{swap} (S) and \textit{discontinuous} (D), which is often split into \textit{discontinuous left} (DL) and \textit{discontinuous right} (DR). However, LRMs are limited to reorderings of neighboring phrases only. \citet{Galley:2008:SEH:1613715.1613824} proposed a hierarchical phrase reordering model (HRM); see Section~\ref{subsub:HRM}, for more global reorderings. HRMs handle long-distance reorderings better than LRMs by grouping adjacent phrase-pairs into longer phrase-pairs and determining the orientations with respect to these longer phrase-pairs.

LRMs and HRMs both use relative frequencies observed in a parallel corpus to estimate the distribution of orientations conditioned on phrase-pairs. As a result, both suffer estimating unreliable distributions for cases that rarely occur during training and therefore have to resort to smoothing methods to alleviate sparsity issues. 

\begin{table*}[t]
\centering
\small
\caption {\label{tbl:1} Examples of similar phrase-pairs and their orientation probabilities using Dirichlet  (Equation~\ref{emc:1}) and Recursive MAP (Equation~\ref{emc:2}) smoothing. M = monotone, S = swap, DL = discontinous left, and DR = discontinous right.}
\begin{tabular}{
cccccccc}
\cline{5-8}
 & & & \multicolumn{1}{c|}{} &\multicolumn{4}{c|}{Dirichlet Smoothed} \\
\cline{2-8}
 & \multicolumn{1}{|c}{Source} & \multicolumn{1}{|c}{Target} & \multicolumn{1}{|c|}{Freq} & M & S & DL & \multicolumn{1}{c|}{DR} \\
\cline{2-8}
a & \multicolumn{1}{|c}{\zh{中国 政府}} & \multicolumn{1}{|c}{chinese government} & \multicolumn{1}{|c|}{2834} & 0.216 & 0.034 & 0.315 & \multicolumn{1}{c|}{0.433}\\
b & \multicolumn{1}{|c}{\zh{日本 政府}} & \multicolumn{1}{|c}{japanese government} & \multicolumn{1}{|c|}{580} & 0.157 & 0.039 & 0.299 & \multicolumn{1}{c|}{0.503}\\
c & \multicolumn{1}{|c}{\zh{尼泊尔 政府}} & \multicolumn{1}{|c}{nepalese government} & \multicolumn{1}{|c|}{11} & 0.525 & 0.001 & 0.101 & \multicolumn{1}{c|}{0.370} \\
\cline{2-8}
& & & & & & & \\
\cline{5-8}
& & & \multicolumn{1}{c|}{} & \multicolumn{4}{c|}{Recursive Map Smoothed}\\
\cline{2-8}
 & \multicolumn{1}{|c}{Source} & \multicolumn{1}{|c}{Target} & \multicolumn{1}{|c|}{Freq} & M & S & DL & \multicolumn{1}{c|}{DR} \\
 \cline{2-8}
a & \multicolumn{1}{|c}{\zh{中国 政府}} & \multicolumn{1}{|c}{chinese government} & \multicolumn{1}{|c|}{2834} & 0.216 & 0.034 & 0.315 & \multicolumn{1}{c|}{0.432}\\
b & \multicolumn{1}{|c}{\zh{日本 政府}} & \multicolumn{1}{|c}{japanese government} & \multicolumn{1}{|c|}{580} & 0.158 & 0.039 & 0.300 & \multicolumn{1}{c|}{0.501}\\
c & \multicolumn{1}{|c}{\zh{尼泊尔 政府}} & \multicolumn{1}{|c}{nepalese government} & \multicolumn{1}{|c|}{11} & 0.400 & 0.009 & 0.202 & \multicolumn{1}{c|}{0.388}\\
\cline{2-8}
\end{tabular}
\end{table*}

\subsection{Problem Definition}
\label{sec:problem_def}

In order to smooth the original maximum likelihood estimation, LRMs originally back off to the general distribution over orientations:
\begin{equation}
\label{emc:1}
P(o \mid \bar{f}, \bar{e}) = \frac{C(o, \bar{f}, \bar{e})+ \sigma P(o) }{ \sum\nolimits_{o^\prime} {C(o^\prime, \bar{f}, \bar{e})} + \sigma},
\end{equation}
which is also known as Dirichlet smoothing, where $\sigma P(o)$ denotes the parameters of the Dirichlet prior that maximizes the likelihood of the observed data. $C(o, \bar{f}, \bar{e})$ refers to the number of times a phrase-pair co-occurs with orientation $o$, and $\sigma$ is the \textit{equivalent sample size}, i.e., the number of samples required from $ P(o) $ to reflect the observed data \citep{smucker2005investigation}. \citet{cherry:2013:NAACL-HLT} and \citet{chen-foster-kuhn:2013:NAACL-HLT} introduce recursive Maximum a Posteriori (MAP) smoothing, which makes use of more specific priors by recursively backing off to orientation priors, see Equation~\ref{emc:2}:
\begin{equation}
\label{emc:2}
\begin{split}
P(o \mid \bar{f}, \bar{e}) &= \frac{C(o, \bar{e}, \bar{f}) + \alpha_s P_s(o \mid \bar{f}) + \alpha_t P_t(o \mid \bar{e})}{ \sum\nolimits_{o^\prime} {C(o^\prime, \bar{e}, \bar{f})} + \alpha_s + \alpha_t } \\
P_s(o \mid \bar{f}) &= \frac{\sum_{\bar{e}}{C(o, \bar{f}, \bar{e}) + \alpha_g P_g(o)}}{\sum\nolimits_{o^\prime, \bar{e}}{C(o^\prime, \bar{f}, \bar{e})} + \alpha_g } \\
P_t(o \mid \bar{e}) &= \frac{\sum_{\bar{f}}{C(o, \bar{f}, \bar{e}) + \alpha_g P_g(o)}}{\sum\nolimits_{o^\prime, \bar{f}}{C(o^\prime, \bar{f}, \bar{e})} + \alpha_g } \\
P_g(o) &= \frac{\sum_{\bar{f}, \bar{e}}{C(o, \bar{f}, \bar{e})} + \alpha_u / 3}{\sum\nolimits_{o^\prime, \bar{f}, \bar{e}}{C(o^\prime, \bar{f}, \bar{e}) + \alpha_u}}
\end{split}
\end{equation}
\noindent While recursive MAP smoothing factorizes phrase-pairs into source and target phrases, it still considers the phrases themselves as fixed units. 

To better understand what kind of information is ignored by both of the aforementioned smoothing methods, consider the phrase-pairs and their corresponding orientations distributions given in Table~\ref{tbl:1}. The phrase-pairs in rows (a) and (b) are frequently observed during training, resulting in reliable estimates. On the other hand, the phrase-pair in row (c) is infrequent, leading to a very different distribution due to the smoothing prior, while being semantically and syntactically close to (a) and (b). In Table~\ref{tbl:1}, we can also see that recursive MAP smoothing results in slightly more similar distributions compared to plain Dirichlet smoothing but the overall differences remain noticeable. 

In this chapter, we argue that in order to obtain smoother reordering distributions for infrequent phrase-pairs such as the ones in Table~\ref{tbl:1}, one has to take phrase-internal information into account. 

\subsection{Research Questions}

In earlier work, \citet{cherry:2013:NAACL-HLT} builds on top of HRMs and proposes a sparse feature approach that uses word clusters instead of fully lexicalized forms for infrequent words to decrease the effect of sparsity on the estimated model. 

\citet{nagata2006clustered} and \citet{durrani2014investigating} have used part of speech (POS) tags and different types of word classes instead of the original forms of words to address the unreliable estimation of the reordering distribution for infrequent phrases. In particular, \citet{nagata2006clustered} have used the first and the last words of the phrases as the most influential words of the phrase-pairs and have introduced them as separate features to their model. However, they have not provided any argument or evidence for their choice.

In this chapter, we investigate the importance of the internal words in defining reordering behavior of the phrases. This helps to better understand the linguistic phenomena captured by phrase reordering models. Additionally, it also helps improve the estimation of the reordering distribution in case of infrequent phrase-pairs by backing off to the most influential words. We ask the following general research question:

\begin{itemize}[wide, labelwidth=!, labelindent=0pt ]
\item[] \textbf{\ref{rq:reorderingMain}} \textit{\acl{rq:reorderingMain}}
\end{itemize}

\noindent Examples like those in Table~\ref{tbl:1} already show that the ordering of phrase-pairs does not entirely depend on the complete lexical form of the phrase-pairs. Intuitively, those phrase-pairs should follow the same reordering distribution. To achieve this, one has to know to what extent it is required to use the full lexical form of a phrase-pair to be able to estimate the reordering distribution for it. This is important because if we could substitute the full form with a shorter or more abstract form of a phrase-pair to estimate its reordering distribution then the problem of estimating the reordering distribution for infrequent phrase-pairs can be alleviated by finding the right shortened or abstract form.

\ref{rq:reorderingMain} is subdivided into the following three subquestions:

\begin{itemize}[wide, labelwidth=!, labelindent=0pt ]
\item[] \textbf{\ref{rq:reorderingSub1}} \textit{\acl{rq:reorderingSub1}}
\end{itemize}

\noindent Understanding the impact of the internal structure of the phrase-pairs on their reordering behavior can help find more effective representations of phrases to improve the estimation of reordering distributions for infrequent phrase-pairs. We answer this question with two types of experiment. We use a back-off approach following the back-off idea in n-gram language modelling as one approach, and a generalized form approach that keeps \textit{exposed heads} lexicalized and generalizes the remaining words. Exposed heads are words inside a subsequence that are in a dependency relation with a word outside of the subsequence and are dominated by that word~\citep{li2012head}. Based on these experiments, we also answer \ref{rq:reorderingSub2} and \ref{rq:reorderingSub3}.

\begin{itemize}[wide, labelwidth=!, labelindent=0pt ]
\item[] \textbf{\ref{rq:reorderingSub2}} \textit{\acl{rq:reorderingSub2}}
\end{itemize}

\noindent The use of the first and the last word of the phrases as a separate feature of the reordering model, as proposed by \citet{nagata2006clustered} and \citet{cherry:2013:NAACL-HLT}, focuses on the border words of phrases as the most influential words on the ordering of phrases. We experiment with backing off towards the border words of phrases following the back-off idea in smoothing n-gram language models to validate whether the influence of words on reordering increases with their proximity to the borders:

\begin{itemize}[wide, labelwidth=!, labelindent=0pt ]
\item[] \textbf{\ref{rq:reorderingSub3}} \textit{\acl{rq:reorderingSub3}}
\end{itemize}

\noindent Some work has already used class-based generalized forms using POS tags and automatically learned word clusters to estimate the reordering distribution of infrequent phrase-pairs \citep{nagata2006clustered}. However, the authors have done this for all words in a phrase-pair without accounting for potentially different impacts of the internal words on the reordering behavior of the phrase-pair. Here, we experiment with generalizing the words that are less influential on reordering behavior of a phrase-pair and leave the more influential words in their original form to avoid over-generalization.

In this chapter, we propose two types of approach to use the most influential words from inside the original phrase-pairs to estimate better orientation distributions for infrequent phrase-pairs. These approaches perform better at taking phrase-pair similarities into account than the LRM and HRM which use the full lexical form of the phrase-pair. In the first approach, we define a back-off model to shorten towards 
important words inside the original phrase-pairs following the idea of back-off models in language model smoothing. This is to some extent complementary to the HRM by using smaller phrase-pairs to achieve better predictions. The difference is that within HRMs smaller phrase-pairs are merged into longer blocks when possible, while we propose to use shorter forms of phrase-pairs when possible. In the second approach, we propose to produce generalized forms of fully lexicalized phrase-pairs by including important words and marginalizing the remaining words allowing for smoothed distributions that better capture the true distributions over orientations. Here, we use syntactic dependencies from the original phrase-pair to generalize and shorten in a more linguistically informed way. 

This chapter makes the following contributions:

\begin{enumerate}
\item We introduce new methods that use shortened and generalized forms of a phrase-pair to smooth the original phrase orientation distributions. 
We show that our smoothing approaches result in improvements in a phrase-based machine translation system, even when compared against a strong baseline using both LRM and HRM together. These methods do not require any changes to the decoder and do not lead to any additional computational costs during decoding. 

\item As our second contribution, we provide a detailed analysis showing that orientation distributions conditioned on long phrase-pairs typically depend on a few words within phrase-pairs and not the whole lexicalized form. These words are syntactically important words based on the dependency parse of the sentences. Our analysis contributes to the interpretability of reordering models and shows that capturing syntax is an important aspect for better machine translation quality. In Chapter~\ref{chapter:research-03}, we investigate how syntactic information is captured by neural machine translation architectures which are currently the state-of-the-art models. Moreover, in Chapter~\ref{chapter:research-02}, we show how attention in recurrent neural machine translation model learns to attend to modifiers of a word, as defined by a dependency parse, while generating the translation of the word. Note that exposed heads used in the current chapter are modifiers that are modifying a word outside of their containing phrases.  Our analysis of reordering behavior, as described in this chapter, supports and adds to the sparse reordering features of \citet{cherry:2013:NAACL-HLT}.
\end{enumerate}

\section{Related Work}


The problem of data sparsity of training LRMs has first been addressed by \citet{nagata2006clustered}, who propose to use POS tags and word clustering methods and distinguish the first or last word of a phrase as the head of the phrase.



Complementary to our work, \citet{Galley:2008:SEH:1613715.1613824} introduced hierarchical reordering models that group phrases occurring next to the current phrase into blocks, ignoring the internal phrase composition within a block, which biases orientations more towards monotone and swap. At the same time, orientations are still conditioned on entire phrase-pairs, which means that their approach suffers from the same sparsity problems as LRMs. This problem has been more directly addressed by \citet{cherry:2013:NAACL-HLT}, who uses unsupervised word classes for infrequent words in phrase-pairs in the form of sparse features. Like \citet{nagata2006clustered}, the first and last words of a phrase-pair are used as features in this model. Unfortunately, this approach also introduces thousands of additional sparse features, many of which have to be extracted \emph{during} decoding, requiring changes to the decoder as well as a sizable tuning set.


\citet{durrani2014investigating} investigate the effect of generalized word forms on reordering in an n-gram-based operation sequence model, where they use different generalized representations including POS tags and unsupervised word classes to generalize reordering rules to similar cases with unobserved lexical operations. 

While the approaches above use discrete representations, \citet{li2014neural} propose a discriminative model using continuous space representations of phrase-pairs to address data sparsity problems. They train a neural network classifier based on recursive auto-encoders to generate vector space representations of phrase-pairs and base reordering decisions on those representations. They apply their model as an additional hyper-graph reranking step since direct integration into the decoder would make hypothesis recombination more complex and substantially increase the size of the search space. 

In addition to reordering models, several approaches have used word classes to improve other models within a statistical machine translation system, including translation \citep{wuebker2013improving} and language models, where the problem of data sparsity is particularly exacerbated for morphologically rich target languages \citep{chahuneau2013translating,bisazza2014class}.

\section{Model Definition} \label{ModelDef}
In this section, we propose two different types of model that use different words as a source of information to better estimate reordering distributions for sparse phrase-pairs. Each model uses a different generalization scheme to obtain less sparse but still informative representations.

Additionally, each model follows a different notion of what is defined to be an important word. For the first type of models, we follow \citet{nagata2006clustered} assuming the border words of a phrase-pair as the most important words. However, instead of just using the border words as \citet{nagata2006clustered} do, we use all words, but assume that the importance increases by getting closer to the borders of a phrase-pair. For the second type, we assume the exposed heads to be the most important words, following \citep{chelba2000structured, garmash2015bilingual, li2012head}.
 
\subsection{Interpolated Back-Off Sub-Phrases}\label{subsec:int_backoff}
In n-gram language modeling shorter n-grams have been used to smooth the probability distributions of higher-order, sparse n-grams. Lower-order n-grams form the basis of Jelinek-Mercer, Katz, Witten-Bell and absolute discount smoothing methods \citep{chen1999empirical}. For instance, Jelinek-Mercer smoothing linearly interpolates distributions of lower orders to smooth the distributions of higher-order n-grams.

We use this as a motivation that shorter phrase-pairs in lexicalized reordering models could play the role of lower-order n-grams in language model smoothing. But while backing off is obvious in language modeling, it is not straightforward in the context of lexicalized reordering models as there are several plausible ways to shorten a phrase-pair, which is further complicated by the internal word alignments of the phrase-pairs.

\begin{figure*}[thb]
\centering
\begin{subfigure}[t]{0.5\textwidth}
	\centering
	\includegraphics[scale=1]{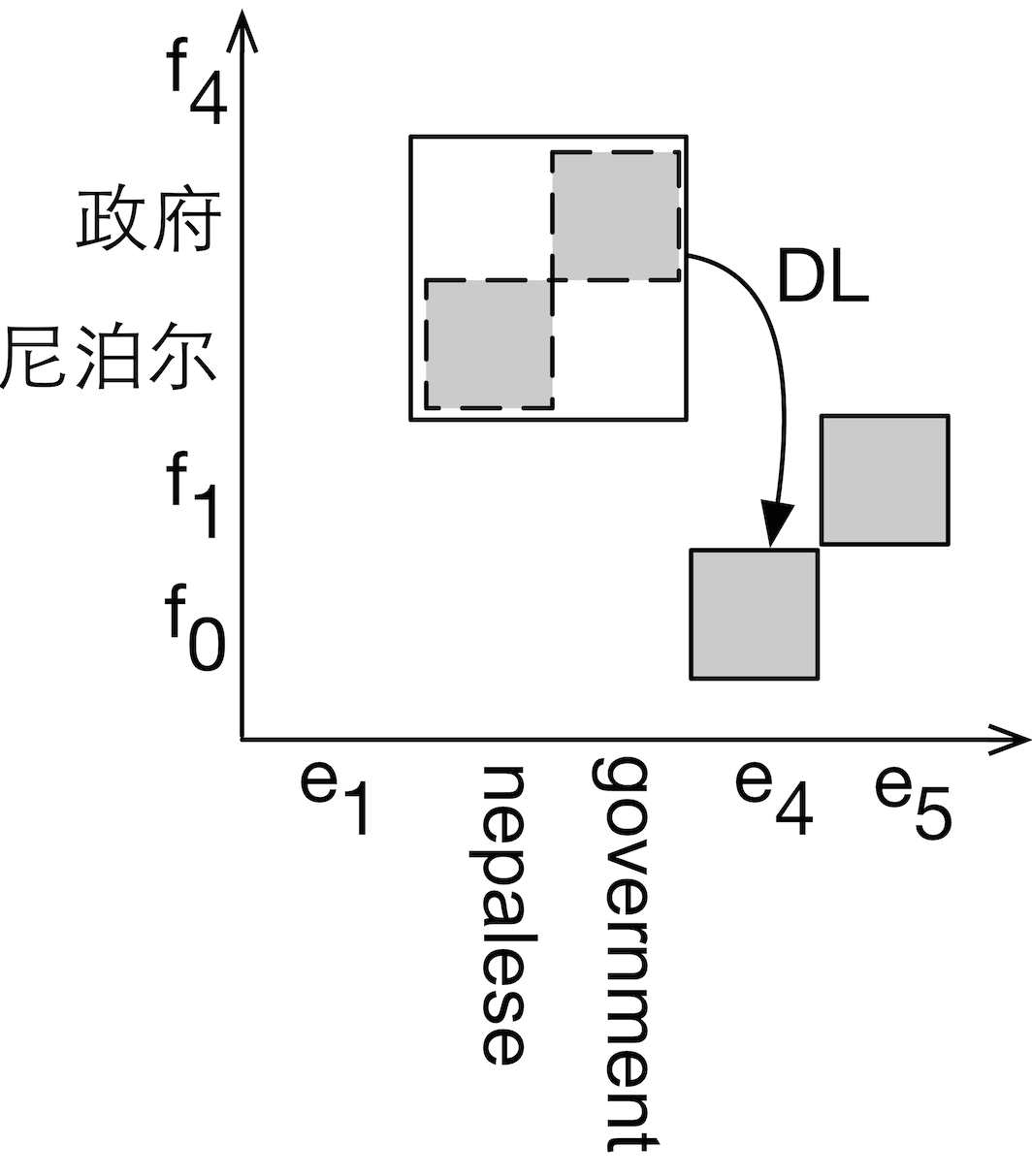}
	\caption{Original phrase-pair}\label{fig:oPhrase}
\end{subfigure}%
~
\begin{subfigure}[t]{0.5\textwidth}
	\centering
	\includegraphics[scale=1]{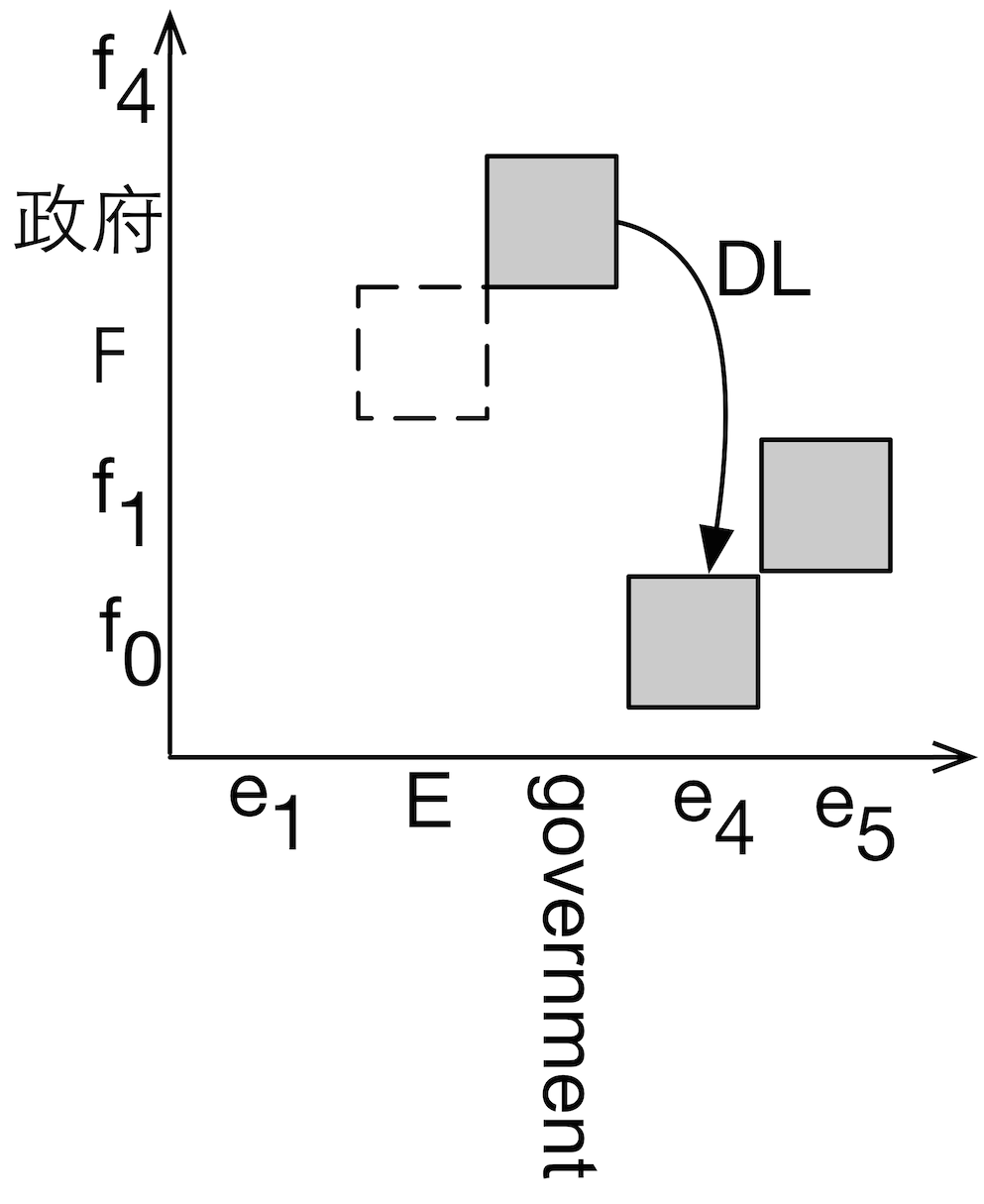}
	\caption{Eligible shortening}\label{fig:eligible}
\end{subfigure}%
\\
\par\bigskip 
\begin{subfigure}[t]{0.5\textwidth}
	\centering
	\includegraphics[scale=1]{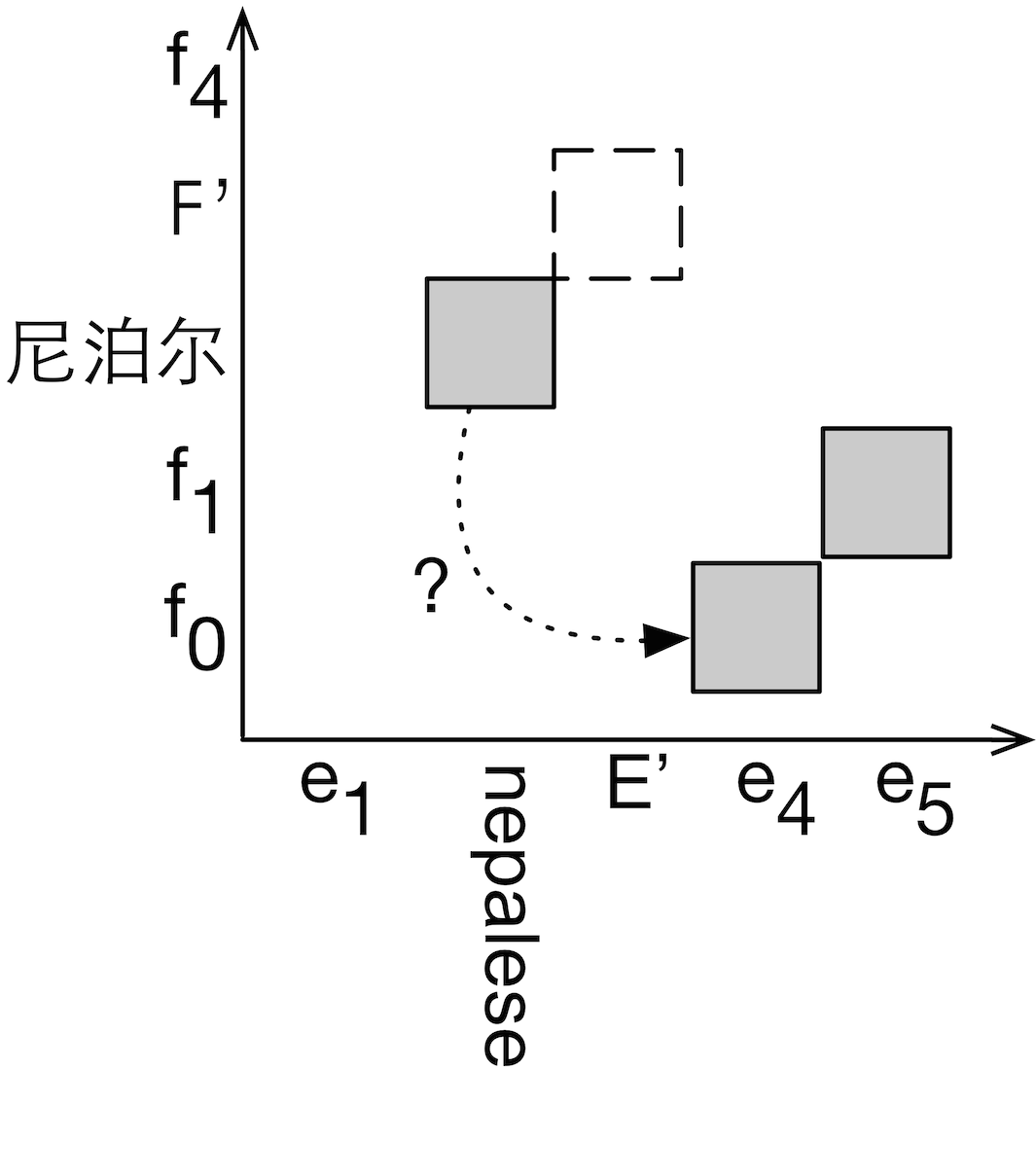}
	\caption{Ineligible shortening}\label{fig:Ineligible}
\end{subfigure}%
~
\begin{subfigure}[t]{0.5\textwidth}
	\centering
	\includegraphics[scale=1]{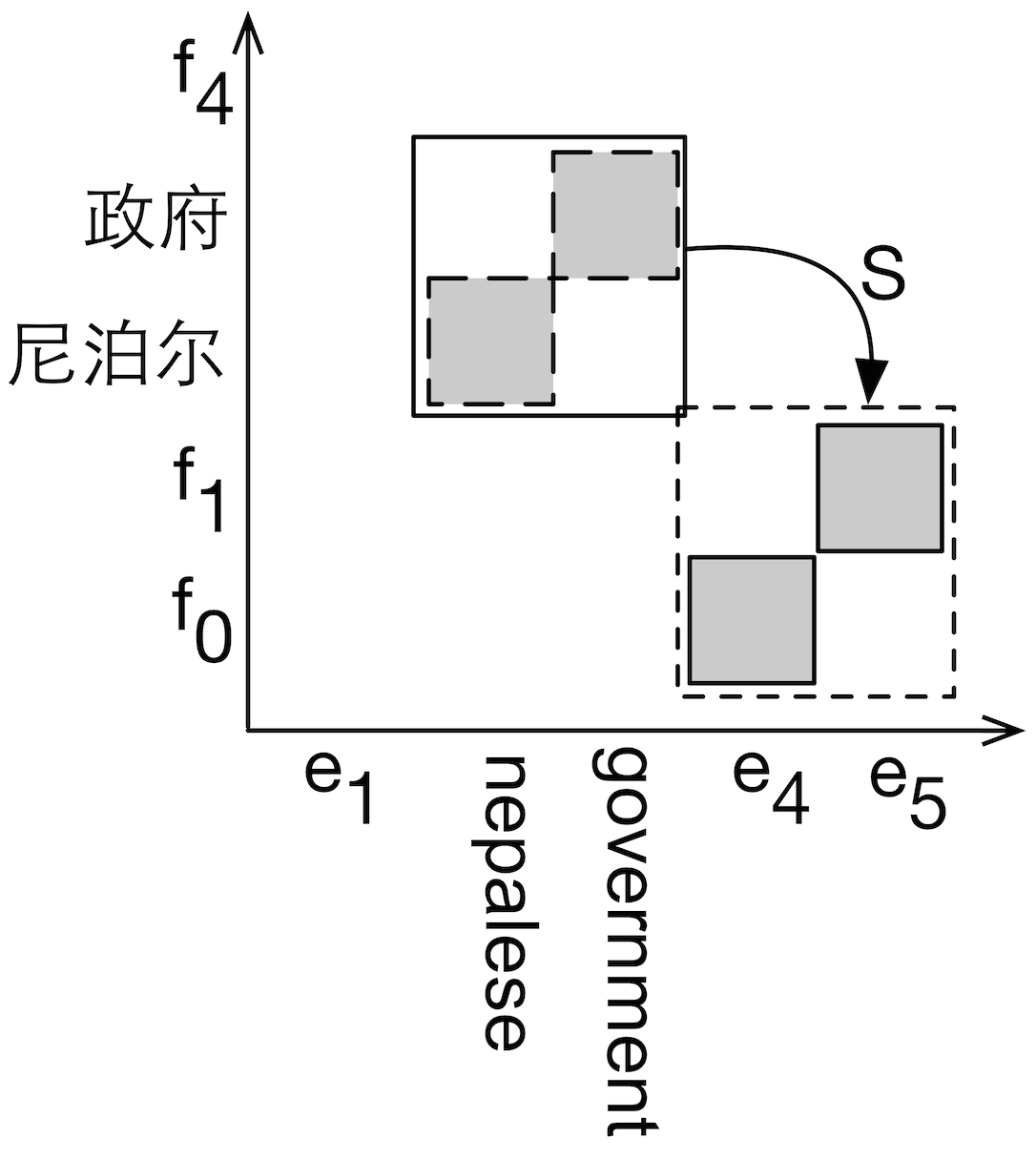}
	\caption{Hierarchical reordering situation}\label{fig:HRM}
\end{subfigure}
	\caption{An example of an original phrase-pair (a) and backing off to shorter phrase-pairs using eligible sub-phrase-pairs (b). For comparison, we also include an example of a grouping of phrases as done in HRM (d). We have also included an invalid example of shortening (c) to provide a comprehensive case.}
\label{fig:subphrase}
\end{figure*}      

The example in Figure~\ref{fig:subphrase} illustrates how sub-phrase-pairs (Figure \ref{fig:eligible} and \ref{fig:Ineligible}) of the longer phrase-pair (Figure \ref{fig:oPhrase}) can be used to estimate the \textit{discontinuous left} orientation for a longer, infrequent phrase-pair. Following the strategy within language modeling to back off to shorter n-grams, we back off to the sub-phrase-pairs that are consistent with the inside alignment of the longer phrase-pair and provide a shorter and less sparse history.
In this example, the number of times that sub-phrase-pair (\zh{政府}, government), see Figure~\ref{fig:eligible}, appears with a \textit{discontinuous left} jump of the length of the previous phrase-pair for the next translation is considered when estimating the \textit{discontinuous left} orientation for the longer phrase-pair. 
On the other hand, the sub-phrase-pair (\zh{尼泊尔}, nepalese), see Figure~\ref{fig:Ineligible}, cannot be used to predict a future \textit{discontinuous left} of the long phrase-pair, as there is no direct way to connect it to ($f_0$, $e_4$). The difference between this model and HRM can be seen in Figure~\ref{fig:HRM}. HRM groups small phrase-pairs from the context into longer blocks and determines the orientation with respect to the grouped block, while our model looks into the phrase-pair itself and uses possible shortenings to better estimate the orientation distribution conditioned on the original phrase-pair. Our model can be applied to HRMs as well as LRMs to estimate a better distribution for long and infrequent phrase-pairs. This can be done without requiring to retrain the models and by only using the statistics resulting from training.

In order to provide a formal definition of which sub-phrase-pairs to consider when backing off, let us assume that $A$ is the set of alignment connections between the source $\bar{f}$ and target $\bar{e}$ side of a longer phrase-pair. The set of eligible sub-phrase-pairs, $E_{\bar{f}, \bar{e}}$, is defined as follows:
\begin{equation}
\label{emc:eligibleSubPhrases}
\begin{aligned}
E_{\bar{f}, \bar{e}}= \{(\bar{f}^{[l,k]}, \bar{e}^{[l^\prime,n]}) \mid & \ 1 \leq k \leq m, \ 0 \leq l \leq k, 0 \leq l^\prime \leq n \ \ \text{and } \\ 
& (\bar{f}^{[l,k]}, \bar{e}^{[l^\prime,n]}) \text{ is consistent with } A \ \text{ if } l > 0 \wedge l^\prime > 0\ \},
\end{aligned}
\end{equation}
where $\bar{f}^{[l,k]}$ is a sub-phrase of $\bar{f}$ with length $l$ that ends at the $k$-th word of $\bar{f}$, $m$ and $n$ are the lengths of $\bar{f}$ and $\bar{e}$ respectively and the consistency with the alignment is ensured by the following three conditions \citep{koehn05iwslt}:

\begin{enumerate}\setlength{\itemsep}{-2pt}
\item $\exists e_i \in \bar{e}^{[l^\prime,n]}, f_j \in \bar{f}^{[l,k]} : (i,j) \in A$
\item $\forall e_i \in \bar{e}^{[l^\prime,n]} : (i,j) \in A \Rightarrow f_j \in \bar{f}^{[l,k]}$
\item $\forall f_i \in \bar{f}^{[l,k]}: (i,j) \in A \Rightarrow e_i \in \bar{e}^{[l^\prime,n]}$
\end{enumerate}

\begin{figure}[thb!]
\begin{center}
	\includegraphics[scale=0.50]{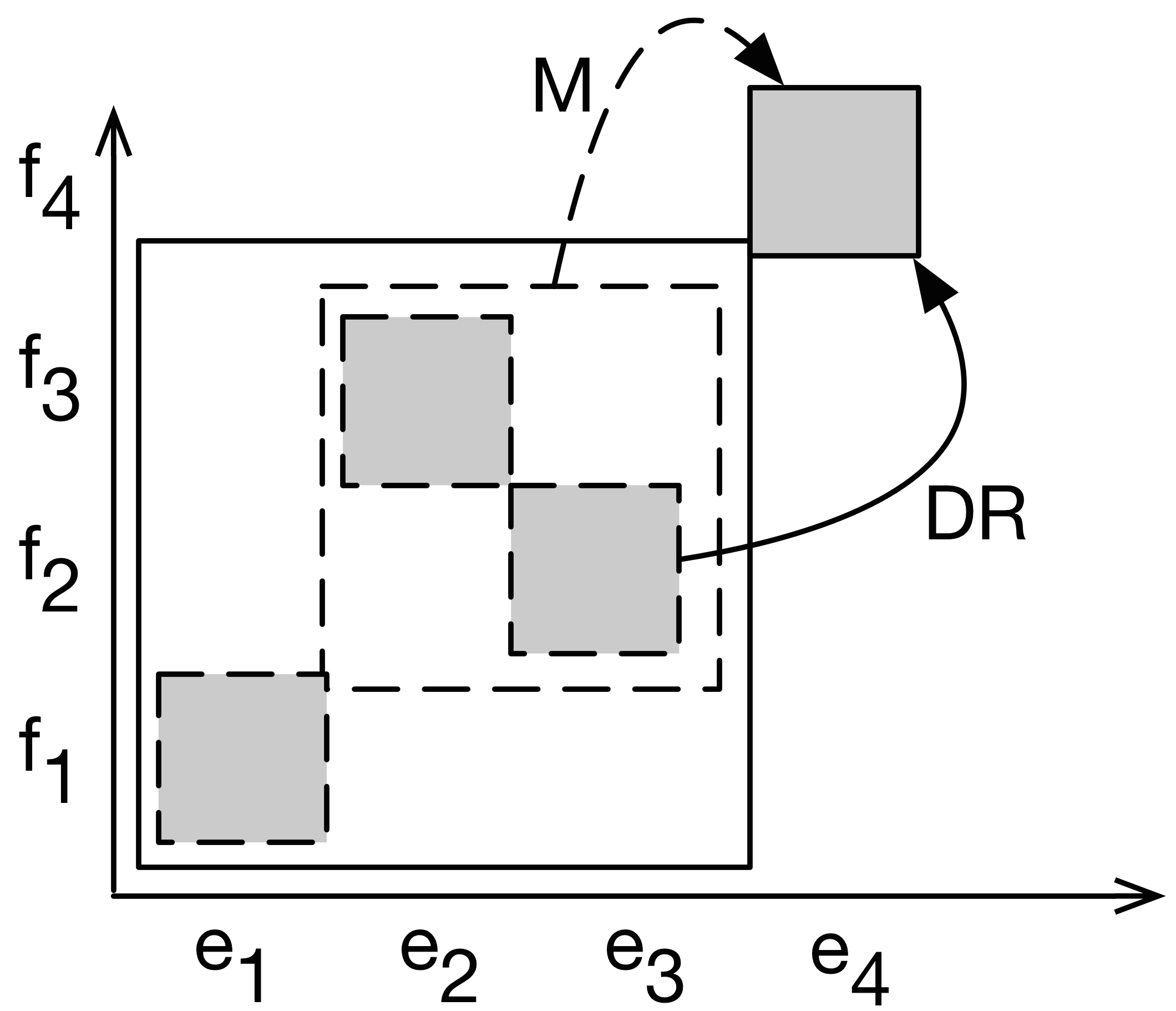}
\end{center}
	\caption{The distributions needed to estimate the conditional probability $\hat{p}(M \mid f_{1}f_{2}f_{3},e_{1}e_{2}e_{3})$ include 
$p(DR\mid f_{2},e_{3})$ and $p(M\mid f_{2}f_{3},e_{2}e_{3})$}
	\label{fig:sutiation}
\end{figure}

Considering Figure~\ref{fig:sutiation}, it is clear why ($\bar{f}^{[1,2]}$, $\bar{e}^{[1,3]}$) and ($\bar{f}^{[2,3]}$, $\bar{e}^{[2,3]}$) are considered eligible shortenings. Other possible shortenings such as ($\bar{f}^{[2,2]}$, $\bar{e}^{[3,3]}$) and ($\bar{f}^{[1,3]}$, $\bar{e}^{[1,2]}$) either violate the consistency conditions or do not run up to the end of the target side of the original phrase-pair as in the definition above. Note that $n$ is a constant here and $\bar{e}^{[l^\prime,n]}$ means that all sub-phrase-pairs must finish at the end of the target side of the original phrase-pair. This is necessary as otherwise one cannot directly determine the orientation with respect to the next phrase-pair. This condition makes sure that cases such as the one in Figure~\ref{fig:Ineligible}, are identified as an invalid way of shortening.



In our first model, we compute the smoothed orientation distribution conditioned on a phrase-pair by linearly interpolating the distribution of all eligible sub-phrases as follows:
\begin{equation}
\label{emc:linearInterpolation}
\hat{P}(o \mid \bar{f}, \bar{e}) = \sum\nolimits_{(\bar{f}^{[l,k]}, \bar{e}^{[l^\prime,n]}) \in E_{\bar{f}, \bar{e}}} \lambda_{l,l^\prime}P(\Omega(\bar{f}^{[l,k]}, \bar{f}, o) \mid \bar{f}^{[l,k]}, \bar{e}^{[l^\prime,n]}).
\end{equation}

\noindent Here $E_{\bar{f}, \bar{e}}$ is the set of eligible sub-phrase-pairs and $\bar{f}^{[l,k]}$ indicates a sub-phrase of $\bar{f}$ with length $ l $ ending at the $k$-th word of $ \bar{f} $, $ \bar{e}^{[l^\prime,n]} $ is a sub-phrase of $ \bar{e} $ with the length of  $ l^\prime $ that ends at the last word of $\bar{e}$,  and the function $\Omega(\bar{f}^{[l,k]}, \bar{f}, o) $ returns the correct orientation considering the position of source sub-phrase $ \bar{f}^{[l,k]} $ with respect to either end of the source phrase $\bar{f}$ and orientation $o$.

In order to compute the linear interpolation over the conditional distributions of the sub-phrase-pairs, we have to determine the weight of each term in the linear interpolation (Equation \ref{emc:linearInterpolation}). Here, we use expectation-maximization (EM) to find a set of weights that maximizes the likelihood over a held-out data set, which is word-aligned using GIZA++ \citep{och2003systematic}. The alignments are refined using the grow-diag-final-and~\citep{Koehn:2003:SPT:1073445.1073462} heuristic. We extract phrase-pairs using a common phrase extraction algorithm \citep{koehn05iwslt} and count the number of occurrences of orientations for each phrase-pair. These counts are used with unsmoothed reordering probabilities learned over the training data to compute the likelihood over the held-out data. We designed the EM algorithm to learn a set of lambda parameters for each length combination of the original phrase-pairs. To reduce the number of parameters, we assume that all sub-phrases with the same length on the source and target side share the same weight. 
This model is referred to as the Back-off model in the remainder of this chapter.

Next, we introduce a model based on the same scheme of shortening into sub-phrase-pairs, but a different way of combining them to smooth the target distribution.

\subsection{Recursive Back-Off MAP Smoothing}\label{sec:recbackoff}
Above we used linear interpolation to estimate the orientation distributions of shorter sub-phrase-pairs. Here we investigate another method which aims to affect the distributions of frequent phrase-pairs to a lesser extent than those of non-frequent ones.

To this end, we use recursive Maximum a Posteriori (MAP) smoothing~\citep{cherry:2013:NAACL-HLT} to estimate the distribution of the original phrase-pair. In our interpolated back-off model, see Section~\ref{subsec:int_backoff}, all phrase-pairs with the same length will get the same fraction of their estimated distribution from their sub-phrase-pairs of the same length since the interpolation parameters are the same for these phrase-pairs and sub-phrase-pairs. On the other hand, for more frequent phrase-pairs, the maximum likelihood distribution of the phrase-pair itself is more reliable than the distributions of its sub-phrase-pairs. Thus, a model relying more on the distribution of the original phrase-pairs for frequent phrase-pairs would be desirable.

\begin{figure}[thb]
\begin{center}
	\includegraphics[scale=0.50]{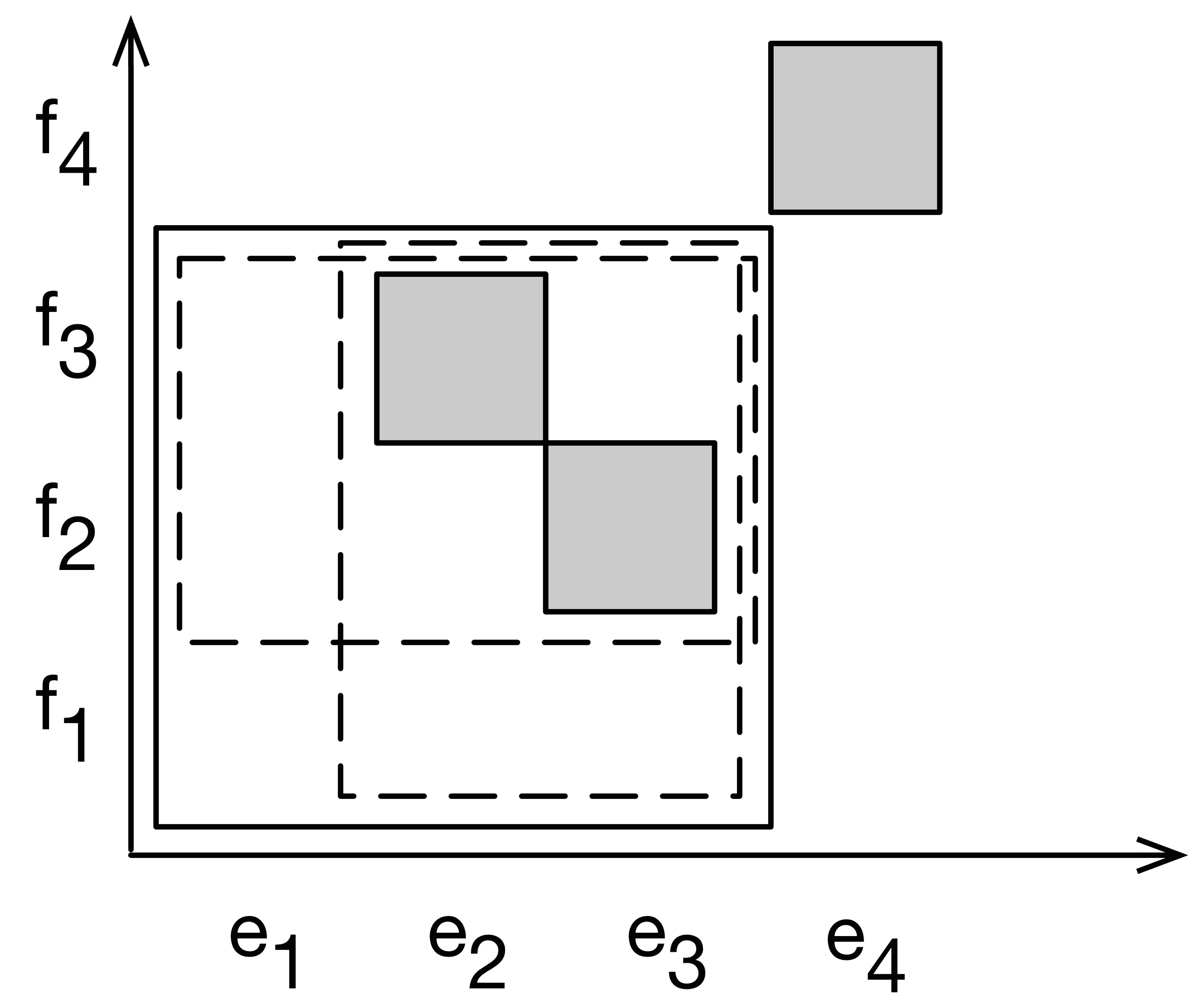}
\end{center}
	\caption{Illustration of sub-phrases where neither of the two pairs $(f_1f_2f_3, e_2e_3)$ and $(f_2f_3, e_1e_2e_3)$ of the original phrase-pair $(f_1f_2f_3, e_1e_2e_3)$ is a sub-phrase of the other.}
	\label{fig:nonChildSubPhr}
\end{figure}

To achieve this, we use a formulation similar to recursive MAP smoothing (Equation \ref{emc:2}) with recursively backing off to the distributions of shorter sub-phrase-pairs. At each recursion step we use the distribution of the longest sub-phrase-pair as the prior distribution. Taking our definition for eligible sub-phrase-pairs into account (Equation \ref{emc:eligibleSubPhrases}), for every two sub-phrase-pairs one is always a sub-phrase-pair of the other, if the original phrase-pair does not include unaligned words. For original phrase-pairs that include unaligned words, see the example in Figure~\ref{fig:nonChildSubPhr}, there could be pairs of sub-phrases where neither of them is a sub-phrase of the other. In these cases, we include the distributions of all sub-phrases with the same \textit{equivalent sample size} as prior distributions, see Section~\ref{sec:problem_def}. The estimated probability distribution of a phrase-pair ($\bar{f}$,$\bar{e}$) is defined as follows:
\begin{equation}
\label{emc:recursiveBackoff}
\hat{P}(o \mid \bar{f}, \bar{e}) = \frac{C(o,\bar{f},\bar{e}) + \sum_{(\bar{f}_{L},\bar{e}_{L}) \in L_{\bar{f},\bar{e}}}\alpha\hat{P}(\Omega(\bar{f}_{L}, \bar{f}, o) \mid \bar{f}_{L}, \bar{e}_{L})}{\sum_{o \in O}C(o,\bar{f},\bar{e}) + \sum_{(\bar{f}_{L},\bar{e}_{L}) \in L_{\bar{f}, \bar{e}}}\alpha},
\end{equation}
where $L_{\bar{f},\bar{e}}$ refers to the set of eligible sub-phrase-pairs of $(\bar{f},\bar{e})$ that are not sub-phrase-pairs of each other and which is defined as follows:
\begin{multline}
L_{\bar{f},\bar{e}} = \biggl\{ (\bar{f^{\prime}}, \bar{e^{\prime}}) \in E_{\bar{f},\bar{e}}\setminus \left\{(\bar{f}, \bar{e})\right\} | \ \neg\exists (\bar{f^{\prime\prime}}, \bar{e^{\prime\prime}}) \in E_{\bar{f},\bar{e}}\setminus\left\{(\bar{f}, \bar{e}), (\bar{f^{\prime}}, \bar{e^{\prime}})\right\}: \\ 
\shoveright{\bar{f^{\prime}} \sqsubseteq \bar{f^{\prime\prime}} \wedge \bar{e^{\prime}} \sqsubseteq \bar{e^{\prime\prime}} \biggr\}}\\
\end{multline}
\noindent Here, $\bar{f^{\prime}} \sqsubseteq \bar{f^{\prime\prime}}$ means that $\bar{f^{\prime}}$ is a sub-phrase of or equal to $\bar{f^{\prime\prime}}$. As a result, $L_{\bar{f},\bar{e}}$ is the set of longest eligible sub-phrase-pairs of the original phrase-pair where none of them is a sub-phrase of the others. For $E_{\bar{f},\bar{e}}$, see Equation~\ref{emc:eligibleSubPhrases}. $O$ is the set of possible orientations. 
We refer to this model as RecursiveBackOff model from now on. We also refer to both this model and the BackOff model described in the previous section as back-off models.



\subsection{Dependency Based Generalization}\label{DepSec}


The methods described so far generalize the original phrase-pairs by shortening towards the last aligned words as the most important words to define the reordering behavior of a phrase-pair. In the remainder of this section, we use dependency parses to define how to generalize the original phrase-pair and shorten towards important words.

\begin{figure}[thb]
\begin{center}
	\includegraphics[scale=0.8]{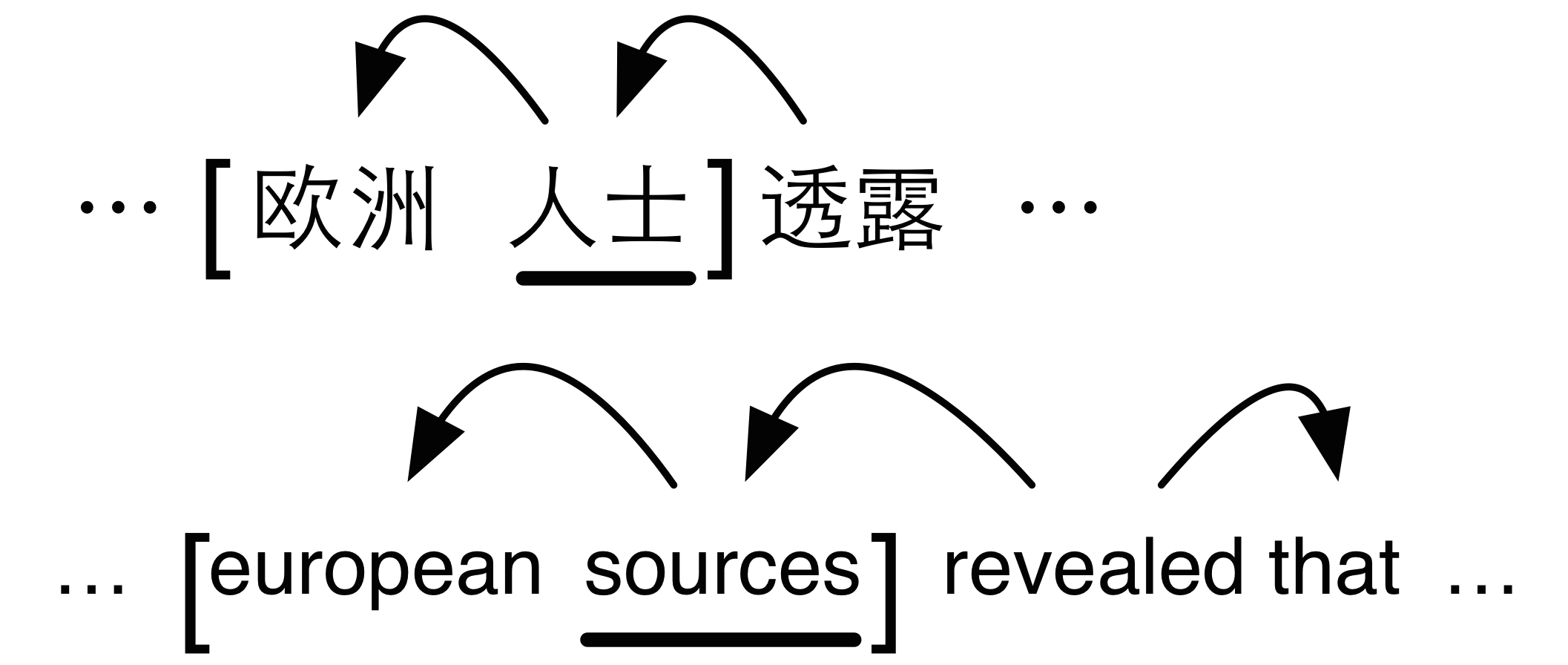}
\end{center}
	\caption{Examples of exposed heads in a Chinese-English phrase-pair (between square brackets). The underlined words are the exposed heads since they have an incoming dependency originating outside of the phrase.}
	\label{fig:exposedHeads}
\end{figure}

\begin{sidewaysfigure}
\begin{center}
	\includegraphics[scale=7]{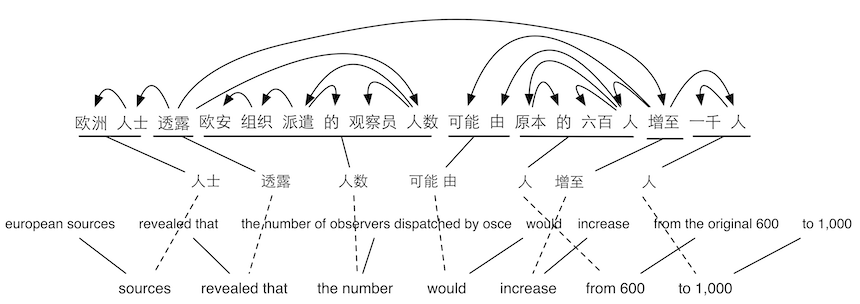}
	\caption{A source (Chinese) and the target (English) sentence pair with the extracted phrases and their corresponding exposed heads of each phrase from the source and the target side. Solid lines connect a phrase to its exposed heads and dashed lines connect aligned words. The target side dependency parse is not shown for visualization purposes.}
	\label{fig:exposedHeads_bi_sentence}
\end{center}
\end{sidewaysfigure}

Head-driven hierarchical phrase-based translation \citep{li2012head} argues that using heads of phrases can be beneficial for better reordering. In our work, we define the heads of a phrase to be its \textit{exposed heads}. Given a dependency parse, the exposed heads are all words inside a subsequence that play the role of a modifier in a dependency relation with a word outside of the subsequence. Figure~\ref{fig:exposedHeads} shows an example of exposed heads in a phrase-pair. The underlined words are the exposed heads of the phrase-pair. Exposed heads have been used in multiple linguistically motivated approaches as strong predictors of the next word in structured language models \citep{chelba2000structured,garmash2015bilingual} and the next translation rule to be applied in a hierarchical phrase-based translation model~\citep{chiang2007hierarchical, li2012head}. Figure~\ref{fig:exposedHeads_bi_sentence} shows a sentence pair with the extracted phrase-pairs and their corresponding exposed heads.

In our model, besides training a regular lexicalized or hierarchical reordering model on surface forms of phrases, we train another reordering model which keeps the exposed heads lexicalized and replaces the remaining words in a phrase-pair by a generalized representation.

Assume that $RE$ is the set of dependency relations in the dependency parse tree of sentence $S$. We consider each relation as an ordered pair $(w_l^\prime, w_k)$ which means that word $w$ at index $k$ of sentence $S$ modifies word $w^\prime$ at index $l$. In addition, assuming $f_{i}^{j}$ is a phrase in $S$, starting from the $i$-th and ending with the $j$-th word in sentence $S$, then the generalization $w_G$ of a word is:
\[
w_G =
\begin{cases}
w_k & \mathrm{if}\ w_k \in f_{i}^{j}, \exists (w_{l}^{\prime}, w_k) \in RE:\\ & l < i \text{ or } l > j \text{ or } w^\prime=\mathit{ROOT}\\
\mathit{Gen}(w_k) & \mathrm{otherwise}.
\end{cases}
\]
Here, $k$ and $l$ are indices of words $w$ and $w^\prime$ in sentence $S$ and $\mathit{ROOT}$ is the root of the dependency parse of sentence $S$. The function $\mathit{Gen}(w)$ returns a generalization form for word $w$. We define this function in three different ways to create three different models.
\theoremstyle{definition}
\begin{definition}
\label{def:1}
$\mathit{Gen}(w_k)$ = POS\_tag($w_k$)
\end{definition}

\begin{definition}
\label{def:2}
$\mathit{Gen}(w_k)$ = \texttt{<mod>}  if $w_{k-1}$ is not equal to \texttt{<mod>} and nothing otherwise (replaces all consecutive modifiers with a single \texttt{<mod>} symbol)
\end{definition}

\begin{definition}
\label{def:3}
$\mathit{Gen}(w_k)$ remove $w_k$.
\end{definition}
\noindent Here, \texttt{<mod>} is a designated symbol that replaces the modifiers of the exposed heads. The question is how to use these generalizations to improve the estimation of reordering distribution for each phrase-pair. Our first model applies a generalization to the bilingual training data and creates a reordering model similar to the regular lexicalized reordering model, but based on the relative frequency of the generalized phrase-pairs. In practice, a phrase-pair can have multiple generalizations due to different dependency parses in different contexts. Since it is difficult to produce a dependency parse for the target side during decoding, we assume that a phrase-pair will always have one possible generalization. Under this assumption, we can approximate the orientation distribution of a phrase-pair to be the orientation distribution of its generalized form:
\begin{equation}
\label{emc:3}
\hat{P}(o \mid \bar{f}, \bar{e}) = P(o \mid \bar{f_{G}}, \bar{e_{G}}).
\end{equation}   
\noindent Here, $\bar{f_{G}}$ and $\bar{e_{G}}$ are word by word generalizations of $\bar{f}$ and $\bar{e}$. In the case of multiple generalizations we use the one maximizing $P(\bar{f_{G}}, \bar{e_{G}} \mid \bar{f}, \bar{e})$. Depending on which of the three definitions for $\mathit{Gen}(w_k)$ we use, we name our models PMLH (POSed Modifiers, Lexicalized Heads), see Definition~\ref{def:1}, MMLH (Merged Modifiers, Lexicalized Heads), see Definition~\ref{def:2}, and LH (Lexicalized Heads), see Definition~\ref{def:3}.

As an alternative model, we propose to use the generalized distributions as a prior distribution in Dirichlet smoothing, where the distribution of each phrase-pair is smoothed with the distribution of its generalized form. This results in the following smoothing formulation:
\begin{equation}
\label{emc:LHSmoothed}
P(o \mid \bar{f}, \bar{e}) = \frac{C(o, \bar{f}, \bar{e})+ \sigma P(o \mid \bar{f_{G}}, \bar{e_{G}})}{ \sum\nolimits_{o^\prime} {C(o^\prime, \bar{f}, \bar{e})} + \sigma}.
\end{equation}
\noindent Here, $P(o \mid \bar{f_{G}}, \bar{e_{G}})$ is the generalized distribution, see Equation~\ref{emc:3}, and $C(o, \bar{f}, \bar{e})$ refers to the number of times a phrase-pair co-occurs with orientation $o$, and $\sigma$ is the equivalent sample size, as defined in Section~\ref{sec:problem_def}. Our assumption is that this should result in more accurate distributions since it affects the distributions of frequent phrase-pairs to a lesser extent. 


\section{Experiments}
\label{sec:exp_research_1}
We evaluate our models for Chinese-to-English translation. Our training data consists of the parallel and monolingual data released by NIST's OpenMT campaign, with MT04 used for tuning and news data from MT05 and MT06 for testing, see Table~\ref{tbl:3}. Case-insensitive BLEU \citep{papineni2002bleu}, RIBES~\citep{isozaki2010automatic} and translation error rate (TER) \citep{snover2006study} are used as evaluation metrics. 

\begin{table}[t]
\begin{center}
\small
\caption{\label{tbl:3} Statistics for the Chinese-English bilingual corpora used in all experiments. Token counts for the English side of dev and test sets are averaged across all references.}
\begin{tabular}{
|l|r|r|r|}
\hline
\multicolumn{1}{|c|}{Corpus} & \multicolumn{1}{c|}{Lines} & \multicolumn{1}{c|}{Tokens (ch)} & \multicolumn{1}{c|}{Tokens (en)}\\
\hline
train & 937K &  22.3M & 25,9M\\

MT04 (dev) & 1,788 & 49.6K & 59.2K\\

MT05 (test) & 1,082 & 30.3K & 35.8K\\

MT06 (test) & 1,181 & 29.7K & 33.5K\\ 

\hline
\end{tabular}
\end{center}
\end{table}

\subsection{Baseline}
We use an in-house implementation of a phrase-based statistical machine translation system similar to Moses \citep{koehn2007moses}. Our system includes all the commonly used translation, lexical weighting, language, lexicalized reordering, and hierarchical reordering models as described in Chapter~\ref{chapter:background}. We use both lexicalized and hierarchical reordering models together, since this is the best model reported by \citet{Galley:2008:SEH:1613715.1613824} and our smoothing methods can be easily applied to both models. Word alignments are produced with GIZA++ \citep{och2003systematic} and the grow-diag-final-and alignment refinement heuristic~\citep{Koehn:2003:SPT:1073445.1073462}. A 5-gram language model is trained on the English Gigaword corpus containing 1.6B tokens using interpolated, modified Kneser-Ney smoothing. The lexicalized and the hierarchical reordering models are trained with relative and smoothed frequencies using Dirichlet smoothing (Equation~\ref{emc:1}), for both left-to-right and right-to-left directions distinguishing four orientations: monotone (M), swap (S), discontinuous left (DL), and discontinuous right (DR). Feature weights are tuned using PRO \citep{hopkins2011tuning} and statistical significance of the differences are computed using approximate randomization \citep{riezler2005some}.

In addition to the baseline, we reimplemented the 2POS model by \citet{nagata2006clustered}, which uses the POS tag of the first and last words of a phrase-pair to smooth the reordering distributions. The 2POS model is used in combination with the baseline lexicalized and hierarchical reordering models. Comparing our models to the 2POS model allows us to see whether backed-off sub-phrases and exposed heads of phrase-pairs yield better performance than simply using the first and last words. 

\begin{table*}[tbh!]
\centering
\small
\caption{\label{tbl:news4} Model comparison using BLEU, TER (lower is better), and RIBES over news data, which is a combination of newswire and broadcast\_news for MT06 and just newswire for MT05. Scores better than the baseline are in italics. $\blacktriangleup$ and $\smalltriangleup$ indicate statistically significant improvements at $p < 0.01$ and $p < 0.05$, respectively. The left hand side $\blacktriangleup$ or $\smalltriangleup$ is with respect to Lex+Hrc and the right hand side with respect to Nagata's 2POS. PMLH refers to the model using POS tags for modifiers and keeps exposed heads lexicalized. MMLH merges modifiers and keeps exposed heads lexicalized. LH removes modifiers. LHSmoothed uses the LH model with Dirichlet smoothing.}
\resizebox{1\linewidth}{!}{
\begin{tabular}{
|p{2.3 cm}|l|l|l|l|l|l|l|}
\hline
Model & \multicolumn{2}{c|}{MT05} & \multicolumn{2}{c|}{MT06} & \multicolumn{3}{c|}{MT05 + MT06} \\
\hline
 & \multicolumn{1}{c|}{BLEU$\uparrow$} & \multicolumn{1}{c|}{TER$\downarrow$}  & \multicolumn{1}{c|}{BLEU$\uparrow$} & \multicolumn{1}{c|}{TER$\downarrow$} & \multicolumn{1}{c|}{BLEU$\uparrow$} & \multicolumn{1}{c|}{TER$\downarrow$} & \multicolumn{1}{c|}{RIBES$\uparrow$}\\
 \cline{2-8}
Lex+Hrc & 32.25 & 60.13  & 33.00 & 57.17 & 32.84 & 58.62 & 79.24 \\
Nagata's 2POS & 32.20 & 60.31 & 33.13 & 57.07 & 32.87 & 58.66 & 79.11\\
\hline
BackOff  & \textit{32.40} & 60.45  & \textit{33.10} & \textit{57.00} & \textit{33.00} & 58.69 & 79.07\\
RecursiveBackOff & \textit{32.37} & 60.29 & \textit{33.34{$^{\smalltriangleup, \castlinghyphen}$}} & \textit{57.08} & \textit{33.05{$^{\smalltriangleup, \castlinghyphen}$}} & 58.65 & \textit{79.28} \\
PMLH  & \textit{32.41} & \textit{60.08}  & \textit{33.26{$^{\smalltriangleup, \castlinghyphen}$}} & \textit{57.05} & \textit{33.04{$^{\smalltriangleup, \castlinghyphen}$}} & \textit{58.53} & \textit{79.26} \\
MMLH  & \textit{32.62{$^{\blacktriangleup, \smalltriangleup}$}} & \textit{59.84{$^{\castlinghyphen, \blacktriangleup}$}} & \textit{33.18}  & \textit{56.91{$^{\smalltriangleup, \castlinghyphen}$}} & \textit{33.04{$^{\smalltriangleup, \castlinghyphen}$}} & \textit{58.35{$^{\blacktriangleup, \smalltriangleup}$}} & \bfseries{79.43}\\
LH  & \textit{32.64{$^{\blacktriangleup, \smalltriangleup}$}} & \textit{59.85{$^{\smalltriangleup, \blacktriangleup}$}} & \textit{33.26{$^{\smalltriangleup, \castlinghyphen}$}} & \textit{56.85{$^{\blacktriangleup, \castlinghyphen}$}} & \textit{33.11{$^{\blacktriangleup, \smalltriangleup}$}} & \textit{58.32{$^{\blacktriangleup, \blacktriangleup}$}} & \textit{79.34}\\
LHSmoothed  & \bfseries{32.65{$^{\blacktriangleup, \blacktriangleup}$}} & \bfseries{59.80{$^{\smalltriangleup, \blacktriangleup}$}}  & \bfseries{33.38{$^{\smalltriangleup, \castlinghyphen}$}} & \bfseries{56.77{$^{\blacktriangleup, \smalltriangleup}$}} & \bfseries{33.20{$^{\blacktriangleup, \smalltriangleup}$}} & \bfseries{58.25{$^{\blacktriangleup, \blacktriangleup}$}} & \textit{79.35}\\ 
\hline
\end{tabular}}
\end{table*}

\subsection{Experimental Results}
\label{subsec:result_ch1}

We compare the baseline to the models described in Section~\ref{ModelDef}. For all systems other than the baseline, the lexicalized and the hierarchical reordering models are replaced by the corresponding smoothed models. When computing the RecursiveBackOff model (Section~\ref{sec:recbackoff}), using Equation~\ref{emc:recursiveBackoff}, we set the value of $\alpha$ to 10, following \citep{cherry:2013:NAACL-HLT} and \citep{chen-foster-kuhn:2013:NAACL-HLT}.


For the model using exposed heads, we use the dependency parses of the source and the target side of the training corpus. The Stanford Neural-network dependency parser \citep{chen2014fast} is used to generate parses for both sides of the training corpus. From a dependency parse, we extract the smallest subtree that includes all incoming and outgoing relations of the words of a phrase. This is done for both the source and the target side phrases. Considering these subtrees, all words with an incoming connection from outside are exposed heads. Having identified the exposed heads, we apply the generalization methods introduced in Section~\ref{DepSec} to produce the estimated generalized distributions.


The experimental results for all models are shown in Table~\ref{tbl:news4}. As one can see, all our models achieve improvements over the baselines in terms of BLEU. The improvements for our back-off models are statistically significant only for RecursiveBackOff for MT06 and MT05+MT06. The improvements for MT05 by our dependency-based shortenings are statistically significant for all models except PMLH.
%
%
In the case of MT06, only the improvements resulting from MMLH are not statistically significant. However, both the PMLH and the MMLH model achieve the same improvements for MT05 and MT06 combined, and both are statistically significant. The LH model performs better than these models and also achieves higher improvements on MT05+MT06. This model generalizes much more than the other models and is the only model that changes the distributions of single word phrases which are among the most frequently used phrase-pairs. 
However, for frequent phrase-pairs, being mapped to the same generalization form can be potentially harmful, since it may change the reordering distribution for phrase-pairs which are frequent enough for reliable estimation in their original form. In order to be able to control the effect of the model on phrase-pairs based on their frequency, we use the distributions in LH as a prior distribution with Dirichlet smoothing (Equation~\ref{emc:LHSmoothed}). Using this formulation, frequent phrase-pairs are smoothed to a lesser degree. This results in the LHSmoothed model shown in the last row of  Table~\ref{tbl:news4} which achieves the best improvements for both MT05 and MT06.

In addition to BLEU, we also report results using TER. The results for TER are in line with BLEU. BLEU and TER are general translation quality metrics, which are known to be not very sensitive to reordering changes \citep{birch2010metrics}. To this end we also include RIBES \citep{isozaki2010automatic}, a reordering-specific metric that is designed to directly address word-order differences between translation candidates and reference translations in translation tasks with language pairs often requiring long-distance reorderings. As the values of the RIBES metric shows, our dependency models improve reordering more than the other models.

\subsection{Analysis}\label{sec:analysis}
The improvements achieved by our BackOff and RecursiveBackOff methods show that these models capture useful generalizations by shortening the phrase-pairs towards the last aligned words in the target side as the most important words. The difference between the two models indicates that shortening is less beneficial for frequent phrase-pairs and shorter phrase-pairs, which are less affected by lower-order distributions in the case of RecursiveBackOff. As a result, the RecursiveBackOff model achieves a slightly better performance.

The improvements achieved by our generalization models support our hypothesis that not all words inside a phrase-pair have the same impact on the reordering properties of the phrase-pair as a whole. The experimental results for the PMLH model show that the lexicalized form of modifier words inside a phrase-pair may increase data sparsity.

Observing the improvements achieved by MMLH, we can go further and say that even the number of modifiers of an exposed head in a phrase does not influence the reordering properties of a phrase-pair. 
The improvements achieved by our LH model show not only that the number of modifiers but also the mere presence or absence of them does not significantly influence the reordering properties of a phrase-pair.

\begin{table*}[thb!]
\small
\centering
\caption{\label{tbl:5} Number of times that phrase-pairs with different lengths and a frequency of less than 10 in the training data have been used by the baseline for MT06. Phrase-pairs occurring less than 10 times account for 
72\% of all phrases used during decoding of MT06.}
\begin{tabular}{
|c|l|r|r|r|r|r|r|r|}
\hline
 & \multicolumn{8}{c|}{Target Length}\\
 \cline{2-9}
 &  \multicolumn{1}{c}{}& \multicolumn{1}{c}{1} & \multicolumn{1}{c}{2} & \multicolumn{1}{c}{3} & \multicolumn{1}{c}{4} & \multicolumn{1}{c}{5} & \multicolumn{1}{c}{6} & \multicolumn{1}{c|}{7} \\
 \cline{2-9}
 \multirow{7}{*}{\rotatebox[origin=c]{90}{Source Length}} & 1 & 8232 & 2719 & 879 & 269 & 89 & 18 & 7 \\
 & 2 & 2344 & 1777 & 1055 & 410 & 158 & 58 & 18 \\
 & 3 & 316 & 390 & 376 & 252 & 100 & 61 & 29\\
 & 4 & 42 & 46 & 97 & 63 & 29 & 28 & 14 \\
 & 5 & 2 & 3 & 11 & 11 & 10 & 8 & 12 \\
 & 6 & 0 & 1 & 1 & 3 & 3 & 5 & 2 \\
 & 7 & 0 & 0 & 0 & 3 & 1 & 1 & 1 \\
  \hline
\end{tabular}
\end{table*}

One thing to bear in mind is that the PMLH and MMLH models do not change the distribution of single word phrase-pairs that are mostly frequent phrase-pairs, while the LH model does change these distributions as well. With the LHSmoothed model we have controlled this effect and decreased it to a negligible degree for frequent single word phrase-pairs. However, it still changes the distributions of infrequent single word phrase-pairs. Comparing the results of LH and LHSmoothed in Table~\ref{tbl:news4}, we suspect that the difference between the models for MT06 is due to the effect of frequent single word phrase-pairs.\footnote{We also used the distributions from PMLH and MMLH as priors in Dirichlet smoothing, but it did not lead to any noticeable changes in the results.} However, comparing the results of LHSmoothed with the other models we can say that even infrequent single word phrase-pairs benefit from the higher generalization level offered by this model. The statistics of infrequent phrase-pairs used during testing and their lengths are shown in Table~\ref{tbl:5}. This indicates why this model achieves the highest improvements. The table shows that 41\% of the infrequent phrase-pairs (frequency $<$ 10) used while translating MT06 have a length of one on both sides. Therefore, our models, apart from LH and LHSmoothed, do not affect almost half of the infrequent phrase-pairs. This probably explains why the back-off models achieve such small improvements when 75\% of infrequent phrase-pairs have lengths of less than 3 on both sides.    
\begin{table*}[thb]
\small
\centering
\caption{\label{tbl:6} Examples from MT05 illustrating reordering improvements over the baseline.}
\resizebox{1\linewidth}{!}{
\begin{tabular}{|l|p{10.5cm}|}
\hline
Source & \zh{... 由于 东京 和 汉城 当局 均 寄望 能 于 二○○五年 底 前 签署 自由 贸易 协定 ...} \\
\cline{1-2}
Baseline & ... tokyo and seoul authorities are to be placed in 2005 before the end of the signing of a free trade agreement ... \\
\cline{1-2}
LHSmoothed & ... tokyo and seoul authorities both in the hope of signing a free trade agreement before the end of 2005 ...\\
\cline{1-2}
Ref &  ... tokyo and seoul both hoped to sign a fta agreement by the end of 2005 ...\\
\hline
\multicolumn{1}{c}{} & \multicolumn{1}{c}{} \\
\hline
Source & \zh{俄罗斯 多 次 指控 西方 插手 东欧 事务 ...}\\
\cline{1-2}
Baseline & russia has repeatedly accused of meddling in the affairs of the western and eastern europe ...\\
\cline{1-2}
LHSmoothed & russia has repeatedly accused western intervention in the eastern european affairs ...\\
\cline{1-2}
Ref & russia has been accusing the west of interfering in the affairs of eastern europe ...\\
\hline
\end{tabular}}
\end{table*}

\begin{table*}[thb]
\small
\centering
\caption{\label{tbl:examplePP} Changes to the orientation distributions for some infrequent phrase-pairs as a result of applying our LHSmoothed generalization. These infrequent phrase-pairs are used with the correct orientation in the translations by the LHSmoothed model in Table~\ref{tbl:6}. The used orientations are also shown. Note the shifts between the LHSmoothed distributions and the corresponding baseline distributions. The original frequencies are also shown.}
\resizebox{1\linewidth}{!}{
\begin{tabular}{
|c|c|l|l|l|l|}
\hline

\hline
 \multicolumn{1}{|c|}{\zh{寄望}} & \multicolumn{1}{c|}{ in the hope} & \multicolumn{1}{c}{M} & \multicolumn{1}{c}{S} & \multicolumn{1}{c}{DL} & \multicolumn{1}{c|}{DR}  \\
\hline
\multirow{3}{*}{Monotone with previous} & Baseline & 0.10 & 0.01 & 0.11 & 0.78\\
& LHSmoothed & 0.28 & 0.01 & 0.01 & 0.70\\
& Counts in training & 0 & 0 & 0 & 1 \\ 
\hline
 \multicolumn{1}{|c|}{\zh{能 于}} & \multicolumn{1}{c|}{of} & \multicolumn{1}{c}{} & \multicolumn{1}{c}{} & \multicolumn{1}{c}{} & \multicolumn{1}{c|}{}  \\
\hline
\multirow{3}{*}{Discontinuous right with next}& Baseline & 0.10 & 0.01 & 0.12 & 0.77\\
& LHSmoothed & 0.06 & 0.01 & 0.07 & 0.87 \\
& Counts in training & 0 & 0 & 0 & 1\\
\hline 
 \multicolumn{1}{|c|}{\zh{签署 自由 贸易 协定}} & \multicolumn{1}{c|}{signing a free trade agreement} & \multicolumn{1}{c}{} & \multicolumn{1}{c}{} & \multicolumn{1}{c}{} & \multicolumn{1}{c|}{} \\
\hline
\multirow{3}{*}{Discontinuous right with previous}& Baseline & 0.10 & 0.02 & 0.68 & 0.20 \\
& LHSmoothed &  0.21 & 0.03 & 0.30 & 0.46 \\
& Counts in training & 0 & 0 & 1 & 0\\
\hline
\multicolumn{1}{c}{} & \multicolumn{1}{c}{} & \multicolumn{1}{c}{} & \multicolumn{1}{c}{} & \multicolumn{1}{c}{} & \multicolumn{1}{c}{}\\
\hline
\multicolumn{1}{|c|}{\zh{插手}} &  \multicolumn{1}{c|}{meddling in the} & \multicolumn{1}{c}{} & \multicolumn{1}{c}{} & \multicolumn{1}{c}{} & \multicolumn{1}{c|}{}\\
\hline
\multirow{2}{*}{Discontinuous right with previous}& Baseline & 0.10 & 0.01 & 0.68 & 0.21 \\
& LHSmoothed & 0.42 & 0.01 & 0.51 & 0.06  \\
(correctly replaced by the phrase-pair below) & Counts in training & 0 & 0 & 1 & 0\\
\hline
\multicolumn{1}{|c|}{\zh{插手}} &  \multicolumn{1}{c|}{intervention in the}& \multicolumn{1}{c}{} & \multicolumn{1}{c}{} & \multicolumn{1}{c}{} & \multicolumn{1}{c|}{}\\
\hline
\multirow{3}{*}{Monotone with previous} & Baseline & 0.67 & 0.02 & 0.11 & 0.20 \\
& LHSmoothed & 0.75 & 0.01 & 0.03 & 0.21 \\
& Counts in training & 1 & 0 & 0 & 0 \\
\hline
\end{tabular}}
\end{table*}

In general, our dependency-based models change the distributions to a larger degree than our back-off models. In our back-off models, not all long phrase-pairs have eligible sub-phrase-pairs, while the dependency models more often result in shorter generalizations. Table~\ref{tbl:6} provides some examples where our LHSmoothed model has improved the translations by better modeling of reorderings.
To further the understanding of how the reordering distributions of infrequent phrase-pairs, as used in the examples in Table~\ref{tbl:6}, are affected by our models, we show some examples in Table~\ref{tbl:examplePP}. This table lists the examples before and after applying our model.  
As a result of our model, for the phrase-pair (\zh{寄望}, in the hope) the monotone orientation probability is increased although it has a frequency of zero for this orientation. For the next phrase-pair (\zh{能 于}, of) the discontinuous right probability is increased resulting in the correct orientation during translation. The most interesting case is the substantial decrease in the probability of discontinuous left and the increase for discontinuous right for the phrase-pair (\zh{签署 自由 贸易 协定}, signing a free trade agreement), although it has frequency 1 for the former orientation and 0 for the latter. 
In the second example, the shifts in the distributions for the phrase-pairs (\zh{插手}, meddling in the) and (\zh{插手}, intervention in the) caused the first translation option to be dropped and replaced with the second translation option, leading to the correct reordering of the word \textit{western}.
These shifts within the  probability distributions lead to the better translation generated by the LHSmoothed model for the first example in  Table~\ref{tbl:6}.

\section{Conclusions}
\label{sec:conclusion_ch1}

Our reimplementation of the 2POS method by \citet{nagata2006clustered} shows that the full lexical form does not play a significant role in estimating reliable reordering distributions. However, the improvements achieved by keeping important words in their surface form show that the lexical forms for the exposed heads still have an impact on estimating the reordering distributions more precisely. This provides the answer to our \ref{rq:reorderingMain}, which asks:

\begin{itemize}[wide, labelwidth=!, labelindent=0pt ]
\item[]  \textbf{\ref{rq:reorderingMain}} \textit{\acl{rq:reorderingMain}}
\end{itemize}

\noindent Further, we have introduced a novel method that builds on the established idea of backing off to shorter histories, commonly used in language model smoothing, and shown that it can be successfully applied to smoothing of lexicalized and hierarchical reordering models in statistical machine translation. 
%
%
We have also shown that not all sub-phrase-pairs are equally influential in determining the most likely phrase reordering. We experimented with different general forms of the phrase-pairs in which we keep the exposed heads in their original forms and generalize the remaining words by using word classes or simply removing them. The experimental findings show that sub-phrase-pairs consisting of just exposed heads tend to be the most important ones and most other words inside a phrase-pair have negligible influence on reordering behavior. 

\begin{itemize}[wide, labelwidth=!, labelindent=0pt ]
\item[] \textbf{\ref{rq:reorderingSub1}} \textit{\acl{rq:reorderingSub1}}
\end{itemize}

\noindent Earlier approaches, such as \citep{nagata2006clustered} and \citep{cherry:2013:NAACL-HLT}, often assume that the last and the first word of a phrase-pair are the most important words for defining reordering behavior, but our experiments show that exposed heads tend to be stronger predictors. We experiment with both backing off towards the border words by assuming those as the important words, following \citet{nagata2006clustered} and \citet{cherry:2013:NAACL-HLT}, and assuming exposed heads as the most important words, following \citet{chelba2000structured} and \citet{garmash2015bilingual}. We showed that generalized representations of phrase-pairs based on exposed heads can help decrease sparsity and result in more reliable reordering distributions.


\begin{itemize}[wide, labelwidth=!, labelindent=0pt ]
\item[] \textbf{\ref{rq:reorderingSub2}} \textit{\acl{rq:reorderingSub2}}
\end{itemize}

\noindent The results show that backing off to the shorter sub-phrase-pairs that include the ending or the starting word of the phrase helps estimate the reordering distributions. However, our back-off models do not achieve the best results even though they perform better than the baselines. Our generalized models achieve better results in comparison to our back-off models because exposed heads are in general more important for phrase reordering compared to border words.


\begin{itemize}[wide, labelwidth=!, labelindent=0pt ]
\item[] \textbf{\ref{rq:reorderingSub3}} \textit{\acl{rq:reorderingSub3}}
\end{itemize}

\noindent Our generalized representations of phrase-pairs based on exposed heads achieved the best results in our experiments. This is mostly due to the better choice of the most influential internal words rather than the approach we took for the generalization of less important words. However, among the models that generalize phrase-pairs by keeping exposed heads intact as the most influential words, the way less influential words are generalized has a negligible effect on performance. Interestingly, simply removing the less influential words yields the best performance improvements in our experiments. We show that exposed heads and border words do provide good indicators for the reordering behavior of phrase-pairs and that exposed heads are more influential compared to border words on average.


Considering the analysis of the length of infrequent phrase-pairs used during translation, we also conclude that a smoothing model that would be able to further improve the distribution of single word phrase-pairs is crucial for achieving higher improvements during translation.
 
 In this chapter, we showed how the internal words of phrase-pairs impact reordering behavior of phrase-pairs. This is a step towards better understanding how reordering models capture different reordering phenomena. This also helps us in interpreting the behavior of attention models in neural machine translation which is the focus of the next chapter.
 

\graphicspath{ {04-research-02/images/} }

\chapter{What does Attention in Recurrent Neural Machine Translation Pay Attention to?}
\label{chapter:research-02}

\fancyhead{}
\fancyhead[RO]{\sffamily \rightmark}
\fancyhead[LE]{\sffamily \thechapter.\ What does Attention in Recurrent NMT Pay Attention to?}



\section{Introduction}

In this chapter, we shift from phrase-based to neural machine translation. Neural machine translation has gained a lot of attention due to its substantial improvements in machine translation quality over phrase-based machine translation and achieving state-of-the-art performance for many languages~\citep{luong-EtAl:2015:ACL-IJCNLP, jean-EtAl:2015:ACL-IJCNLP, bentivogli-etal-2016-neural, wu2016google, NIPS2017_7181, wang2018multiagent, ott-etal-2018-scaling, chen-etal-2018-best, edunov-etal-2018-understanding, pmlr-v97-wang19f}. The core architecture of neural machine translation models is based on the general encoder-decoder approach~\citep{sutskever2014sequence}. 

In Chapter~\ref{chapter:research-01}, we proposed feature extraction algorithms that identify the words that effect reordering most, in order to improve reordering distributions in phrase-based machine translation. However, neural machine translation model learns to handle reorderings more indirectly. Neural machine translation is an end-to-end approach that learns to encode source sentences into distributed representations and decode these representations into sentences in the target language. Encoding sentence information into distributed representations makes neural machine translation models even harder to interpret, compared to phrase-based machine translation. However, attention-based neural machine translation~\citep{bahdanau-EtAl:2015:ICLR, DBLP_journals_corr_LuongPM15} explicitly identifies the most relevant parts of the source sentence at each translation step. This capability also makes the attentional model superior in translating longer sentences~\citep{bahdanau-EtAl:2015:ICLR, DBLP_journals_corr_LuongPM15}. 

\begin{figure}[thb]
\centering
\includegraphics[scale=0.50, width=0.70\textwidth]{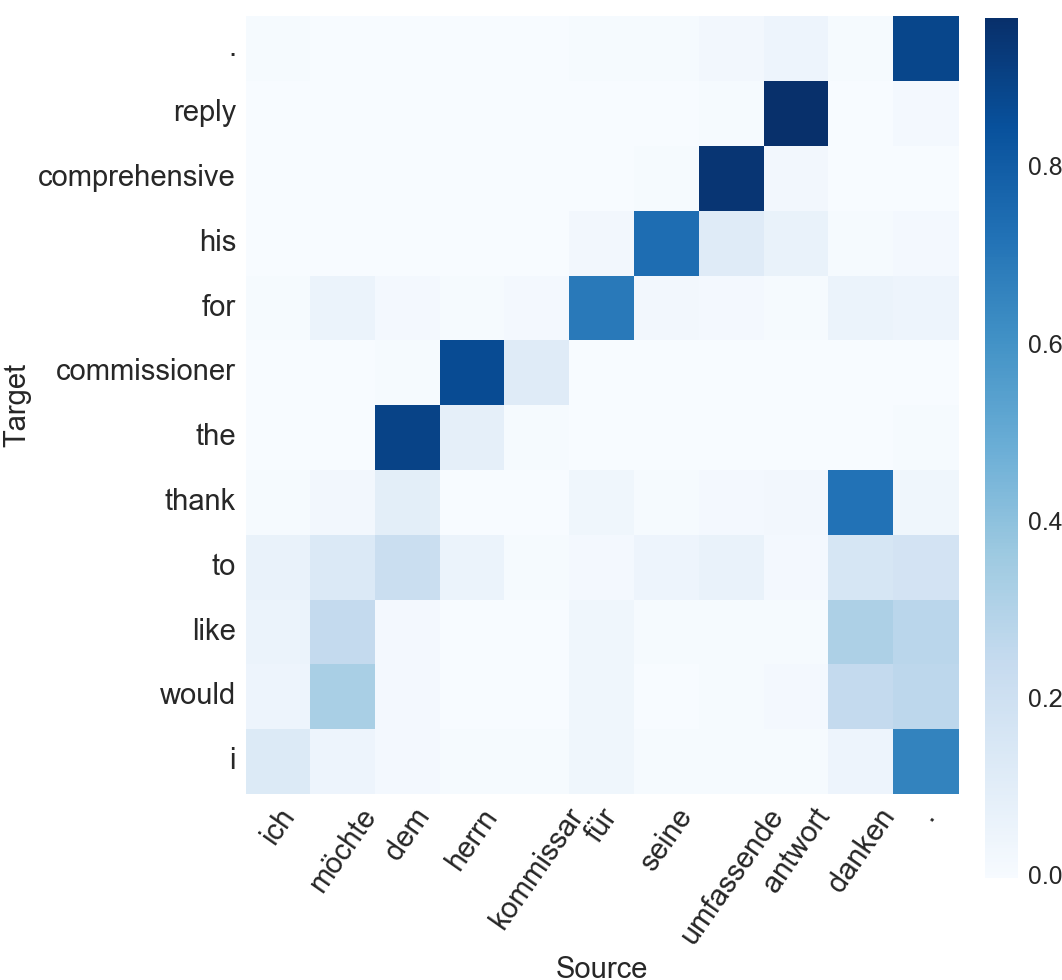}
\caption{Visualization of the attention paid to the relevant parts of the source sentence for each generated word of a translation example. Note how the attention is `smeared out' over multiple source words in the case of ``would" and ``like".}
\label{fig:heatmap}
\end{figure}

In order to understand how neural machine translation captures different syntactic phenomena, especially better word ordering in comparison to phrase-based machine translation, it is important to understand the behavior of the attention model. Due to the capability of attending to the most relevant parts of the source sentence at each translation step, the attention model is often considered similar to reordering models \citep{liu-EtAl:2016:COLING}.

Figure~\ref{fig:heatmap} shows an example of how attention uses the most relevant source words to generate a target word at each step of the translation. In this chapter, we focus on studying the relevance of the attended parts, especially cases where attention is `smeared out' over multiple source words and where their relevance is not entirely obvious, see, e.g., ``would" and ``like" in Figure~\ref{fig:heatmap}. Here, we ask whether this is due to errors or noise of the attention mechanism or is a desired behavior of the model.

Since the introduction of attention models in neural machine translation~\citep{bahdanau-EtAl:2015:ICLR} various modifications such as integrating inductive biases from traditional alignments into attention~\citep{cohn-EtAl:2016:N16-1}, training an attention model to follow traditional alignments~\citep{liu-EtAl:2016:COLING, chen2016guided}, improving the architecture to be simpler and more general~\citep{DBLP_journals_corr_LuongPM15}, and integrating a coverage model into the attention model~\citep{tu-etal-2016-modeling}, have been proposed. However, to the best of our knowledge there is no study that provides an analysis of what kind of phenomena are being captured by attention. There is some work that has looked at attention as being similar to traditional word alignment~\citep{alkhouli-EtAl:2016:WMT, cohn-EtAl:2016:N16-1, liu-EtAl:2016:COLING, chen2016guided}. Some of these approaches also experimented with training the attention model using traditional alignments~\citep{alkhouli-EtAl:2016:WMT, liu-EtAl:2016:COLING, chen2016guided}.~\citet{liu-EtAl:2016:COLING} have shown that attention can be seen as a reordering model as well as an alignment model.

In this chapter, we focus on investigating the differences between attention and alignment and what is being captured by the attention mechanism in general. The general question that we are aiming to answer in this chapter is: 

\begin{itemize}[wide, labelwidth=!, labelindent=0pt ]
\item[] \textbf{\ref{rq:attnMain}} \textit{\acl{rq:attnMain}}
\end{itemize}

\noindent We analyze the attention model behavior using two methods: (i) by looking into its difference with the behavior of traditional alignment and (ii) by studying the distributional behavior of the attention model for different syntactic phenomena. The following subquestions elaborate the analyses we propose in this chapter.

\begin{itemize}[wide, labelwidth=!, labelindent=0pt ]
\item[] \textbf{\ref{rq:attnSub1}} \textit{\acl{rq:attnSub1}}
\end{itemize}

\noindent To answer this question, we first define a metric to compute the difference between the attention distribution and traditional alignment. Then, we study the relation between the quality of translation and the convergence of attention to alignment. We show that the relation differs for different syntactic phenomena, meaning that closer attention to alignments leads to a better translation for some cases, but hurts performance in some other cases. Our analysis suggests that attention and traditional alignment models are different and it is not a good idea to train an attention model to follow the traditional alignments under all circumstance. 

\begin{itemize}[wide, labelwidth=!, labelindent=0pt ]
\item[] \textbf{\ref{rq:attnSub2}} \textit{\acl{rq:attnSub2}}
\end{itemize}

\noindent Here, we analyze whether the differences between the attention model and the traditional alignment model are due to errors in the attention or whether they are side products of the attention model learning meaningful information in addition to what a traditional alignment model learns. We carry out these analyses by studying the distributional behavior of the attention model, especially for cases in which attention deviates from the alignment model. We also study the relation between attention distribution and translation quality to see whether there are meaningful correlations across different syntactic phenomena.

Furthermore, we study the word types that receive the most attention when it is paid to words other than alignment points. This also helps better understand whether the differences between the attention and the alignment models are meaningful.

\begin{itemize}[wide, labelwidth=!, labelindent=0pt ]
\item[] \textbf{\ref{rq:attnSub3}} \textit{\acl{rq:attnSub3}}
\end{itemize}

\noindent In order to achieve a more fine-grained analysis, we also look into the changes in the distributional behavior of the attention model for different syntactic phenomena. We use POS tags as the basis for this analysis since they are established syntactic classes. Additionally, POS tags are simply computed using available tools. We show how the attention behavior changes with respect to the POS tag of the word being generated. We show that this is a key difference between attention and alignment models that causes the attention model to capture additional information compared to traditional alignment models. We show that this difference should not be reversed by training the attention model to closely follow the alignment model.

\begin{itemize}[wide, labelwidth=!, labelindent=0pt ]
\item[] \textbf{\ref{rq:attnSub4}} \textit{\acl{rq:attnSub4}}
\end{itemize}

\noindent In order to understand what additional information the attention model captures compared to the traditional alignment model, it is useful to know what word types are attended to by the attention model based on the type of the generated words. Are the attended words always the same as the aligned words? Or is attention paid to words other than the alignment points. To answer these questions, we look into the average portion of attention that is paid to words other than the alignment points based on the POS tag of the generated word. The observations provide good evidence for our hypothesis that the attention model captures information that the traditional alignment model is unable to capture.
%
%

Our analysis shows that attention follows traditional alignment in some cases more closely while it captures information beyond alignment in others. For instance, attention agrees with traditional alignments to a high degree in the case of nouns. However, in the case of verbs, it captures information beyond just translational equivalence.




This chapter makes the following contributions: 
\begin{enumerate}
\item We provide a detailed comparison of attention in neural machine translation and traditional word alignment. We show that attention and alignment have different behaviors and attention is not necessarily an alignment model.
 
\item We show that while different attention mechanisms can lead to different degrees of compliance with respect to word alignments, full compliance is not always helpful for word prediction.
%
\item We additionally show that attention follows different patterns depending on the type of word being generated. By providing a detailed analysis of the attention model, we contribute to its interpretability. Our analysis explains the mixed results of previous work that has used alignments as an explicit signal to train attention models~\citep{chen2016guided, alkhouli-EtAl:2016:WMT, liu-EtAl:2016:COLING}. 

\item We provide evidence showing that the difference between attention and alignment is due to the capability of the attention model to attend to the context words influencing the current word translation. We show that the attention model attends not only to the translational equivalent of a word being generated, but also to other source words that can affect the form of the target word. For example, while generating a verb on the target side, the attention model regularly attends to the preposition, subject or object of the translational equivalent of the verb on the source side.

\end{enumerate}

\section{Related Work}
\label{sec:relatedWork}

The attention model was an important addition to neural machine translation~\citep{bahdanau-EtAl:2015:ICLR} and it has been frequently used and adapted by other work since. \citet{DBLP_journals_corr_LuongPM15} propose a global and a local attention model. They argue that their global attention model is similar to the original attention model by~\citet{bahdanau-EtAl:2015:ICLR} while having a simpler architecture and requiring less computation. In their local model they predict an alignment point at each step and make the attention focus on a window with the aligned point at the centre. They also propose the input-feeding attention model which takes as input previous attention decisions to better maintain coverage of the source side.

\citet{liu-EtAl:2016:COLING} investigate how training the attention model in a supervised manner can benefit machine translation quality. To this end, they use traditional alignments obtained by running automatic alignment tools (GIZA++~\citep{och2003systematic} and fast\_align~\citep{dyer-chahuneau-smith:2013:NAACL-HLT}) on the training data and feed it as ground truth to the attention network. They report some improvements in translation quality, arguing that the attention model has learned to better align source and target words. Training attention using traditional alignments has also been proposed by others~\citep{mi-etal-2016-supervised, chen2016guided, alkhouli-EtAl:2016:WMT}.

\citet{chen2016guided} show that guiding attention with traditional alignments helps in the domain of e-commerce translation which includes many out-of-vocabulary (OOV) product names and placeholders. On the other hand, they did not report any notable improvements for other domains.

\citet{alkhouli-EtAl:2016:WMT} have separated the alignment model and translation model, reasoning that this avoids propagation of errors from one model to the other as well as providing more flexibility with respect to model types and the training of models. They use a feed-forward neural network as their alignment model that learns to model jumps on the source side using HMM/IBM alignments obtained by running GIZA++. 

\citet{mi-etal-2016-supervised} also propose to train the attention model to minimize a distance metric between attention and alignments given by an automatic tool. They show that separate optimization of the distance and translation objectives leads to a performance drop, although joint optimization of the two objectives achieves small gains. They also experiment with first transforming the traditional alignments by using a Gaussian distribution followed by joint optimization which achieves larger performance improvements. 


\citet{shi-padhi-knight:2016:EMNLP2016} show that various kinds of syntactic information are being learned and encoded in the output hidden states of the encoder. The neural system for their experimental analysis is not an attention-based model and they argue that attention does not have any impact on learning syntactic information.  However, performing the same analysis for morphological information, \citet{belinkov2017neural} show that attention also has some effect on the information that the encoder of a neural machine translation system encodes in its output hidden states. As part of their analysis, they show that a neural machine translation system that has an attention model can learn the POS tags of the source side more efficiently than a system without attention.

\citet{koehn2017six} carry out a simple analysis of the extent to which attention and alignment correlate with each other in different languages by measuring  the probability mass that attention gives to alignments obtained from an automatic alignment tool. They also report differences based on the words that received the highest attention. They perform their experiments for both directions for three language pairs. They observe that close to half of the probability mass from the attention distribution is paid to words other than the alignment points in almost all experiments. However, the German-English experiment appears to be an outlier in their experiments, since the alignment points are one position off from the highest attention points. Although we also use German-English for our experiments in this chapter, we do not observe the same shift.

\citet{tang-etal-2018-analysis} investigate attention behavior when translating ambiguous words. They hypothesize that the attention model spreads out its weights to the context words that help disambiguate the correct sense of the ambiguous words. They carry out their experiments on word sense disambiguation datasets that involve the translation of ambiguous nouns. However, they observe the opposite behavior to their hypothesis. Interestingly, the attention model concentrates more of its weight on the source word while translating ambiguous nouns compared to the cases of translating other types of words. 

 
\section{Attention Models}
\label{sec:attentionModels}

\begin{figure}[thb!]
\centering
\begin{subfigure}[t]{0.70\textwidth}
	\includegraphics[scale=0.20, width=1\linewidth]{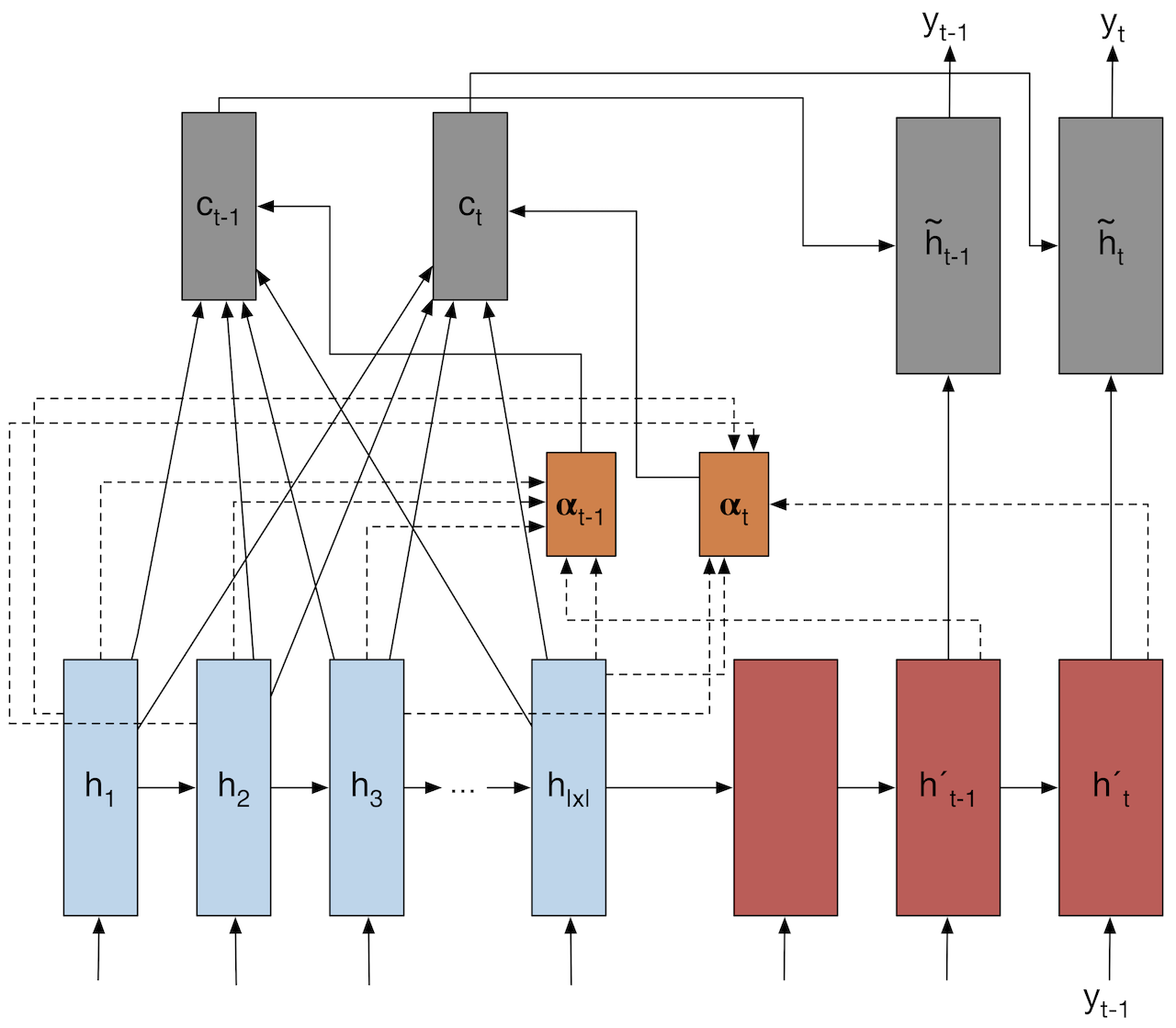}
	\caption{Global attention model}
	\label{fig:attention_non_recurrent}
\end{subfigure}%

\begin{subfigure}[t]{0.70\textwidth}
	\includegraphics[scale=0.20, width=1\linewidth]{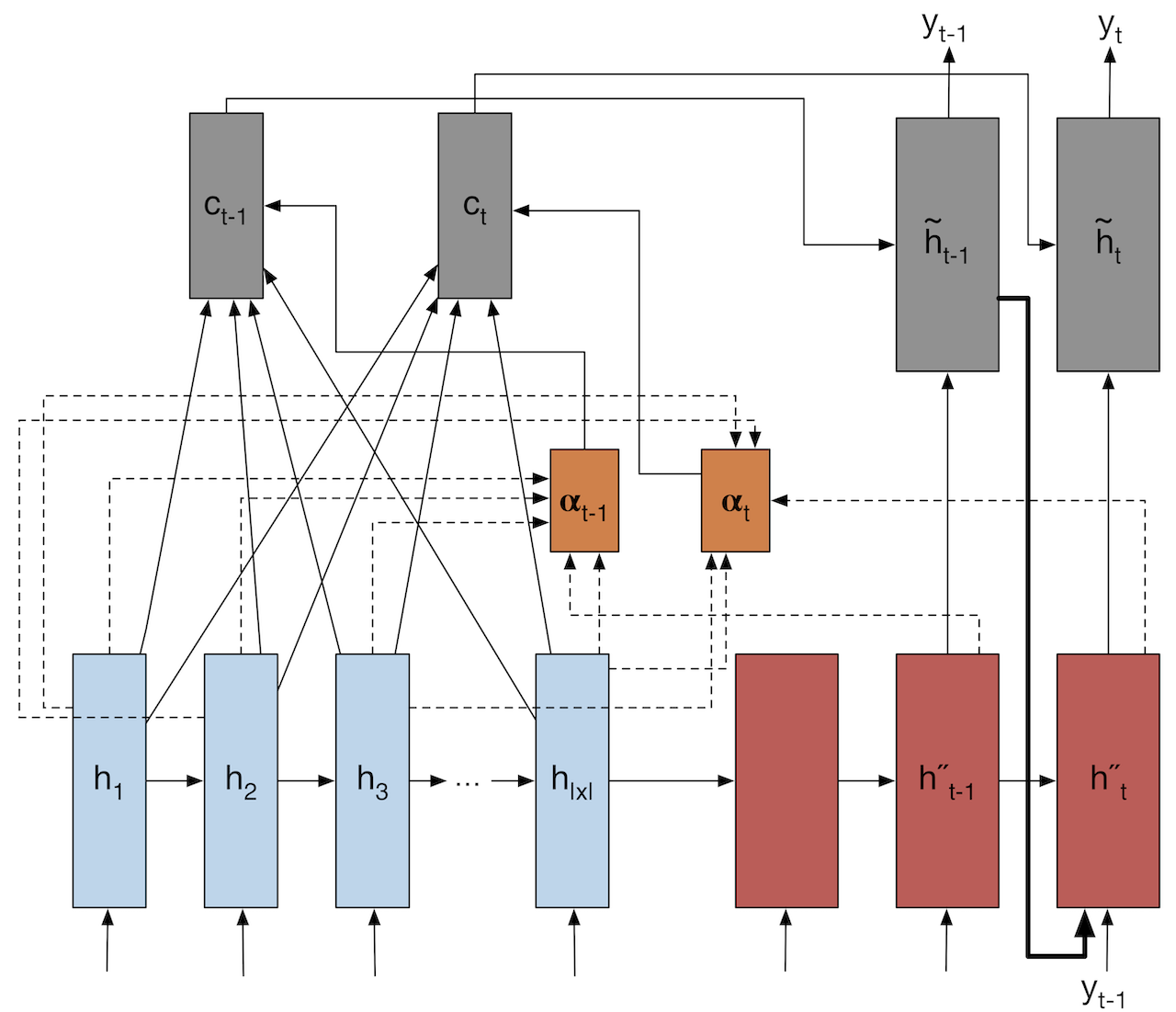}
	\caption{Input-feeding attention model}
	\label{fig:attention_recurrent}
\end{subfigure}
\caption{Comparison of global and input-feeding attention models. These figures show two time steps of decoding. The difference between the two models is shown by the highlighted feedback connection between $\tilde{h}_{t-1}$ and $h^{\prime\prime}_{t}$ in the input-feeding model. This connection feeds the context-dependent output of the decoder from the previous time step back to the decoder as input to the current time step.}
\end{figure}

This section discusses the two popular attention models, which are also used in this chapter. The first model is the non-recurrent attention model, which is equivalent to the ``global attention" method proposed by \citet{DBLP_journals_corr_LuongPM15}. The second attention model is \emph{input-feeding} by \citet{DBLP_journals_corr_LuongPM15}. Figure~\ref{fig:attention_non_recurrent} and \ref{fig:attention_recurrent} show the architecture for global and input-feeding attention models, respectively. The figures are drawn to present the architectures in a way that simply highlights the difference between the two attention models. The bold connection in Figure~\ref{fig:attention_recurrent} highlights the architectural difference between the two models. Below we describe the details of both models.

Both global and input-feeding models compute a context vector $c_i$ at each decoding time step. Subsequently, they concatenate the context vector to the hidden state of the decoder and pass it through a non-linearity before it is fed into the softmax output layer of the translation network:
\begin{equation}
\tilde{h}_{t} = \tanh(W_c[c_t;h^{\prime}_{t}]),
\label{eq:attOutput}
\end{equation}
where  $h^{\prime}_{t}$ and $c_t$ are the hidden state of the decoder and the context vector at time $t$, respectively; $h^{\prime}_{t}$  is defined as follows:
\begin{equation}
\label{eq:global}
h^{\prime}_{t}=f(y_{t-1}, h^{\prime}_{t-1}).
\end{equation}

\noindent Here, $y_{t-1}$ is the last generated target word, $h^{\prime}_{t-1}$ is the previous hidden state of the decoder and $f$ can be an LSTM or GRU recurrent layer.

In the global model, the hidden state of the decoder is compared to each hidden state of the encoder. Often, this comparison is realized as the dot product of vectors, see Equation~\ref{eq:chHiddenComparison}. Then the result is fed to a softmax layer to compute the attention weight distribution, see Equation~\ref{eq:attentionWeight}:
\begin{align}
e_{t,i} &= h^{T}_{i} h^{\prime}_{t} \label{eq:chHiddenComparison}\\
\alpha_{t,i} &= \frac{\mathrm{exp}(e_{t,i})}{\sum_{j=1}^{|x|} \mathrm{exp}(e_{t,j})}\label{eq:attentionWeight}.
\end{align}
\noindent Here $h^{\prime}_{t}$ is the hidden state of the decoder at time $t$, $h_{i}$ is the $i$-th hidden state of the encoder and $|x|$ is the length of the source sentence. The attention weights are used to calculate a weighted sum over the encoder hidden states, which results in the context vector mentioned above:
\begin{equation}
c_i = \sum_{i=1}^{|x|}\alpha_{t,i}h_i.
\end{equation}
\noindent The difference between global and input-feeding models is based on how the context vector is computed. The input-feeding model changes the context vector computation such that at each step $t$ the context vector is aware of the previously computed context $c_{t-1}$. To this end, the input-feeding model feeds back $\tilde{h}_{t-1}$ as input to the network and uses the resulting hidden state instead of the context-independent $h^{\prime}_{t}$. This is defined in the following equations:
\begin{equation}
h^{\prime\prime}_{t}=f(W[\tilde{h}_{t-1};y_{t-1}], h^{\prime\prime}_{t-1}),
\label{eq:stackLSTM}
\end{equation}
where the corresponding hidden state in global attention is defined as in Equation~\ref{eq:global}. Here, $f$ can be an LSTM or GRU recurrent layer, $y_{t-1}$ is the last generated target word, and $\tilde{h}_{t-1}$ is the output of the previous time step of the input-feeding network itself. $e_{t,i}$ for the input-feeding model is defined as follows:
\begin{equation}
e_{t,i} = h_{i}^T h^{\prime\prime}_{t} .
\label{eq:inputFeedingComp}
\end{equation} 

The main difference between the input-feeding and global model is the feed-back in Equation~\ref{eq:stackLSTM}. Global attention is a simpler model that is computationally more efficient than the input-feeding model~\citep{DBLP_journals_corr_LuongPM15}. 

\section{Comparing Attention with Alignment}
As mentioned above, it is a commonly held assumption that attention corresponds to word alignment~\citep{bahdanau-EtAl:2015:ICLR, liu-EtAl:2016:COLING, cohn-EtAl:2016:N16-1, chen2016guided}. To verify this assumption, we investigate whether higher consistency between attention and alignment leads to better translations. To this end, we define three metrics, one to measure the discrepancy between attention and alignment, one to measure attention concentration, and one to measure word prediction loss. Then, we measure the correlation between the two attention metrics and the word prediction loss. 

\begin{table}
\centering
\small
\caption{\label{tbl:RWTHdataStatistics} Statistics of manual alignments provided by RWTH German-English data.}
\begin{threeparttable}
\begin{tabular}{|l|r|}
\hline
& RWTH data\\
\hline
\# sentences & 508\\
\# alignments & 10534 \\
\% sure alignments & 91\%\\
\% possible alignments & 9\%\\
\% 1-to-1 alignments & 74\%\\
\% 1-to-2 alignments & 12\%\\
\% 2-to-1 alignments & 7\%\\
\% 1-to-n\tnote{1} alignments & 5\%\\
\% n\tnote{1}-to-1 alignments & 2\%\\
\hline
\end{tabular}
\begin{tablenotes}
\item[1]where $\text{n} \geq 3$
\end{tablenotes}
\end{threeparttable}
\end{table}

\subsection{Measuring Attention-Alignment Accuracy}

In order to compare attentions of multiple systems as well as to measure the difference between attention and word alignment, we convert the hard word alignments into soft ones, following \citet{mi-etal-2016-supervised}, and use cross entropy between attention and soft alignment as a loss function. For this purpose, we use manual alignments provided by the RWTH German-English dataset~\citep{iwslt06:TP:vilar} as the hard alignments. The statistics of the data are shown in Table~\ref{tbl:RWTHdataStatistics}. Almost 10 percent of the alignments are possible alignments. As explained in Section~\ref{subsec:word_alignment}, for possible alignments human experts are not in consensus.

We convert the hard alignments to soft alignments using Equation~\ref{cs:alignmentDef}. For unaligned words, we first assume that they have been aligned to all words on the source side and then convert them as follows:
\begin{equation}
Al(x_i, y_t) =
\begin{cases}
\frac{1}{|A_{y_t}|} & \mathrm{if}\ x_i \in A_{y_t}\\
0 & \mathrm{otherwise}.
\end{cases}
\label{cs:alignmentDef}
\end{equation}
\noindent Here, $A_{y_t}$ is the set of source words aligned to target word $y_t$ and $|A_{y_t}|$ is the number of source words in the set.

After conversion of the hard alignments to soft ones, we compute the \emph{attention loss} as follows:
\begin{equation}
L_{At}(y_t) = -\sum_{i=1}^{|x|}{Al(x_i,y_t)\log(At(x_i,y_t))}.
\label{eq:attnLoss}
\end{equation}

\noindent Here, $x$ is the source sentence and $Al(x_i,y_t)$ is the weight of the alignment link between source word $x_i$ and the target word $y_t$, see Equation~\ref{cs:alignmentDef}. $At(x_i,y_t)$ is the attention weight $\alpha_{t,i}$ (see Equation~\ref{eq:attentionWeight}) of the source word $x_i$, when generating the target word $y_t$ . 

In our analysis, we also look into the relationship between translation quality and the quality of the attention with respect to the alignments. For measuring the quality of attention, we use the attention loss as defined in Equation~\ref{eq:attnLoss}. As a measure of translation quality, we choose the loss between the output of our NMT system and the reference translation at each translation step, which we call \emph{word prediction loss}. The word prediction loss for word $y_t$ is the logarithm of the probability given in Equation~\ref{eq:NMToutput}. 
\begin{equation}
p_{\mathit{nmt}}(y_t\mid y_{<t}, x) = \mathit{\softmax}(W_o\tilde{h}_{t}).
\label{eq:NMToutput}
\end{equation}

\noindent Here, $x$ is the source sentence, $y_t$ is target word at time step $t$, $y_{<t}$ is the target history given by the reference translation and $\tilde{h}_t$ is given by Equation~\ref{eq:attOutput} for either global attention or input-feeding attention.

\begin{figure}[thb]
\centering
\includegraphics[scale=0.50, width=0.70\textwidth]{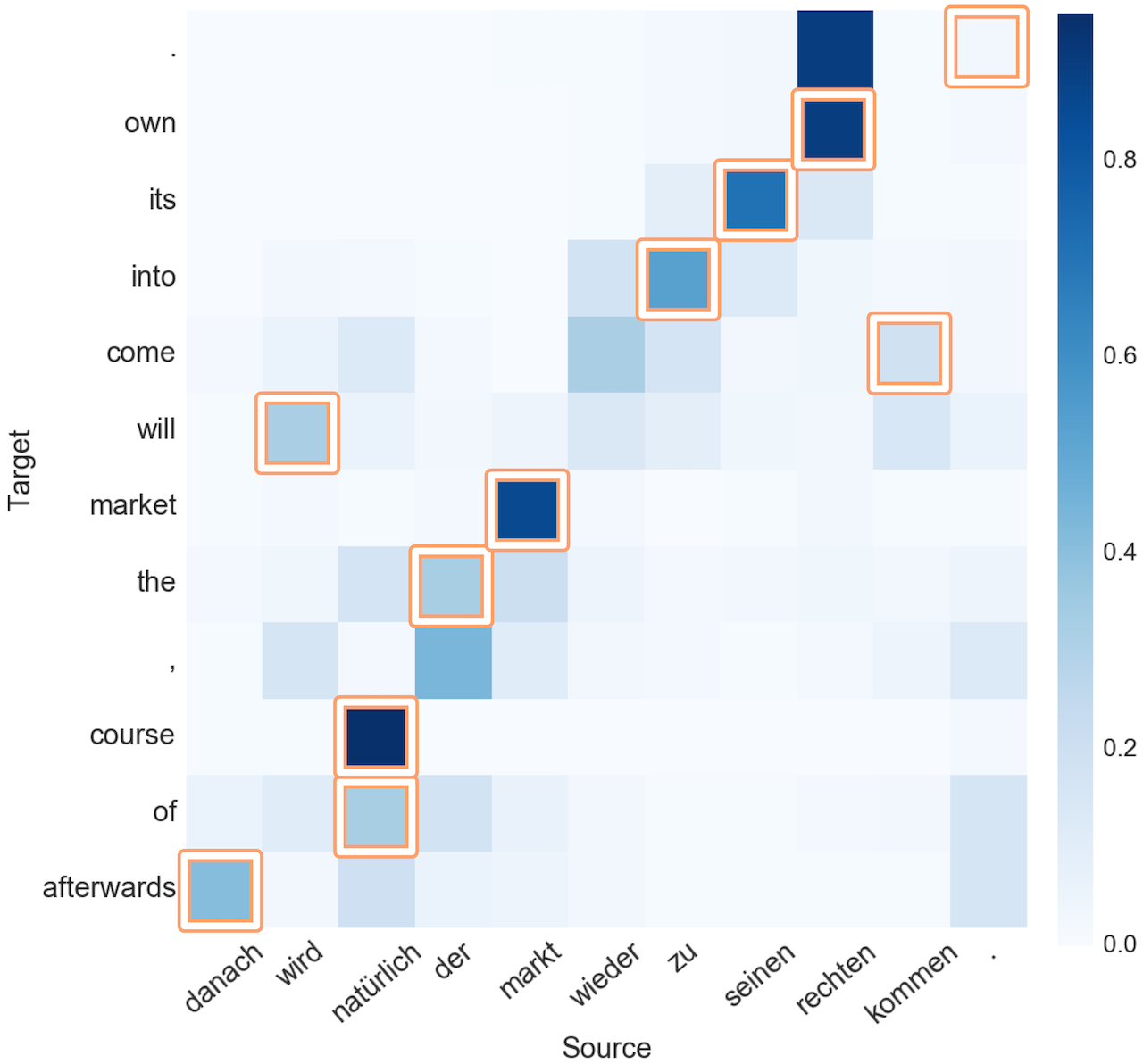}
\caption{An example of inconsistent attention and alignment. The outlined cells show the manual alignments from the RWTH dataset (see Table~\ref{tbl:RWTHdataStatistics}). Note how attention deviates from alignment points in the case of ``will" and ``come".}
\label{fig:attention_alignment}
\end{figure}

Spearman's rank correlation is used to compute the correlation between attention loss and word prediction loss:
\begin{equation}
\rho = \frac{\mathrm{Cov}(R_{L_{At}}, R_{L_{WP}})}{\sigma_{R_{L_{At}}}\sigma_{R_{L_{WP}}}},
\end{equation}
where $R_{L_{At}}$ and $R_{L_{WP}}$ are the ranks of the attention losses and word prediction losses, respectively, $\mathrm{Cov}$ is the covariance between two input variables, and $\sigma_{R_{L_{At}}}$ and $\sigma_{R_{L_{WP}}}$ are the standard deviations of $R_{L_{At}}$ and $R_{L_{WP}}$.

If there is a close relationship between word prediction quality and consistency of attention versus alignment, then there should be a high correlation between word prediction loss and attention loss. Figure~\ref{fig:attention_alignment} shows an example with different levels of consistency between attention and word alignments. For the target words ``will" and ``come" the attention is not focused on the manually aligned word but distributed between the aligned word and other words. The focus of this chapter is to examine cases where attention does not strictly follow alignment, answering the question whether those cases represent errors or are a desirable behavior of the attention model.

\subsection{Measuring Attention Concentration}

As another informative variable in our analysis, we look into the attention concentration. While most word alignments only involve one or a few words, attention can be distributed more freely. We measure the concentration of attention by computing the entropy of the attention distribution:
\begin{equation}
E_{At}(y_t) = -\sum_{i=1}^{|x|}{At(x_i,y_t)\log(At(x_i,y_t))}.
\label{eq:attnEntropy}
\end{equation}
\noindent As in Equation~\ref{eq:attnLoss}, $x$ is the source sentence and $At(x_i,y_t)$ is the attention weight $\alpha_{t,i}$ (see Equation~\ref{eq:attentionWeight}) of the source word $x_i$, when generating the target word $y_t$. 

By measuring attention concentration, we can closely trace shifts in the attention distribution and see for which syntactic phenomena the attention model learns to focus its weight and for which phenomena it learns to spread out the weights to other relevant source words. We also investigate the relationship between attention concentration and translation quality to see whether attention concentration has any direct effect on translation quality for different syntactic phenomena. 


\begin{table}[thb!]
\centering
\small
\caption{\label{tbl:dataStats} Statistics for the parallel corpus used to train our models. The length statistics are based on the source side.}
\begin{tabular}{|l|c|c|c|c|}
\hline
Data & \# of Sent & Min Len & Max Len & Average Len\\
\hline
WMT15 & 4,240,727 & 1 & 100 & 24.7\\
\hline
\end{tabular}
\end{table}

\section{Empirical Analysis of Attention Behavior}
\label{sec:emp_analysis_ch2}

We conduct our analysis using the two attention models described in Section~\ref{sec:attentionModels}. Our first attention model is the global model as introduced by \citet{DBLP_journals_corr_LuongPM15}. The second model is the input-feeding model \citep{DBLP_journals_corr_LuongPM15}, which uses recurrent attention.  Our NMT system is a unidirectional encoder-decoder system with 4 recurrent layers as described in \citet{DBLP_journals_corr_LuongPM15}.

We train the systems with a dimension size of 1,000 and a batch size of 80 sentences for 20 epochs. The vocabulary for both source and target side is set to be the 30K most common words. The learning rate is set to 1 and a maximum gradient norm of 5 has been used.  We also use a dropout rate of 0.3 to avoid overfitting.


\begin{table*}[thb!]
\centering
\caption{\label{tbl:BLEUScores} Performance of our experimental system in BLEU on different standard WMT test sets.}
\begin{tabular}{|l|c|c|c|c|}
\hline
\multicolumn{1}{|l|}{System} & \multicolumn{1}{c|}{test2014} & \multicolumn{1}{c|}{test2015} & \multicolumn{1}{c|}{test2016}  & \multicolumn{1}{c|}{RWTH}\\
\hline
Global & 17.80 & 18.89 & 22.25 & 23.85\\
\hline
Input-feeding & 19.93 & 21.41 & 25.83 & 27.18\\
\hline
\end{tabular}
\end{table*}

\subsection{Impact of Attention Mechanism}
\label{sec:attnMechImpact}

We train both systems on the WMT15 German-to-English training data, see Table~\ref{tbl:dataStats} for statistics of the data. Table~\ref{tbl:BLEUScores} shows the BLEU scores \citep{papineni2002bleu} for both systems on different test sets.

Since we use POS tags and dependency roles in our analysis, both of which are based on words, we chose not to use BPE \citep{sennrichP16-1162}, which operates at the sub-word level.


\begin{table}[thb]
\centering
\small
\caption{\label{tbl:alignmentErrorRate} Alignment error rate (AER) of the hard alignments produced from the output attentions of the systems with input-feeding and global attention models. We use the most attended source word for each target word as the aligned word. The last column shows the AER for the alignment generated by GIZA++.}
\begin{tabular}{|l|c|c|c|}
\hline
 & Global & Input-feeding & GIZA++ \\
 \hline
 AER & 0.60 & 0.37 & 0.31 \\
 \hline
\end{tabular}
\end{table}

\begin{table}[thb]
\centering
\small
\caption{\label{tbl:averageAttnVsAlignLoss} Average loss between attention generated by input-feeding and global systems and the manual alignment over RWTH German-English data.}
\begin{tabular}{|l|c|c|}
\hline
 & Global & Input-feeding  \\
 \hline
 \makecell{Attention loss} & 0.46 & 0.25 \\
 \hline
\end{tabular}
\end{table}




To compute AER over attentions, we follow \citet{DBLP_journals_corr_LuongPM15} to produce hard alignments from attention distributions by choosing the most attended source word for each target word. We also use GIZA++ \citep{och2003systematic} to produce automatic alignments over the data set to allow for a comparison between automatically generated alignments and the attentions generated by our systems. GIZA++ is run in both directions and alignments are symmetrized using the grow-diag-final-and refined alignment heuristic~ \citep{och2003systematic}.


As shown in Table~\ref{tbl:alignmentErrorRate}, the input-feeding system not only achieves a higher BLEU score, but also uses attentions that are closer to the human alignments.


Table~\ref{tbl:averageAttnVsAlignLoss} compares input-feeding and global attention in terms of attention loss computed using Equation~\ref{eq:attnLoss}. Here, the losses between the attentions produced by each system and the human alignments is reported. As expected, 
the difference in attention loss is in line with AER.


The difference between these comparisons is that AER only takes the most attended word into account while attention loss considers the entire attention distribution. 


\begin{table}[thb]
\centering
\caption{List of the universal POS tags used in our analysis.}
\label{tbl:POStags}
\begin{tabular}{|l|l|l|}
\hline
\multicolumn{1}{|l|}{Tag} & \multicolumn{1}{l|}{Meaning} & \multicolumn{1}{l|}{Example}\\
\hline
ADJ & Adjective & large, latest \\
ADP & Adposition & in, on, of \\
ADV & Adverb & only, whenever\\
CONJ & Conjunction & and, or\\
DET & Determiner & the, a\\
NOUN & Noun & market, system\\
NUM & Numeral & 2, two\\
PRT & Particle & 's, off, up\\
PRON & Pronoun & she, they\\
PUNC & Punctuation & ;, .\\
VERB & Verb & come, including\\
\hline
\end{tabular}
\end{table}

%

\subsection{Alignment Quality Impact on Translation}

Based on the results in Section~\ref{sec:attnMechImpact}, one might be inclined to conclude that the closer the attention is to the word alignments, the better the translation. However, earlier research  \citep{chen2016guided, alkhouli-EtAl:2016:WMT, liu-EtAl:2016:COLING} reports mixed results by optimizing their NMT system with respect to word prediction and alignment quality. These findings warrant a more fine-grained analysis of attention. To this end, we include POS tags in our analysis and study the patterns of attention based on the POS tags of the target words. We choose POS tags because they exhibit simple syntactic characteristics. We use the coarse grained universal POS tags given in Table~\ref{tbl:POStags} \citep{petrov2012universal}.

\begin{figure}[thb!]
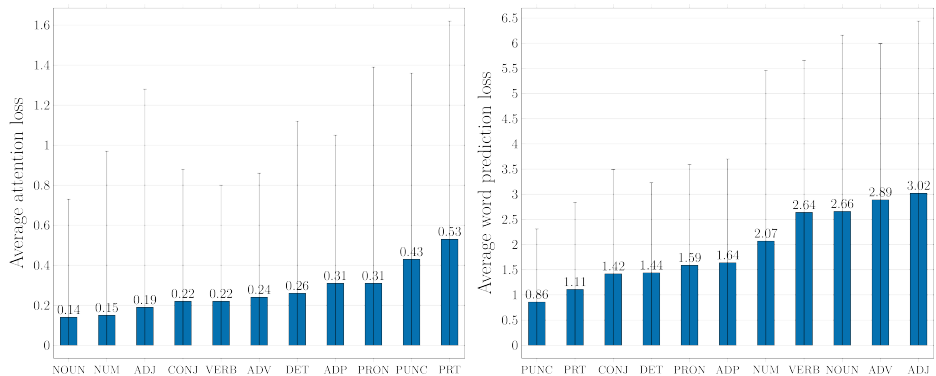

\centering
\begin{subfigure}[t]{0.5\textwidth}
	\centering
	\includegraphics[scale=1, width=1\linewidth]{attenLoss}
	\caption{Average attention loss based on the POS tags of the target side.}
	\label{fig:attentionLoss}
\end{subfigure}%
~
\begin{subfigure}[t]{0.5\textwidth}
	\centering
	\includegraphics[scale=1, width=1\linewidth]{ModelScore}
	\caption{Average word prediction loss based on the POS tags of the target side.}
	\label{fig:transLoss}
\end{subfigure}
\caption{Average attention losses and word prediction losses from the input-feeding system.}
\end{figure}

To better understand how attention accuracy affects translation quality, we analyse the relationship between attention loss and word prediction loss for individual POS classes. Figure~\ref{fig:attentionLoss} shows how attention loss differs when generating different POS tags. One can see that  attention loss varies substantially across different POS tags. In particular, we focus on the cases of NOUN and VERB which are the most frequent POS tags in the dataset. As shown, the attention of NOUN is the closest to alignments on average. But the average attention loss for VERB is almost two times larger than the loss for NOUN. 

Considering this difference and the observations in Section~\ref{sec:attnMechImpact}, a natural follow-up would be to focus on forcing the attention of verbs to be closer to alignments. However, Figure~\ref{fig:transLoss} shows that the average word prediction loss for verbs is actually smaller compared to the average loss for nouns. In other words, although the attention for verbs is substantially more inconsistent with the word alignments than for nouns, on average, the NMT system translates verbs more accurately than nouns.

\begin{figure}[thb!]
\centering
\includegraphics[scale=0.50, width=0.5\textwidth]{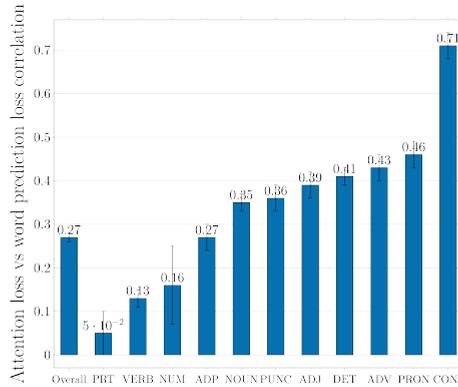}
\caption{Correlation between word prediction loss and attention loss for the input-feeding model.}
\label{fig:transLossVsAlignLoss}
\end{figure}

To formalize this relationship we compute Spearman's rank correlation between word prediction loss and attention loss, based on the POS tags of the target side which is shown in~Figure~\ref{fig:transLossVsAlignLoss} for the input-feeding model.

The low correlation for verbs confirms that attention to other parts of the source sentence rather than the aligned word is necessary for translating verbs correctly and that attention does not necessarily have to follow alignments. However, the higher correlation for nouns means that consistency of attention with alignments is more desirable. This can partly explain the mixed results reported for training attention using alignments \citep{chen2016guided,liu-EtAl:2016:COLING, alkhouli-EtAl:2016:WMT}. This is especially apparent in the results by \citet{chen2016guided}, where large improvements are achieved for the e-commerce domain, which contains many OOV product names and placeholders, but no or only very small improvements are achieved for general domains.  




\begin{figure}[thb!]
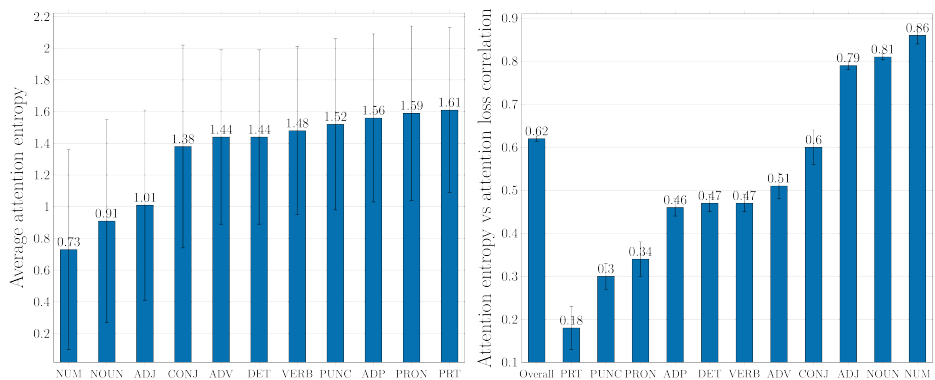

\centering
\begin{subfigure}[t]{0.50\textwidth}
	\includegraphics[scale=0.12, width=1\linewidth]{AttentionEntropy}
	\caption{Average attention entropy based on the POS tags.}
	\label{fig:attnEntropy_a}
\end{subfigure}%
~
\begin{subfigure}[t]{0.50\textwidth}
	\includegraphics[scale=0.12, width=1\linewidth]{AttnEntVsAlignLossCorrelation}
	\caption{Correlation between attention entropy and attention loss.}
	\label{fig:attnEntropy_b}
\end{subfigure}
\caption{Attention entropy and its correlation with attention loss for the input-feeding model.}
\label{fig:AttnEntropyAndCorrelation}
\end{figure}

\subsection{Attention Concentration}
\label{subse:attention_con_ch2}

In word alignment, most target words are aligned to one source word~\citep{graca-etal-2010-learning}, see Table~\ref{tbl:dataStats}. The average number of source words aligned to nouns and verbs is 1.1 and 1.2, respectively. To investigate to what extent this also holds for attention, we measure the attention concentration by computing the entropy of the attention distribution, see Equation~\ref{eq:attnEntropy}.

Figure~\ref{fig:attnEntropy_a} shows the average entropy of attention based on POS tags. As one can see, nouns have one of the lowest entropies, meaning that on average, the attention for nouns tends to be concentrated. This also explains the closeness of the attention to alignments for nouns. In addition, the correlation between attention entropy and attention loss in the case of nouns is high, as shown in Figure~\ref{fig:attnEntropy_b}. This means that attention entropy can be used as a measure of closeness of attention to alignment in the case of nouns.

\begin{figure}[thb!]
\centering
\includegraphics[scale=0.50, width=0.50\textwidth]{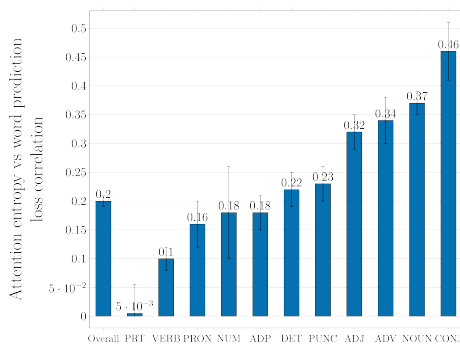}
\caption{Correlation of attention entropy and word prediction loss for the input-feeding model.}
\label{fig:transLossVsattnEntropy}
\end{figure}

The higher attention entropy for verbs, see Figure~\ref{fig:attnEntropy_a}, shows that the attention is more distributed than for nouns. The low correlation between attention entropy and word prediction loss, see Figure~\ref{fig:transLossVsattnEntropy}, shows that attention concentration is not required when translating into verbs. This also confirms that the correct translation of verbs requires the systems to pay attention to different parts of the source sentence.

Another interesting observation is the low correlation for pronouns (PRON) and particles (PRT), see Figure~\ref{fig:transLossVsattnEntropy}. For example, as can be seen in Figure~\ref{fig:attnEntropy_a}, these tags have a wider attention distribution compared to nouns. This could either mean that the attention model does not know where to focus or that it deliberately pays attention to multiple relevant places to be able to produce a better translation. The latter is supported by the relatively low word prediction losses, as shown in Figure~\ref{fig:transLoss}.

\renewcommand{\arraystretch}{1}
\begin{table}[tbh]
\centering
\small
\caption{Distribution of attention probability mass (in \%) paid to alignment points and the mass paid to words other than alignment points for each POS tag.}
\label{tbl:attnPercentage}
\begin{tabular}{|l|c|c|}
\hline
POS tag & \makecell{attention to\\ alignment points \%} & \makecell{attention to\\ other words \%}\\
\hline
NUM & 73 & 27\\
NOUN & 68 & 32\\
ADJ & 66 & 34\\
PUNC & 55 & 45\\
ADV & 50 & 50\\
CONJ & 50 & 50\\
VERB & 49 & 51\\
ADP & 47 & 53\\
DET & 45 & 55\\
PRON & 45 & 55\\
PRT & 36 & 64\\
\hline
Overall & 54 & 46\\
\hline
\end{tabular}
\end{table}
\renewcommand{\arraystretch}{1}

\subsection{Attention Distribution}
\label{subsec:Attn_dist_ch2}

To further understand under which conditions attention is paid to words other than the aligned words, we study the distribution of attention over the source words. First, we measure how much attention, on average, is paid to the aligned words for each POS tag. To this end, we compute the percentage of the probability mass that the attention model has assigned to aligned words for each POS tag, see Table~\ref{tbl:attnPercentage}.

\setcounter{footnote}{0}

\renewcommand{\arraystretch}{1}
\begin{table*}[thb]
\centering
\begin{threeparttable}
\caption{The most attended dependency roles with their received attention percentage from the attention probability mass paid to words other than the alignment points. Here, we focus on the POS tags discussed earlier.}
\label{tbl:dependencyRoles}
\begin{tabular}{|l|l|l|}
\hline
POS tag & roles(attention \%) & description\\
\hline
\multirow{4}{*}{NOUN} & punc(16\%) & Punctuations\tnote{1}\\
& pn(12\%) & Prepositional complements\\
& attr(10\%) & Attributive adjectives or numbers\\
& det(10\%) & Determiners\\
\hline
\multirow{5}{*}{VERB} & adv(16\%) & Adverbial functions including negation\\
& punc(14\%) & Punctuations\\
& aux(9\%) & Auxiliary verbs\\
& obj(9\%) & Objects\tnote{2}\\
& subj(9\%) & Subjects\\
\hline
\multirow{3}{*}{CONJ} & punc(28\%) & Punctuations\\
& adv(11\%) & Adverbial functions including negation\\
& conj(10\%) & All members in a coordination\tnote{3}\\
\hline
\end{tabular}
\begin{tablenotes}
\item[1] Punctuations have the role ``root" in the parse generated using ParZu. However, we use the POS tag to differentiate them from other tokens having the role ``root".
\item[2] Attention mass for all different objects is summed up.
\item[3] Includes all different types of conjunctions and conjoined elements.
\end{tablenotes}
\end{threeparttable}
\end{table*}
\renewcommand{\arraystretch}{1}

\begin{figure}[thb]
\centering
\includegraphics[scale=0.55, width=0.80\textwidth]{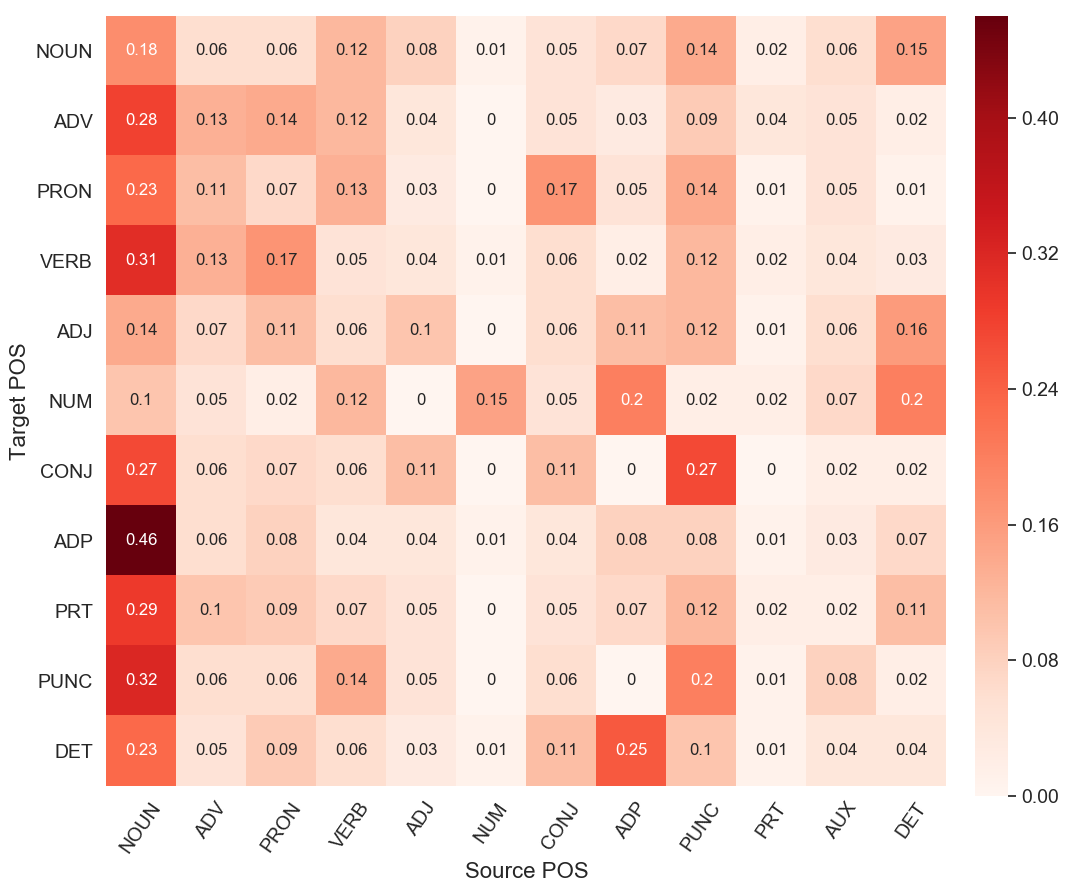}
\caption{The distribution of the most attended words that are not consistent with the alignment points. }
\label{fig:error_dist}
\end{figure}

 It is interesting that numbers, nouns and adjectives are ranked at the top of the table with most of the attention paid to the alignment points. Intuitively, little attention has to be paid to other context words while translating these types of words. These are cases that benefit when the attention model is trained by explicit signals from traditional alignments. The higher gain in the e-commerce data compared to other domains reported by \citet{chen2016guided} also confirms this. However, one can notice that less than half of the attention is paid to alignment points for most of the POS tags. To examine how attention has been distributed for the remaining cases, we measure the attention distribution over dependency roles on the source side. We first parse the source side of the RWTH data using the ParZu parser \citep{sennrich-volk-schneider:2013:RANLP-2013}. Then, we compute how the attention probability mass assigned to words other than the alignment points is distributed over dependency roles. Table~\ref{tbl:dependencyRoles} lists the most attended roles for each POS tag. Here, we focus on the POS tags discussed earlier. One can see that the most attended roles when translating to nouns include adjectives and determiners and, in the case of translating to verbs, they include auxiliary verbs, adverbs (including negation), subjects, and objects. 

\begin{figure}
\centering
\includegraphics[scale=0.55, width=0.8\textwidth]{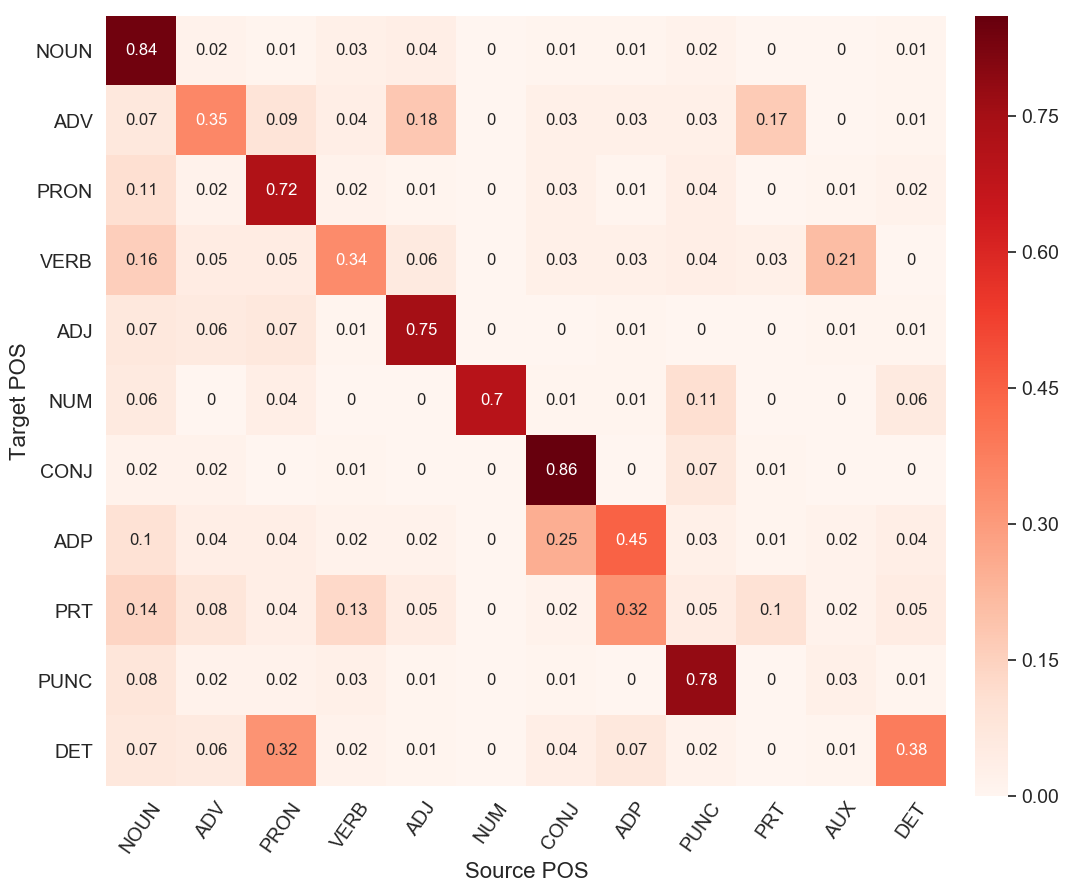}
\caption{The distribution of the most attended words that are consistent with the alignment points.}
\label{fig:consistent_dist}
\end{figure}

\section{Attention as Hard Alignment}

To further analyze how attention is paid to words other than the alignment points, we look to the cases where the most attended word on the source side is not the alignment point of the generated word of the target side. In other words, we convert (soft) attention to hard alignments by taking the most attended source word for each target word as the alignment point of the target word. Next, we compare the resulting alignments to the manual gold alignments. Almost half of the cases of the resulting alignments are not consistent with the gold alignments. Figure~\ref{fig:error_dist} shows how these alignments are distributed based on the POS tags on the source and target side. Note that the numbers in this figure are not attention weights. These are the average portions of cases where the highest attended word has the POS tag given on the horizontal axis while generating a word with the POS tag on the vertical axis.

A number of interesting observations emerge from Figure~\ref{fig:error_dist}. Consider the first row that corresponds to generating nouns on the target side. Interestingly, the most attended cases are almost uniformly distributed over different POS tags. This indicates that these are the cases for which the attention has not properly learned what to attend to. For example, it is clear from the first cell from the top left that in 18 percent of the cases the attention model has put most mass on nouns while generating nouns on the target side. However, considering that these are the cases in which attention is inconsistent with the alignment, it means that the attention model has devoted its highest weight to another noun rather than the translational equivalent of the generated target word. These are the cases that benefit most from an explicit training signal based on traditional alignments.

Another interesting observation relates to verbs. As can be seen in the corresponding row of the target VERB POS tag, in 31, 13 and 17 percent of the cases the highest attended words on the source side are nouns, adverbs and pronouns, respectively. The attended nouns and pronouns are likely to be the subjects and the objects of the verbs. This is consistent with the results reported in Table~\ref{tbl:dependencyRoles}, which shows the attention weights with respect to the dependency roles.

The fact that in almost all cases, the highest attended words on the source side are nouns may warrant further investigation in future research. However, this may also be a side product of the distribution of the word types, since nouns are more frequent than other POS types.

As well as investigating attention-based hard alignments inconsistent with the gold alignments, we also investigated those that are consistent. Figure~\ref{fig:consistent_dist} shows the distribution of the cases where the most attended word is the same as the alignment point of the generated target word.

The focus of weights on the same POS tags in the case of nouns, numbers and adjectives reflects that, in most cases, the translation of these words does not require paying attention to the context. These are also the cases that assign most attention to the alignment points, as shown in Table~\ref{tbl:attnPercentage}.

Figure~\ref{fig:consistent_dist} shows that for verbs attention is not focused. This is because our POS tagger tools separate auxiliary verbs and main verbs on the source side and tag them as only verbs on the target side. However, this is different for verbs in Figure~\ref{fig:error_dist}, which contains inconsistent attention to alignment points.


\section{Conclusion}

In this chapter, we have studied attention in neural machine translation and provided an analysis of the relationship between attention and word alignment. We have used pre-existing and new metrics to analyze the behavior of attention across different syntactic phenomena. Our research in this chapter answers the following sub-research questions:

\begin{itemize}[wide, labelwidth=!, labelindent=0pt ]
\item[] \textbf{\ref{rq:attnSub1}} \textit{\acl{rq:attnSub1}}
\end{itemize}

\noindent We first checked the extent of agreement between two different attention models and traditional word alignments. We did this by measuring alignment error rate (AER) of global~\citep{DBLP_journals_corr_LuongPM15} and input-feeding~\citep{bahdanau-EtAl:2015:ICLR} attention models and GIZA++, a frequently used automatic alignment tool, against word alignments produced by human annotators. 
 
We have shown that attention agrees with traditional alignments to a limited degree. However, the level of agreement differs substantially by attention mechanism and the type of the word being generated. When generating nouns, attention mostly focuses on the translational equivalent of the target word on the source side. Here, attention behaves very similar to traditional alignments. However, its behavior diverges from that of the traditional alignment model when it comes to generating words that need contextual information to be correctly generated. This is especially the case for verbs that need to be in agreement with their subject or are affected by their auxiliary verb.


Next, we investigate whether the divergence from traditional alignment is due to errors in the attention model or whether it is evidence that attention captures more information than traditional alignments. We initiate this investigation with the following question:

\begin{itemize}[wide, labelwidth=!, labelindent=0pt ]
\item[] \textbf{\ref{rq:attnSub2}} \textit{\acl{rq:attnSub2}}
\end{itemize}

\noindent To answer this question, we ensured that the divergence of attention from traditional alignments is not due to errors made by the attention model. To this end, we examine the correlation between translation quality and attention loss as well as attention concentration. The low correlations between these quantities, especially when generating verbs, show that the difference is not due to errors, but a side effect of capturing additional information compared to traditional alignments.

The concentration of attention when generating nouns and adjectives shows that for these syntactic phenomena, attention captures mostly translational equivalence, very similar to traditional alignments. However, while generating verbs more weight is assigned to other context words.

\begin{itemize}[wide, labelwidth=!, labelindent=0pt ]
\item[] \textbf{\ref{rq:attnSub3}} \textit{\acl{rq:attnSub3}}
\end{itemize}

\noindent We have shown that attention has different patterns based on the POS tag of the target word. The concentrated pattern of attention and the relatively high correlations for nouns show that training attention with explicit alignment labels is useful for generating nouns. However, this is not the case for verbs, since a large portion of attention is paid to words other than the alignment points capturing other relevant information. In this case, training attention with alignments will force the attention model to forget this useful information. This explains the mixed results reported when guiding attention to comply with alignments \citep{chen2016guided, liu-EtAl:2016:COLING, alkhouli-EtAl:2016:WMT}.

\begin{itemize}[wide, labelwidth=!, labelindent=0pt ]
\item[] \textbf{\ref{rq:attnSub4}} \textit{\acl{rq:attnSub4}}
\end{itemize}

\noindent To answer this question, we investigated the attention weights that are distributed to the source words other than the alignment point or points of the generated target word. We studied the distribution of these weights with respect to the dependency relations of the alignment points.

We have shown that the attention model learns to assign some attention weights to the dependency relations of a word while translating it. This is specially true in the case of words that need additional context to be translated correctly. For example, verbs, adverbs, conjunctions and particles only receive half of the attention weight while being translated. The rest of the attention is assigned to their dependent words that affect their translations in the target language.

This allows us to answer the overall research question of this chapter:

\begin{itemize}[wide, labelwidth=!, labelindent=0pt ]
\item[] \textbf{\ref{rq:attnMain}} \textit{\acl{rq:attnMain}}
\end{itemize}

\noindent The distributional behavior of the attention model shows that the type of relevant information changes with the syntactic category of the word being generated. We can conclude that for nouns and adjectives, translational equivalents or alignment points are the most important words on the source side. However, for verbs, pronouns, particles, conjunctions and generally most of the other syntactic categories, contextual dependencies are as important as alignment points.
  
In this chapter, we have contributed to interpretability of an attention model by showing how attention behavior changes for different syntactic phenomena and what the important words for attention are. We have also shown that an attention model cannot interpreted as an alignment model.

In the next chapter, we investigate what information is contained in the encoder hidden states which attention models use to compute the context vectors.
\graphicspath{ {04-research-03/images/} }

\chapter{Interpreting Hidden Encoder States}
\label{chapter:research-03}

%


\section{Introduction}

\fancyhead{}
\fancyhead[RO]{\sffamily \rightmark}
\fancyhead[LE]{\sffamily \leftmark}

It is straightforward to train an NMT system in an end-to-end fashion. This has been made possible by an encoder-decoder architecture that encodes the source sentence into a distributed representation and then decodes this representation into a sentence in the target language. While earlier work has investigated what information is captured by the attention mechanism \citep{belinkov2017neural, koehn2017six, tang-etal-2018-analysis, voita-etal-2019-analyzing, moradi-etal-2019-interrogating} and the hidden state representations of an NMT system~\citep{shi-padhi-knight:2016:EMNLP2016,belinkov2017neural,bisazza-tump-2018-lazy,xu2020analyzing}, more analyses are still required to better interpret what linguistic information from the source sentence is captured by the encoders' hidden distributed representations. To this end, the primary research question that we seek to answer in this chapter is as follows:

\begin{itemize}[wide, labelwidth=!, labelindent=0pt ]
\item[] \textbf{\ref{rq:hiddenStateMain}} \textit{\acl{rq:hiddenStateMain}}
\end{itemize}

\noindent In the previous chapter, we investigated the attention model, which computes a weighted sum over the encoder hidden states at each decoding time step. The attention model decides which source words to attend to or ignore at each step. However, this interpretation of the attention model has the underlying assumption that each encoder hidden state represents the corresponding source word. In this chapter, we look deeper inside the hidden states to study what information from the source side is encoded in hidden states. In doing so, we investigate whether the assumption that encoder hidden states represent their underlying tokens is valid. 

Recently, some attempts have been made to shed some light on the information that is being encoded in the intermediate distributed representations \citep{shi-padhi-knight:2016:EMNLP2016,belinkov2017neural}. Feeding the hidden states of the encoder of different sequence-to-sequence systems, including multiple NMT systems, as the input to different classifiers, \citet{shi-padhi-knight:2016:EMNLP2016} aim to investigate what syntactic information is encoded in the hidden states. They provide evidence that syntactic information such as the voice and tense of a sentence and the part-of-speech (POS) tags of words are being learned with reasonable accuracy. They also provide evidence that more complex syntactic information such as the parse tree of a sentence is also learned, but with lower accuracy. 

\citet{belinkov2017neural} follow the same approach as \citet{shi-padhi-knight:2016:EMNLP2016} to conduct more analyses about how syntactic and morphological information is encoded in the hidden states of the encoder. They carry out experiments for POS tagging and morphological tagging. They study the effect of different word representations and different layers of the encoder on the accuracy of their classifiers to reveal the impact of these variables on the amount of syntactic information captured by the hidden states.

Despite the approaches discussed above, attempts to study the hidden states more intrinsically are still missing. For example, to the best of our knowledge, there is no work that studies the encoder hidden states from a nearest neighbor perspective to compare these distributed word representations with the underlying word embeddings. It seems intuitive to assume that the hidden state of the encoder corresponding to an input word conveys more contextual information compared to the embedding of the input word itself. But what type of information is captured and how does it differ from the word embeddings? Answering this question will help us understand what information is captured by the hidden states and find the answer to our general research question \ref{rq:hiddenStateMain}. We start by pursuing a  more fine-grained research question:

\begin{itemize}[wide, labelwidth=!, labelindent=0pt ]
\item[] \textbf{\ref{rq:hiddenStateSub1}} \textit{\acl{rq:hiddenStateSub1}}
\end{itemize}

\noindent We investigate this research question by looking into the nearest neighbors of the hidden states and compare them with the nearest neighbors of the corresponding word embeddings. Understanding the similarities and differences between the lists of the nearest neighbors of embeddings and hidden states facilitate human interpretation of the encoder hidden state properties. 

As mentioned earlier, we intuitively expect the hidden states to capture more contextual information than word embeddings. Additionally, we already know that word embeddings capture a general representation of all possible senses of a word and the most frequent sense of a word in the training data is dominating the embedding representation \citep{iacobacci-etal-2015-sensembed, faruqui-etal-2016-problems}. Observing that neural machine translation systems successfully translate most of the source words with their correct sense in specific contexts, we would expect that hidden states can capture the sense of the words in their context. We further investigate this in the next research question (\ref{rq:hiddenStateSub2}). 

\begin{itemize}[wide, labelwidth=!, labelindent=0pt ]
\item[] \textbf{\ref{rq:hiddenStateSub2}} \textit{\acl{rq:hiddenStateSub2}}
\end{itemize}

\noindent Additionally, we use the similarities and differences between the embeddings and their corresponding hidden states to compare the recurrent neural network model and the transformer model. We compare how much the hidden states of these two types of models contain similar information with their corresponding embeddings. This can help us to understand how much of the capacity of the hidden states in these two models is devoted to other types of information. 

Comparing the nearest neighbors list of the word embeddings and the list of the nearest neighbors of their corresponding hidden states we see that there are some words appearing in both and some that only appear in one of the lists. In our investigations to answer research question \ref{rq:hiddenStateSub2}, we focus on those words that appear in the list of the nearest neighbors of the hidden states and not in the list of the neighbors of the corresponding word embeddings. 

\citet{shi-padhi-knight:2016:EMNLP2016} and \citet{belinkov2017neural} have already shown that the hidden states capture syntactic, morphological and semantic information by feeding the hidden state into diagnostic classifiers. However, we aim to investigate how the captured syntactic and lexical semantic information is reflected in the nearest neighbors list. Knowing this is important as it makes the information captured by the hidden states more interpretable. Additionally, we seek to achieve an estimate of the capacity of the hidden states that has been devoted to capture this information. We also investigate the changes made to these capacities throughout the sentences. This is especially interesting in the case of recurrent models since they have to devote more capacity to capturing context information for the last hidden states in each direction when encoding long sentences. Later on, we can use these estimates for a comparison of the functionality of the hidden states in different neural architectures. This is addressed in the final sub research question: 

\begin{itemize}[wide, labelwidth=!, labelindent=0pt ]
\item[] \textbf{\ref{rq:hiddenStateSub3}} \textit{\acl{rq:hiddenStateSub3}}
\end{itemize}

\noindent In addition to work that has investigated the captured information in hidden states by the use of extrinsic tasks of syntactic and semantic classifications, there are recent approaches that compare different state-of-the-art encoder-decoder architectures in terms of their capabilities to capture syntactic structures \citep{D18_Recurrent} and lexical semantics \citep{D18-1458_Self-Attention}. This work also uses extrinsic tasks for their comparisons. \citet{D18_Recurrent} use subject-verb agreement and logical inference tasks to compare recurrent models with transformers. On the other hand, \citet{D18-1458_Self-Attention} use subject-verb agreement and word sense disambiguation for comparing those architectures in terms of capturing syntax and lexical semantics, respectively. In addition to these tasks, \citet{C18-1054} compare recurrent models with transformers in a multilingual machine translation task.

Having defined human interpretable intrinsic measures to capture lexical semantics and syntactic information in the hidden states, we can compare different neural architectures intrinsically in the complex task of machine translation. This helps to understand how the information that is captured by different architectures differs. Here, we compare recurrent and self-attention architectures, which use entirely different approaches to capture context.

In this chapter, we also try to shed light on the information encoded in the hidden states that goes beyond what is transferred from the word embeddings. To this end, we analyze to what extent the nearest neighbors of words based on their hidden state representations are covered by direct relations in WordNet \citep{wordnet_book,Miller:1995:WLD:219717.219748}. For our German experiments, we use GermaNet~\citep{W97-0802,HENRICH10.264}. From now on, we use \textit{WordNet} to refer to either WordNet or GermaNet.

This chapter does not directly seek improvements to neural translation models, but to further our understanding of the behavior of these models. It explains what information is learned in addition to what is already captured by embeddings. This chapter makes the following contributions:

\begin{enumerate}
\item We provide an analysis of the representations of hidden states in NMT systems highlighting the differences between hidden state representations and word embeddings.

\item We propose an intrinsic approach to study the syntactic and lexical semantic information captured in the hidden state representations based on the nearest neighbors. 

\item We show that the hidden states also capture a positional bias in addition to syntactic and lexical semantic information. The captured positional bias causes some words that do not have any syntactic or lexical semantic similarity to a word to end up in the list of the nearest neighbors of the word.

\item We compare transformer and recurrent models in a more intrinsic way in terms of capturing lexical semantics and syntactic structures. This is in contrast to previous work which focuses on extrinsic performance.

\item We provide analyses of the behavior of the hidden states for each direction layer and the concatenation of the states from the direction layers.
\end{enumerate}

\section{Related Work}

\citet{shi-padhi-knight:2016:EMNLP2016} were the first to use diagnostic classifiers to investigate the intermediate hidden states in sequence-to-sequence recurrent neural models. Using diagnostic classifiers based on the encoder hidden states, they verify that syntactic information is captured by the hidden states to some extent. They use two groups of syntactic features that cover word and sentence level syntactic labels. They use voice and tense of a sentence together with the top level syntactic sequence of the constituent tree as the sentence-level labels. As word-level labels, they use POS tags and the smallest phrase constituent above each word. They show that the hidden states of a sequence-to-sequence model trained for machine translation capture the syntactic labels accurately without having to be retrained for the classification task. They also investigate whether the hidden states capture deeper syntactic information when being trained on a translation task. Interestingly, their reported results show that the hidden states can even capture the constituent structure of the source sentence to a reasonable degree missing only some subtle details.

\citet{belinkov2017neural} build upon the work by \citet{shi-padhi-knight:2016:EMNLP2016} and expand the investigation to different variables, ranging from language specific features to architectural features. The authors use the extrinsic tasks of POS and morphological tagging as classification tasks into which they feed the intermediate hidden states to investigate how good a feature representation they are. 
They investigate four variables that possibly influence the information captured by the hidden state representations: (i) the richness of the morphology of the source and the target languages, (ii) whether a word-based or character-based architecture is used for the sequence-to-sequence model, (iii) the layer within the model from which the hidden states representations are captured, and (iv) the effect of attention on the hidden state representations from the decoder side of the model. 

Using diagnostic classifiers to investigate hidden states has become a popular method. \citet{giulianelli-etal-2018-hood} and \citet{Hupkes:2018:VCR:3241691.3241713} also use this method to inspect subject and verb agreement in the hidden states in language models and hierarchical structures learned by the hidden states in recurrent and recursive neural models. \citet{giulianelli-etal-2018-hood} show that letting diagnostic classifiers fix the intermediate representations can even lead to a boost in language modeling performance.

\citet{wallace-etal-2018-interpreting} take a nearest neighbor approach to investigating feature importance in deep neural networks. They adopt a nearest neighbors based conformity score introduced by \citet{DBLP:journals_corr_abs-1803-04765} to measure the impact of leaving features out on the classification performance. The nearest neighbor based conformity score is shown to be a robust decision measure which does not affect classification performance and also results in robustness and interpretability. This method uses the agreement of the nearest neighbors of the intermediate hidden states, saved during training with their corresponding labels, to make classification decisions during prediction. While this method is difficult to apply to sequence-to-sequence models, it shows the importance of the nearest neighbors of the intermediate hidden states for making robust predictions with neural networks.

\citet{ding-etal-2017-visualizing} use layer-wise relevance propagation (LRP) \citep{bach2015pixel} to study the contributions of words to the intermediate hidden states in a neural machine translation model. The LRP method was originally developed to compute the relevance of a single pixel for the predictions of an image classifier. They modify the method to be able to compute the relevance between two arbitrary hidden states in a neural machine translation model. This way, they can compute the contribution of the source and target word embeddings to the internal hidden states and to the next word being generated on the target side. The authors use the relevance computation to study the commonly observed errors made by neural machine translation systems including word omission, word repetition, unrelated word generation, negation reversion, and spurious word generation. 

\citet{kadar-etal-2017-representation} introduce an omission score to measure the contribution of an input token to the predictions of a recurrent neural network for the two tasks of language modeling and multi-modal meaning representations, where they use textual and visual input. They show that the network in language modeling task learns to pay more attention to tokens with a syntactic function.

In addition to work that pursues interpretability of the hidden states in neural sequence-to-sequence models by either diagnostic classifiers or relevance computation, there is work that seeks interpretability of complex neural models by comparing different neural architectures on some extrinsic tasks \citep{D18_Recurrent,D18-1458_Self-Attention}. 

In this chapter, we use an intrinsic approach to both investigate the information captured by the hidden states and to compare two commonly used neural machine translation architectures. This contributes to the interpretability of hidden states in both architectures.

\section{Datasets and Models}
\label{section:dataset_models}

We conduct our analyses using recurrent~\citep{bahdanau-EtAl:2015:ICLR} and transformer~\citep{NIPS2017_7181} machine translation models. Our recurrent model is a two-layer bidirectional recurrent model with Long Short-Term Memory (LSTM) units \citep{Hochreiter:1997} and global attention \citep{DBLP_journals_corr_LuongPM15}. The encoder consists of a two-layer unidirectional forward and a two-layer unidirectional backward pass. The corresponding output representations from each direction are concatenated to form the encoder hidden state representation for each token. A concatenation and down-projection of the last states of the encoder is used to initialize the first hidden state of the decoder. The decoder uses a two-layer unidirectional (forward) LSTM. We use no residual connections in our recurrent model as they have been shown to result in a performance drop if used on the encoder side of a recurrent model \citep{D17-1151}. Our transformer model is a 6-layer transformer with multi-head attention using 8 heads \citep{NIPS2017_7181}. We choose these settings to obtain competitive models with the relevant core components from each architecture.

\begin{figure}[htb!]
\centering
\includegraphics[scale=2.5]{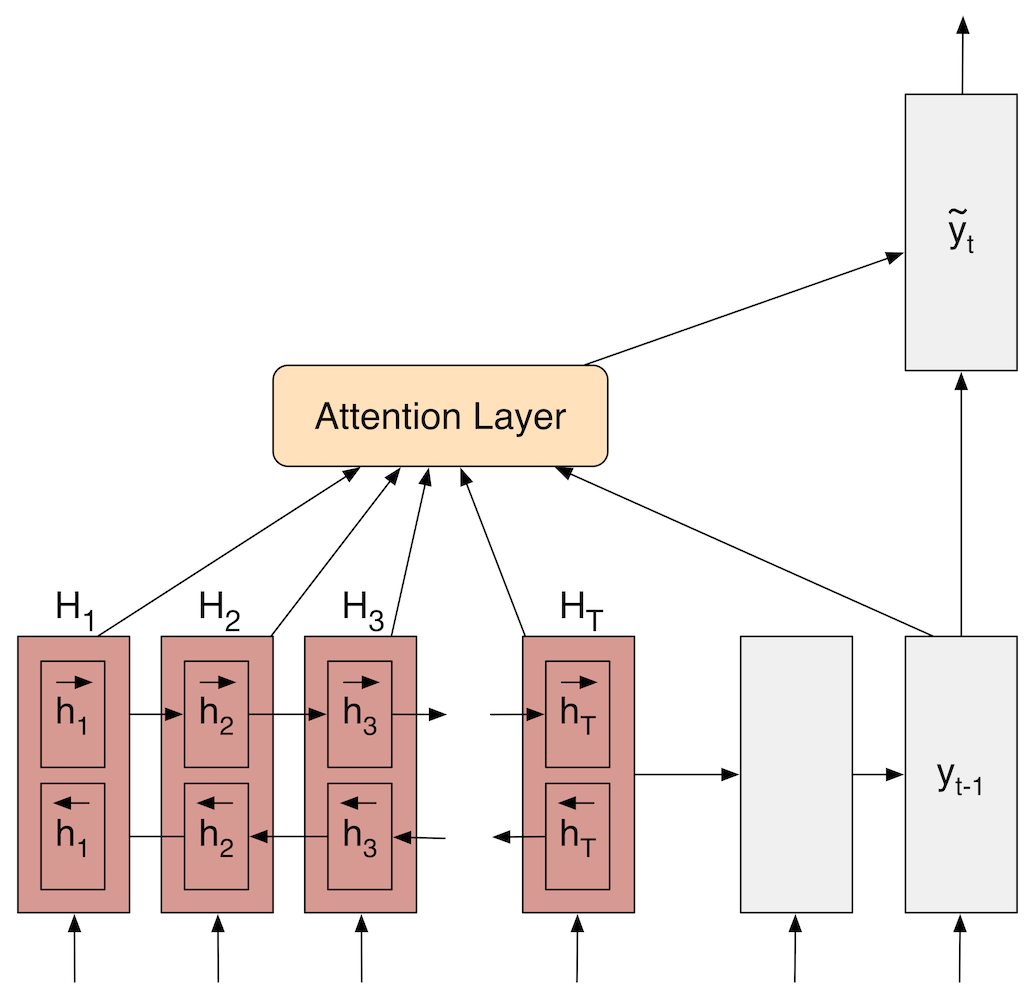}
\caption{Encoder hidden states, $(H_1, H_2, H_3, ..., H_T)$, of a recurrent model that are studied in this chapter.}
\label{fig:rec_hidden_states}
\end{figure}

\begin{figure}[htb!]
\centering
\includegraphics[scale=2.5]{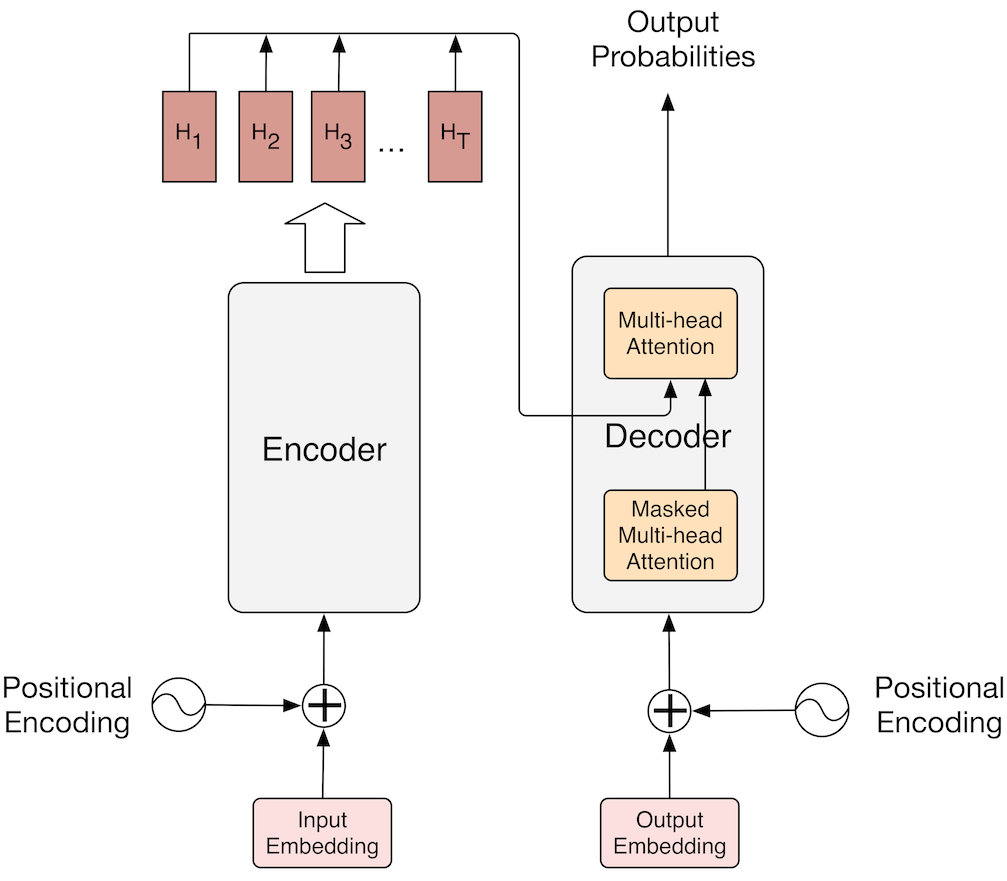}
\caption{Encoder hidden states, $(H_1, H_2, H_3, ..., H_T)$, of a transformer model that are studied in this chapter.}
\label{fig:trans_hidden_states}
\end{figure}

Figure~\ref{fig:rec_hidden_states} and Figure~\ref{fig:trans_hidden_states} show the hidden states ($H_1, H_2, H_3, ..., H_T$) that are of interest for our study in this chapter. These are the encoder hidden states that are fed to the cross attention mechanism in each architecture. Each of the figures shows the encoder hidden states in their respective neural machine translation architecture. Figure~\ref{fig:rec_hidden_states} and Figure \ref{fig:trans_hidden_states} show a recurrent neural architecture and a transformer architecture, respectively. Figure~\ref{fig:rec_hidden_states} also shows the direction-wise hidden states ($\overrightarrow{h}_1, \overleftarrow{h}_1, \overrightarrow{h}_2, \overleftarrow{h}_2, \overrightarrow{h}_3, \overleftarrow{h}_3, ..., \overrightarrow{h}_T, \overleftarrow{h}_T$) that are studied in Section~\ref{subsec:directio_wise}.

We train our models for two translation directions, namely English-German and German-English, both of which use the WMT15 parallel training data~\citep{bojar-etal-2015-findings}. We exclude 100k randomly chosen sentence pairs which are used as our held-out data. Our recurrent system has hidden state dimensions of size 1,024 (512 for each direction) and is trained using a batch size of 64 sentences. The learning rate is set to 0.001 for the Adam optimizer \citep{kingma:adam} with a maximum gradient norm of 5. A dropout rate of 0.3 is used to avoid overfitting.
Our transformer model has hidden state dimensions of 512 and a batch size of 4096 tokens and uses layer normalization~\citep{DBLP:journals/corr/BaKH16}. A learning rate of 2, changed under a warm-up strategy with 8000 warm-up steps, is used for the Adam optimizer with $\beta_1=0.9$, $\beta_2=0.998$ and $\epsilon=10^{-9}$ \citep{NIPS2017_7181}. The dropout rate is set to 0.1, and no gradient clipping is used. The word embedding size of both models is 512. We apply Byte-Pair Encoding (BPE) \citep{sennrichP16-1162} with 32K merge operations.

\begin{table*}[thb]
\centering
\small
\caption{ Performance of our experimental systems in BLEU on WMT \citep{bojar2017findings} German-English and English-German standard test sets. }
\label{table:Bleu}
\begin{tabular}{|l|c|c|c|c|}
 \multicolumn{5}{c}{English-German} \\
\hline
 Model&  test2014 & test2015 & test2016 & test2017 \\
\hline
Recurrent & 24.65 & 26.75 & 30.53 & 25.51\\
Transformer & 26.93 & 29.01& 32.44 & 27.36 \\
\hline
 \multicolumn{5}{c}{}\\
 \multicolumn{5}{c}{German-English} \\
\hline
 Model&  test2014 & test2015 & test2016 & test2017 \\
 \hline
 Recurrent & 28.40 & 29.61 & 34.28 & 29.64 \\
Transformer & 30.15 & 30.92 & 35.99 & 31.80 \\
\hline
\end{tabular}
\end{table*}

We train our models until convergence and then use the trained models to compute and log the hidden states for 100K sentences from a held-out dataset to use in our analyses. 
The remaining sentence pairs of the WMT15 parallel training data are used as our training data.

Table~\ref{table:Bleu} summarizes the performance of our experimental models in terms of BLEU \citep{papineni2002bleu} on different standard test sets. 

\begin{figure}[thb]
\centering
\includegraphics[scale=0.73]{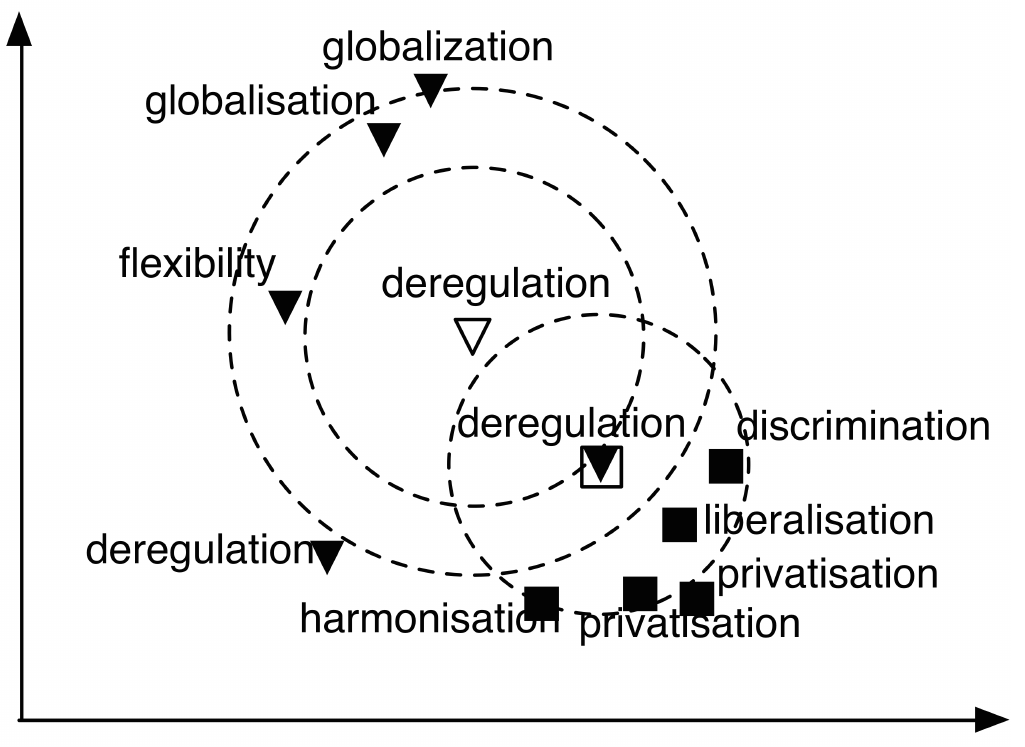}
\caption{An example of 5 nearest neighbors of two different occurrences of the word ``deregulation". Triangles are the nearest neighbors of one occurrence of ``deregulation" shown with the empty triangle. Squares are the nearest neighbors of another occurrence of ``deregulation" shown with the empty square. }
\label{fig:Nearest_Neighbors}
\end{figure}

\section{Nearest Neighbors Analysis}

Following earlier work on word embeddings \citep{mikolov2013distributed,W16-1620}, we choose to look into the nearest neighbors of the hidden state representations to learn more about the information encoded in them.
We treat each hidden state as the representation of the corresponding input token. This way, each \emph{occurrence} of a word has its own representation. Based on this representation, we compute the list of $n$ nearest neighbors of each word occurrence, using cosine similarity as the distance measure. We set $n$ equal to 10 in our experiments. 

In the case of our recurrent neural model, we use the concatenation of the corresponding output representations of our two-layer forward and two-layer backward passes as the hidden states of interest for our main experiments. We also use the output representations of the forward and the backward passes for our direction-wise experiments. In the case of our transformer model, we use the corresponding output of the top layer of the encoder for each word occurrence as the hidden state representation of the word.

Figure~\ref{fig:Nearest_Neighbors} shows an example of 5 nearest neighbors for two different occurrences of the word ``deregulation". Each item in this figure is a specific word occurrence, but we have removed their context information for the sake of simplicity. This figure shows how different occurrences of the same word can share no nearest neighbor in the hidden states space. It is interesting that one of the occurrences of the word ``deregulation" is the nearest neighbor of the other, but not vice versa.

\subsection{Hidden States vs. Embeddings}
\label{sec:hiddenStatesVsEmbeddings}

Here, we count how many of the words in the nearest neighbors lists of hidden states are covered by the nearest neighbors list based on the corresponding word embeddings. Just like the hidden states, the word embeddings used for computing the nearest neighbors are also from the same system and the same trained model for each experiment. The nearest neighbors of the word embeddings are also computed using cosine similarity. It should be noted that we generate the nearest neighbors lists for the embeddings and the hidden states separately and never compute cosine similarity  between word embeddings and the hidden state representations.

Coverage is formally computed as follows:
\begin{equation}
\label{eq:embeddingCov}
cp^{H,E}_{w_{i,j}} = \frac{\left\vert{C^{H,E}_{w_{i,j}}} \right\vert}{\left\vert{N^{H}_{w_{i,j}}}\right\vert},
\end{equation}
where 
\begin{equation}
\label{eq:hiddenEmbeddingIntersection}
C^{H,E}_{w_{i,j}} = N^{H}_{w_{i,j}} \cap N^{E}_{w}
\end{equation}
and $N^{H}_{w_{i,j}}$ is the set of the $n$ nearest neighbors of word $w$ based on hidden state representations. Since there is a different hidden state for each occurrence of a word, we use $i$ as the index of the sentence of occurrence and $j$ as the index of the word in the sentence. Similarly, $N^{E}_{w}$ is the set of the $n$ nearest neighbors of word $w$, but based on the embeddings. 

Word embeddings tend to capture the dominant sense of a word, even in the presence of significant support for other senses in the training corpus \citep{W16-1620}. Additionally, it is reasonable to assume that a hidden state corresponding to a word occurrence captures more of the context-specific sense of the word. Comparing the lists can provide useful insights as to which hidden state-based neighbors are not strongly related to the corresponding word embedding.
Furthermore, it shows in what cases the dominant information encoded in the hidden states comes from the corresponding word embedding and to what extent other information has been encoded in the hidden state.

\subsection{WordNet Coverage}
\label{subsec:WordNet_covrage}

In addition to comparisons with word embeddings, we also compute the coverage of the list of the nearest neighbors of hidden states with the directly related words from WordNet. This can shed further light on the capability of hidden states in terms of learning the sense of the word in the current context. Additionally, it constitutes an intrinsic measure to compare different architectures by their ability to learn lexical semantics. 
To this end, we check how many words from the nearest neighbors list of a word, based on hidden states, are in the list of related words of the word in WordNet.
More formally, we define $R_w$ to be the union of the sets of synonyms, antonyms, hyponyms and hypernyms of word $w$ in WordNet:
\begin{equation}
\label{eq:wordnetCov}
cp^{H,W}_{w_{i,j}} = \frac{\left\vert{C^{H,W}_{w_{i,j}}}\right\vert}{\left\vert{N^{H}_{w_{i,j}}}\right\vert},
\end{equation}
where
\begin{equation}
\label{eq:hiddenWordNetIntersection}
C^{H,W}_{w_{i,j}} = N^{H}_{w_{i,j}} \cap R_w
\end{equation}
and $N^{H}_{w_{i,j}}$ is the set of the $n$ nearest neighbors of word $w$ based on hidden state representations. 




\begin{figure}[thb]
\centering
\begin{subfigure}{.4\linewidth}
\centering
\includegraphics[scale=1]{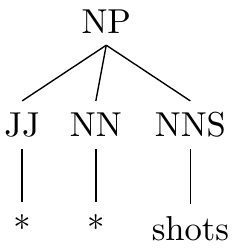}
\caption{$v_{0}$}
\label{fig:word_of_interest}
\end{subfigure} \hfill
\begin{subfigure}{.4\linewidth}
\centering
\includegraphics[scale=1]{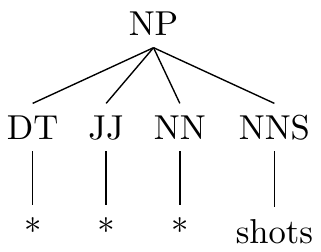}
\caption{$v_{1}$}
\label{fig:neighbor1}
\end{subfigure}

\begin{subfigure}{.4\linewidth}
\centering
\includegraphics[scale=1]{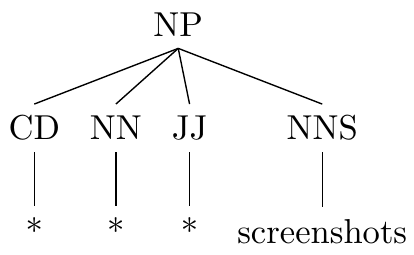}
\caption{$v_{2}$}
\label{fig:neighbor2}
\end{subfigure}\hfill
\begin{subfigure}{.4\linewidth}
\centering
\includegraphics[scale=1]{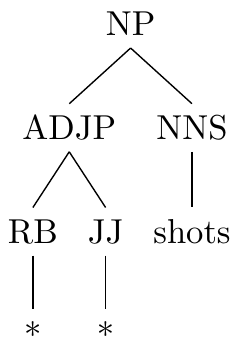}
\caption{$v_{10}$}
\label{fig:neighbor3}
\end{subfigure}
\caption{The figure shows the corresponding word and constituent subtree of query hidden state ($v_{0}$) and the corresponding word and subtree of the first ($v_{1}$) and the second ($v_{2}$) and the last ($v_{10}$) nearest neighbors of it.}
\label{fig:trees}
\end{figure}

\subsection{Syntactic Similarity}
\label{sec:syntax}

Recent  comparisons of recurrent and non-recurrent architectures for learning syntax \citep{D18_Recurrent,D18-1458_Self-Attention} also motivate a more intrinsic comparison of how much different syntactic and lexical semantic information they capture. To this end, we also study the nearest neighbors of hidden states in terms of syntactic similarities. For this purpose, we use the subtree rooting in the smallest phrase constituent above each word, following \citet{shi-padhi-knight:2016:EMNLP2016}. This way, we will have a corresponding parse tree for each word occurrence in our corpus. We parse our corpus using the Stanford constituent parser \citep{P13-1043}. We tag our corpus with POS tags and parse its sentences prior to applying BPE segmentation. Then, after applying BPE, we use the same POS tag and the same subtree of a word for its BPE segments, following \citet{W16-2209}. 

To measure the syntactic similarity between a hidden state and its nearest neighbors, we use the PARSEVAL standard metric \citep{evalb} as similarity metric between the corresponding trees. PARSEVAL computes precision and recall by counting the correct constituents in a parse tree with respect to a ground truth tree and divides the count by the number of constituents in the candidate parse tree and the ground truth tree, respectively. 

Figure~\ref{fig:word_of_interest} shows the corresponding word and subtree of a hidden state of interest. Figures~\ref{fig:trees}(b--d) show the corresponding words and subtrees of its three neighbors. The leaves are substituted with dummy ``*" labels to show that they do not influence the computed tree similarities. We compute the similarity score between the corresponding tree of each word and the corresponding trees of its nearest neighbors. For example, in Figure~\ref{fig:trees} we compute the similarity score between the tree in Figure~\ref{fig:word_of_interest} and each of the other trees.

\subsection{Concentration of Nearest Neighbors}
Each hidden state, together with its nearest neighbors, behaves like a cluster centered around the corresponding word occurrence of the hidden state, where the neighboring words give a clearer indication of the captured information in the hidden state. However, this evidence is more clearly observed in some cases rather than others, as in some cases the neighboring words appear unrelated. 

The stronger the similarities that bring the neighbors close to a hidden state, the more focused the neighbors around the hidden state are. Bearing this in mind, we choose to study the relation between the concentration of the neighbors and the information encoded in the hidden states.  

To make it simple but effective, we estimate the population variance of the neighbors' distances from a hidden state as the measure of the concentration of its neighbors. More formally, this is computed as follows:
\begin{equation}
\label{eq:variance}
v_{w_{i,j}} = \frac{1}{n}\sum_{k=1}^{n}(1-x_{k, w_{i,j}})^2.
\end{equation} 
\noindent Here, $n$ is the number of neighbors and $x_{k, w_{i,j}}$ is the cosine similarity score of the $k$-th neighbor of word $w$ occurring as the $j$-th token of the $i$-th sentence.







\subsection{Positional Distribution of the Nearest Neighbors}
\label{sec:positionalDist}

When inspecting many examples of the nearest neighbors of hidden states, we noticed that some positional information is being encoded in the hidden states. To test this hypothesis, we propose to estimate the distribution of the relative position of the nearest neighbors of the hidden states. 

We define the relative position as the position of a hidden state, or corresponding token, with respect to the length of the sentence that the token appears in. The relative position of a token is computed as follows:
\begin{equation}
rp_{w_{i,j}} = \floor*{\frac{j*10}{l(i)}}.
\end{equation} 

\noindent Here, $w_{i,j}$ is the $j$-th token of the $i$-th sentence and $l(i)$ is the length of the sentence.

We compute the positional distribution of the nearest neighbors for each relative position. Based on this definition, we have 10 bins of relative positions. For each relative position we count the number of neighbors in the same bin. Then we use these numbers to estimate a normal distribution for each relative position. 
%
%
%
%

\section{Empirical Analyses}
\label{sec:emp_analysis_ch3}

We train our systems for English-German and German-English and use our trained model to compute the hidden state representations on held-out data of 100K sentences. We log the hidden state representations together with their corresponding source tokens and their sentence and token indices.

We use the logged hidden states to compute the nearest neighbors of the tokens with frequencies between 10 and 2000 in our held-out data. We compute cosine similarity to find the nearest neighbors.

In addition to the hidden states, we also log the word embeddings from the same system and the same trained model. Similar to the hidden states, we also use embedding representations to compute the nearest neighbors of words. We have to note that in the case of embedding representations we have one nearest neighbor list for each word whereas for hidden states there is one list for each occurrence of a word.

\begin{table*}[thb]
\centering
\small
\caption{Percentage of the nearest neighbors of hidden states covered by the list of the nearest neighbors of embeddings. }
\label{table:EmbedCoverage}
\begin{tabular}{|l|l|c|c|c|c|}
\hline
Model &POS & English-German& $\sigma^2$ & German-English & $\sigma^2$\\
\hline
\multirow{5}{*}{Recurrent} & All POS & 18\%& 4 & 24\% & 7\\
& VERB & 29\% & 5 & 31\%& 5\\
& NOUN & 14\%& 3& 19\%& 8\\
& ADJ & 19\% & 3 & 31\%& 7\\
& ADV & 36\%& 5 &48\%& 2 \\
\hline
\multirow{5}{*}{Transformer} & All POS & 37\%& 14 & 33\%& 10\\
& VERB & 39\% & 8 &  36\%& 7 \\
& NOUN & 38\% & 16 & 31\%&14\\
& ADJ & 32\%& 11 &36\%& 9\\
& ADV & 33\% & 12 & 38\%& 3\\
\hline
\end{tabular}
\end{table*}

\setcounter{footnote}{0}

\subsection{Nearest Neighbors Coverage of Embedding}

As a first experiment, we measure how many of the nearest neighbors based on the embedding representation are retained as nearest neighbors of the corresponding hidden state, as described in Section~\ref{sec:hiddenStatesVsEmbeddings}.

\noindent Table~\ref{table:EmbedCoverage} shows the statistics of the coverage by the nearest neighbors based on embeddings in general and based on selected source POS tags for each of our models. To carry out an analysis based on POS tags, we tagged our training data using the Stanford POS tagger \citep{toutanova2003feature}. We convert the POS tags to the universal POS tags and report only for POS tags available in WordNet. We use the same POS tag of a word for its BPE segments, as described in the Section~\ref{sec:syntax}.

The first row of Table~\ref{table:EmbedCoverage} shows that for our recurrent model only 18\% (for English) and 24\% (for German) of the information encoded in the hidden states is already captured by the word embeddings. Interestingly, in all cases except ADV, the similarity between the hidden states and the embeddings is much higher for the transformer model, and the increase for nouns is much higher than for the rest. This may be a product of not using recurrence in the transformer model which results in a simpler path from each embedding to the corresponding hidden state. We hypothesize that this means that the recurrent model uses the capacity of its hidden states to encode other information that is encoded to a lesser extent in the hidden states of the transformer model.  

\begin{table*}[thb]
\centering
\small
\caption{Percentage of the nearest neighbors of hidden states covered by the list of the directly related words to the corresponding word of the hidden states in WordNet. }
\label{table:wordnetCoverage}
\begin{tabular}{|l|l|c|c|c|c|}
\hline
Model & POS & English-German& $\sigma^2$ & German-English & $\sigma^2$\\
\hline
\multirow{5}{*}{Recurrent} & All POS & 24\%& 6 & 51\%& 12\\
&VERB & 49\% & 9 & 48\% & 10\\
&NOUN & 19\% & 3 & 28\% & 8\\
& ADJ & 15\% & 2 & 60\% & 12 \\
& ADV & 24\% & 4 & 23\% & 1 \\
\hline
\multirow{5}{*}{Transformer} & All POS & 67\%& 16 & 74\%& 10\\
& VERB & 77\% & 9 & 70\% & 9 \\
& NOUN & 65\% & 18 & 63\% & 13\\
& ADJ & 66\%& 14 & 81\% & 9 \\
& ADV & 74\% & 10 & 35\%& 5\\
\hline
\end{tabular}
\end{table*}

\subsection{WordNet Coverage}

Having observed that a large portion of nearest neighbors of the hidden states are still not covered by the nearest neighbors of the corresponding word embeddings, we look for other sources of similarity that cause the neighbors to appear in the list. As our next step, we check to see how many of the neighbors are covered by directly related words of the corresponding word in WordNet.

This does not yield subsets of the nearest neighbors that are fully disjoint with the subset covered by the nearest neighbors from the embedding list. However, it still shows whether this source of similarity is fully covered by the embeddings or whether the hidden states capture information from this source that the embeddings miss.

%
\begin{figure}[thb]
\centering
\includegraphics[scale=0.3]{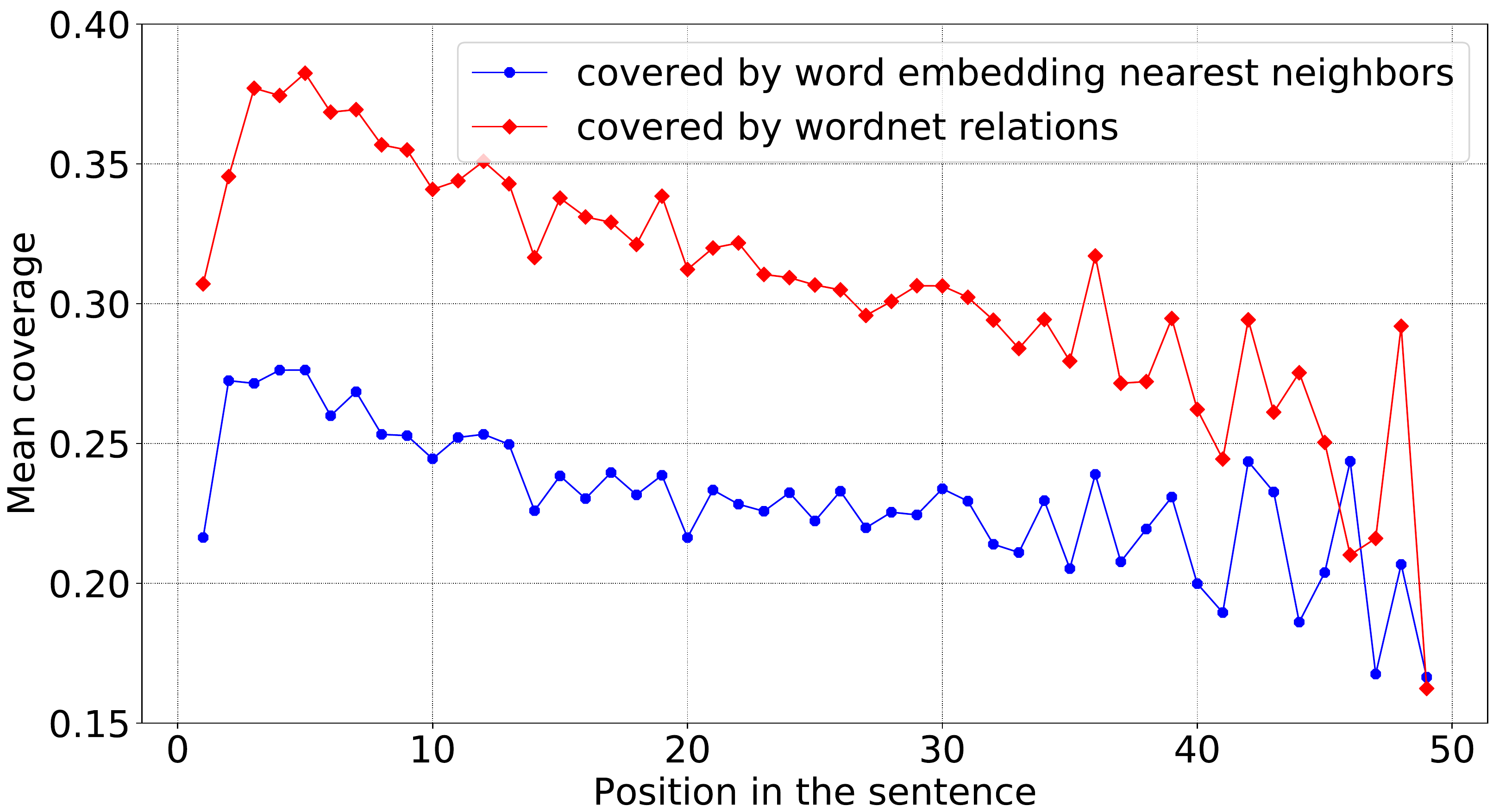}
\caption{The mean coverage per position of the nearest neighbors of hidden states in the recurrent model; (i) by the nearest neighbors of the embedding of the corresponding word, and (ii) by WordNet related words of the corresponding word of the hidden state.}
\label{fig:wordnetAndEmbeddingCoverage}
\end{figure}

\begin{figure}[thb]
\centering
\includegraphics[scale=0.3]{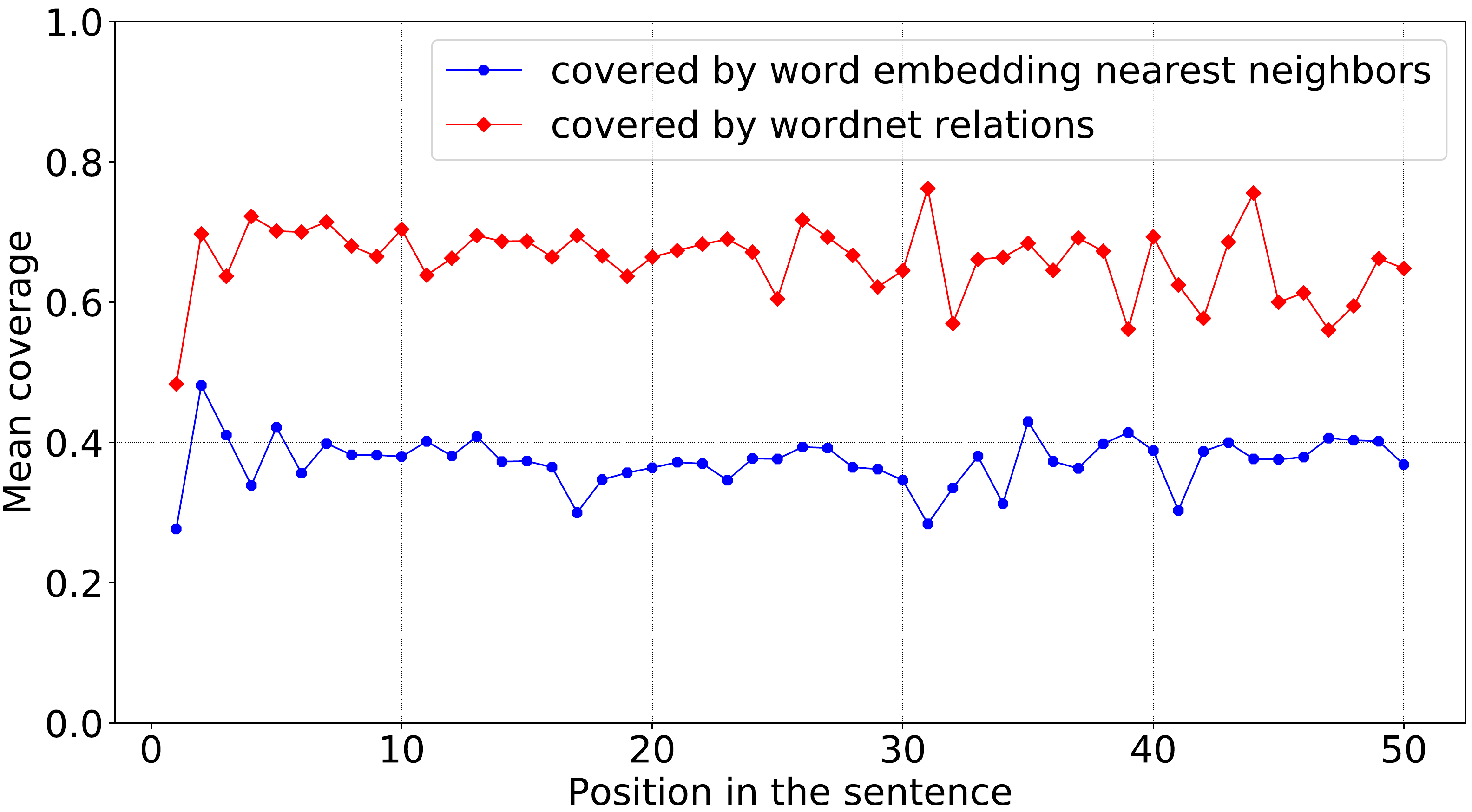}
\caption{The mean coverage per position of the nearest neighbors of hidden states in the transformer model; (i) by the nearest neighbors of the embedding of the corresponding word, and (ii) by WordNet related words of the corresponding word of the hidden state.}
\label{fig:wordnetAndEmbeddingCoverage_transformer}
\end{figure}

\begin{sidewaysfigure}
\centering
\begin{subfigure}{.4\linewidth}
\centering
\includegraphics[scale=0.235]{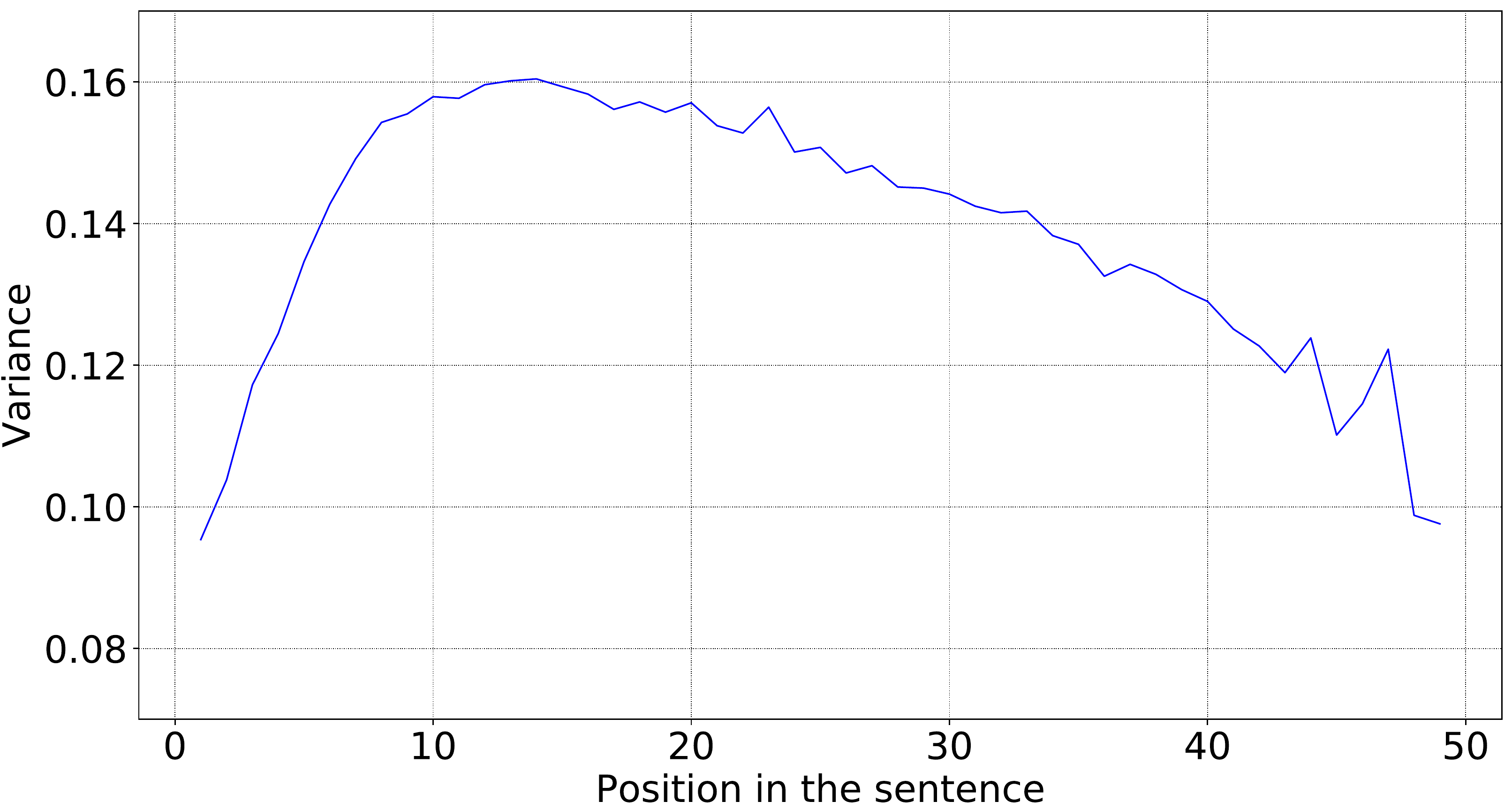}
\caption{English-German, recurrent model}
\label{fig:variance_recur_1}
\end{subfigure}
\begin{subfigure}{.4\linewidth}
\centering
\includegraphics[scale=0.235]{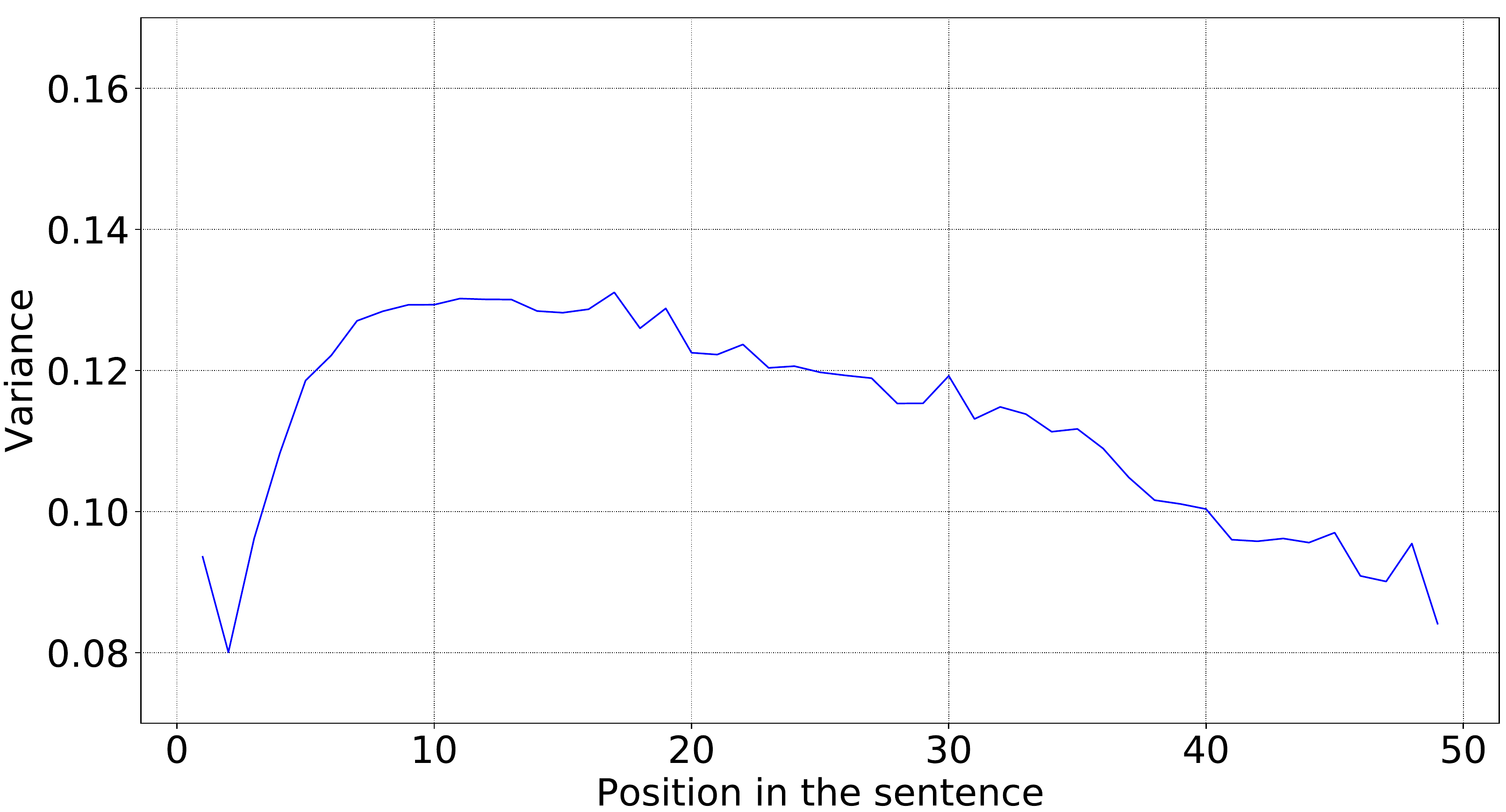}
\caption{German-English, recurrent model }
\label{fig:variance_recur_2}
\end{subfigure}

\begin{subfigure}{.4\linewidth}
\centering
\includegraphics[scale=0.235]{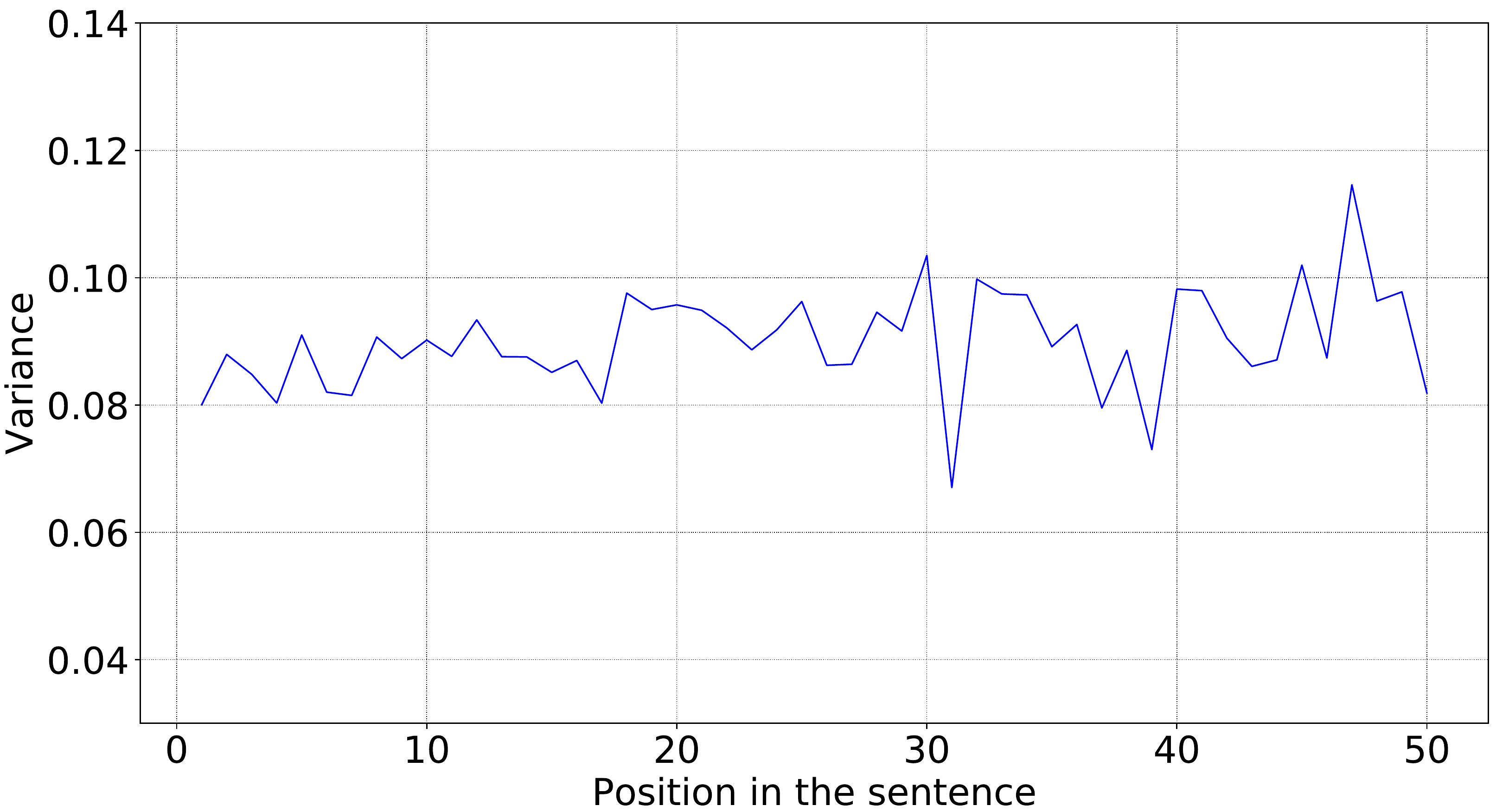}
\caption{English-German, transformer model}
\label{fig:variance_transfor}
\end{subfigure}
\begin{subfigure}{.4\linewidth}
\centering
\includegraphics[scale=0.235]{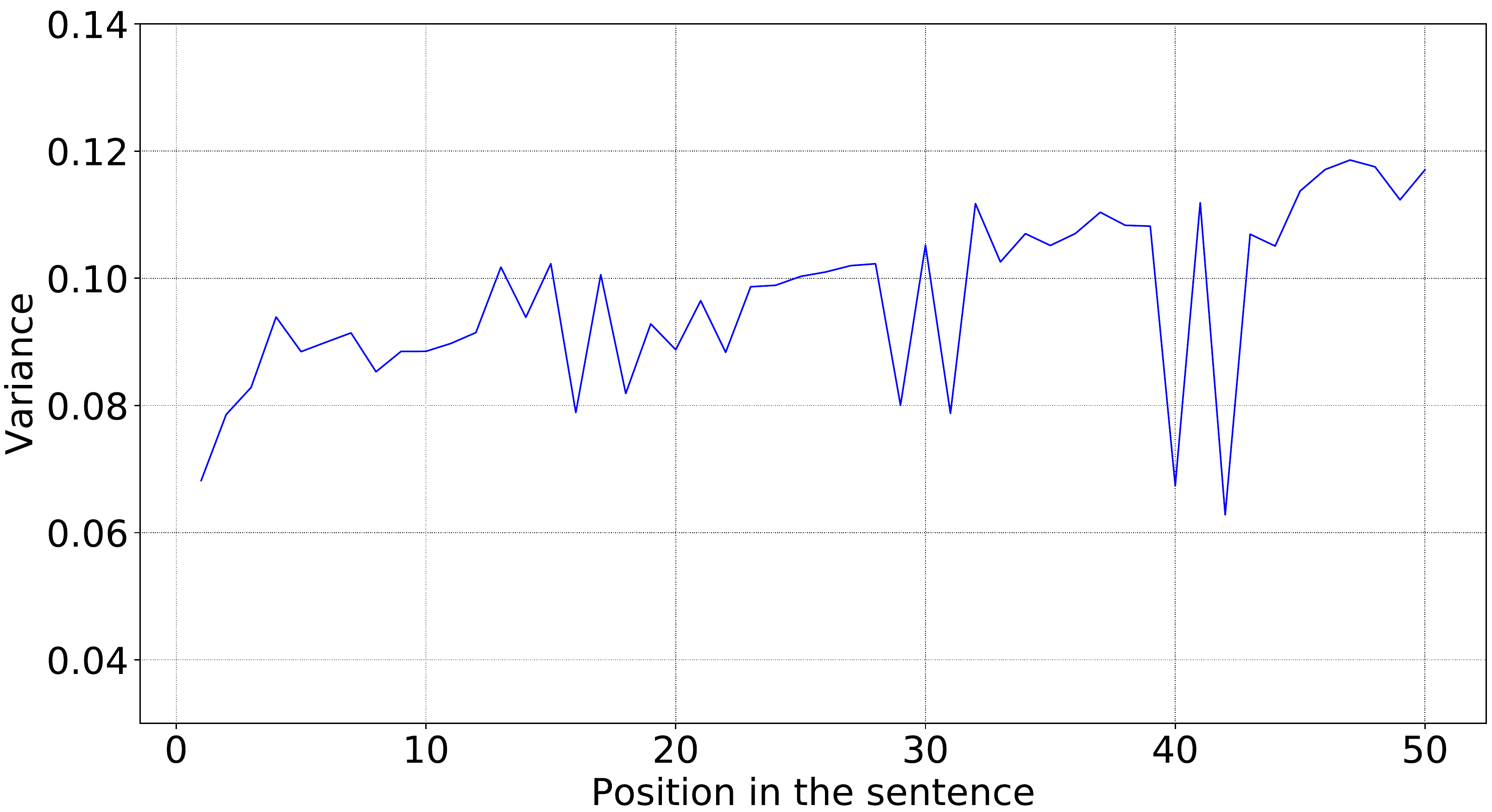}
\caption{German-English, transformer model}
\label{fig:variance_transfor_de_en}
\end{subfigure}
\caption{The average variance of cosine distance scores of the nearest neighbors of words per positions.}
\label{fig:varianceVsPosition}
\end{sidewaysfigure}

Table~\ref{table:wordnetCoverage} shows the general and the POS-based coverage for our English-German and German-English systems. 
The transformer model again has the lead by a large margin. The jump for nouns is again the highest, similar to what we have already seen in Table~\ref{table:EmbedCoverage}. This basically means that more words from the WordNet relations of the word of interest are present in the hidden state nearest neighbors of the word. A simple explanation for this is that the hidden states in the transformer model capture more word semantics than the hidden states in the recurrent model. Or in other words, the hidden states in the recurrent model capture some additional information that brings words beyond the WordNet relations to its neighborhood.  

To investigate whether recurrency has any direct effect on the observed difference between the transformer and the recurrent model in Table~\ref{table:wordnetCoverage}, we compute the mean coverage by direct relations in WordNet per position. Similarly, we also compute the mean coverage by embedding neighbors per position. More formally, we write:
\begin{equation}
acp_{j}^{H,W} = \frac{\sum_{i=1}^{m}(cp^{H,W}_{w_{i,j}})}{\left\vert{S_{l(s) \geq j}}\right\vert}
\end{equation}
and 
\begin{equation}
acp_{j}^{H,E} = \frac{\sum_{i=1}^{m}(cp^{H,E}_{w_{i,j}})}{\left\vert{S_{l(s) \geq j}}\right\vert}
\end{equation}
for the mean coverage by WordNet direct relations per position and the mean coverage by embedding neighbors per position, respectively.

Here $cp^{H,W}_{w_{i,j}}$ and $cp^{H,E}_{w_{i,j}}$ are the values computed in Equation~\ref{eq:wordnetCov} and~\ref{eq:embeddingCov}, respectively. The function $l(s)$ returns the length of sentence $s$ and $S_{l(s) \geq j}$ is the set of sentences that are longer than or equal to $j$.

Figure~\ref{fig:wordnetAndEmbeddingCoverage} shows that for the recurrent model the mean coverage  for both embedding and WordNet is first increasing, but starts to decrease from position 5 onwards. This drop in coverage is surprising, considering that the model is a bidirectional recurrent model. However, this may be a reflection of the fact that the longer a sentence, the less the hidden states encode information about the corresponding word.

Figure~\ref{fig:wordnetAndEmbeddingCoverage_transformer} shows the same mean coverage for the hidden states in the transformer model. The absence of decreasing coverage with regard to position, confirms our hypothesis that the lower coverage for recurrent models indeed directly relates to the recurrency. 

In order to refine the analysis of the positional behavior of the hidden states, we compute the average variance per position of the cosine distance scores of the nearest neighbors based on the hidden states. To compute this value we use the following definition:
\begin{equation}
Av_{j} = \frac{\sum_{i=1}^{m}v_{w_{i,j}}}{\left\vert{S_{l(s) \geq j}}\right\vert}.
\end{equation}
\noindent Here, $v_{w_{i,j}}$ is the variance estimate as defined in Equation~\ref{eq:variance}, $l(s)$ is the function returning the length of sentence $s$ and $S_{l(s) \geq j}$ is the set of sentences that are longer than or equal to $j$ as mentioned earlier. 

\begin{sidewaysfigure}
\centering
\includegraphics[scale=0.40]{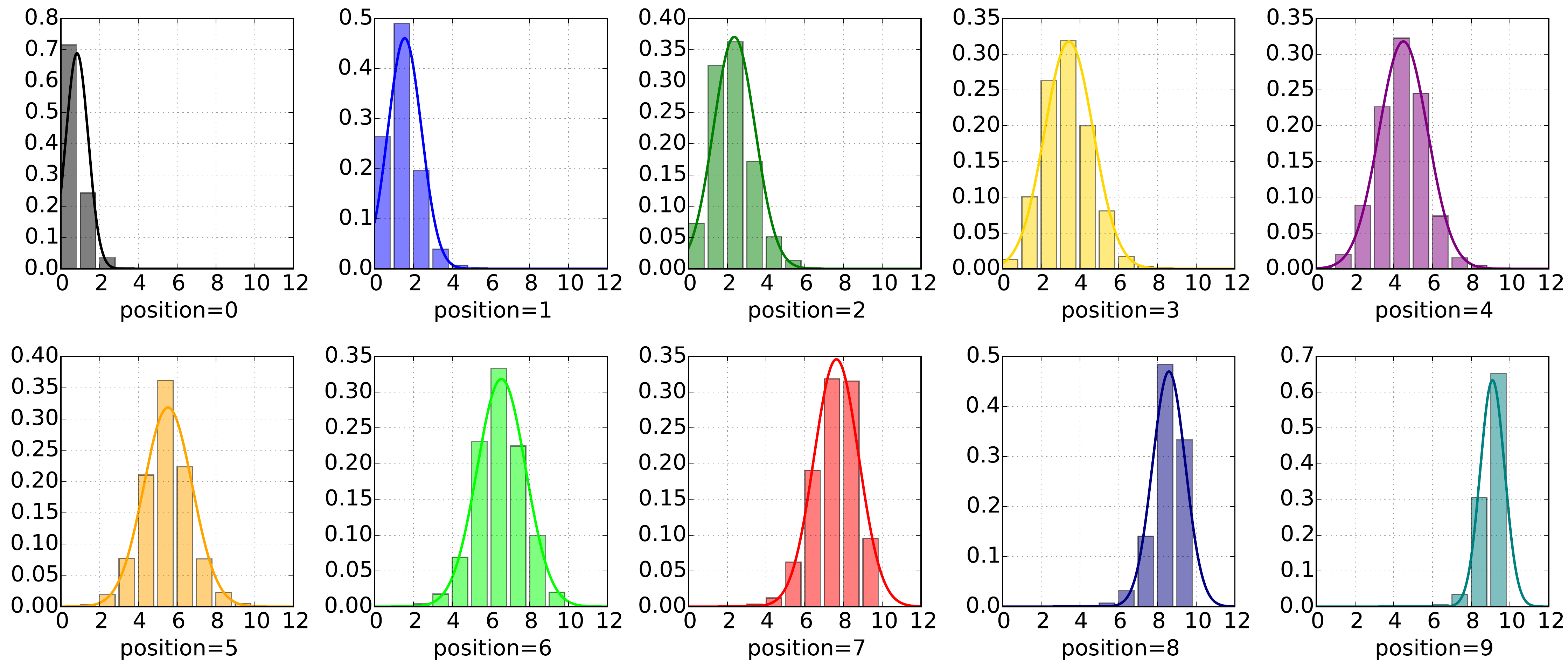}
\caption{Estimated distributions of the relative positions of the nearest neighbors of hidden states, per bin of relative position. This is for the English-German experiment.}
\label{fig:positionDistributions_separate}
\end{sidewaysfigure}

\begin{sidewaysfigure}
\centering
\includegraphics[scale=0.40]{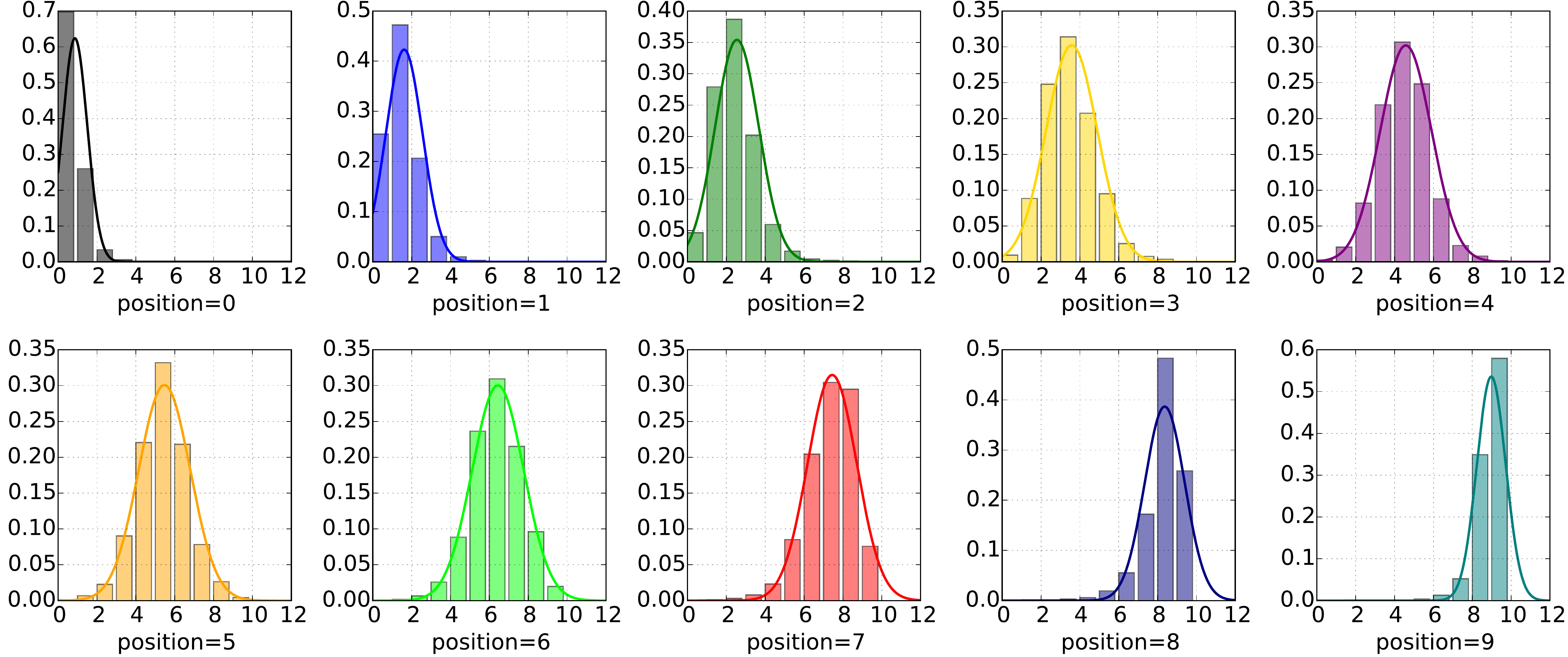}
\caption{Estimated distributions of the relative positions of the nearest neighbors of hidden states, per bin of relative position. This is for the German-English experiment.}
\label{fig:positionDistributions_separate_de}
\end{sidewaysfigure}

\begin{figure}[thb]
\centering
\begin{subfigure}{\linewidth}
\centering
\includegraphics[scale=0.30]{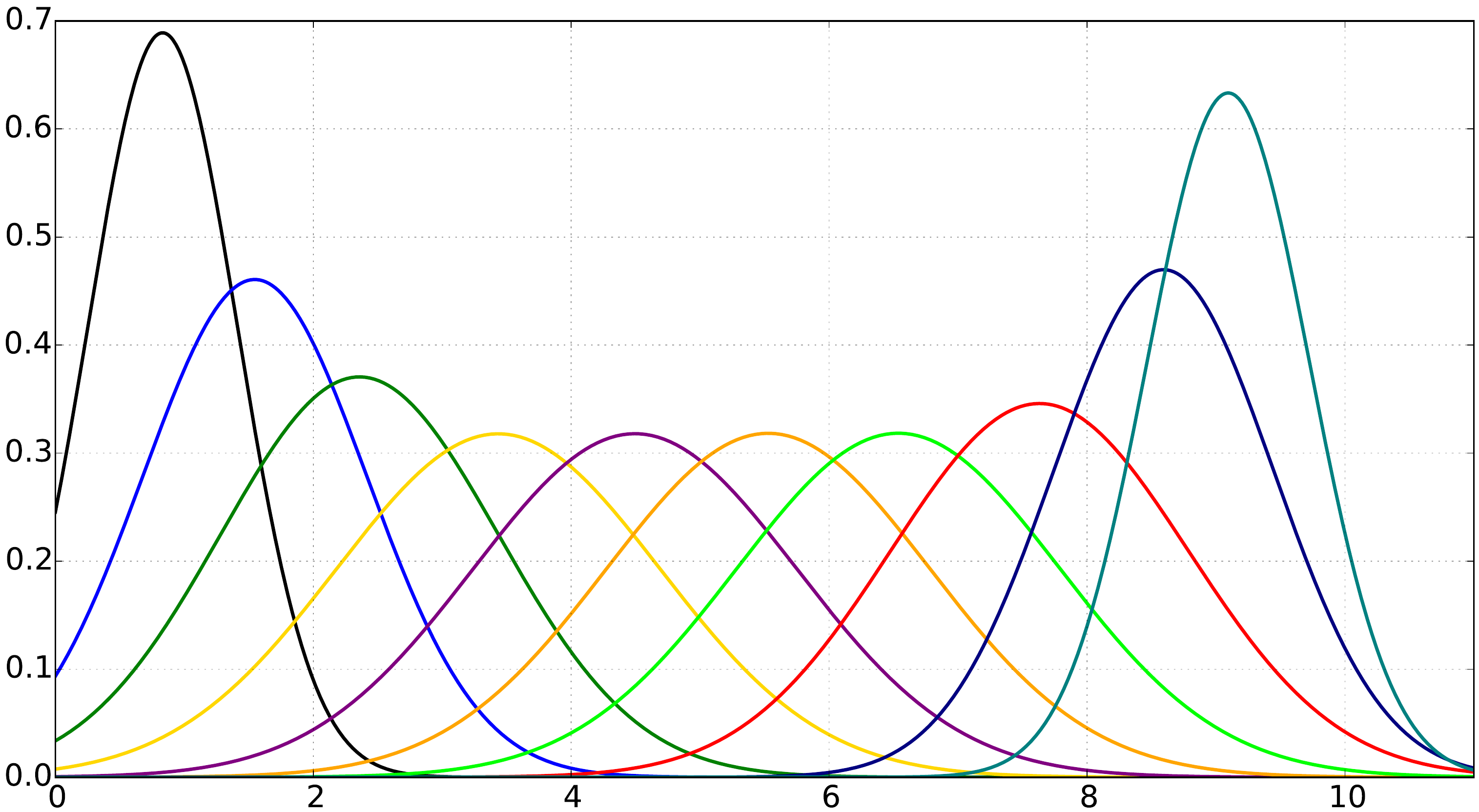}
\caption{English-German}
\end{subfigure}

\begin{subfigure}{\linewidth}
\centering
\includegraphics[scale=0.30]{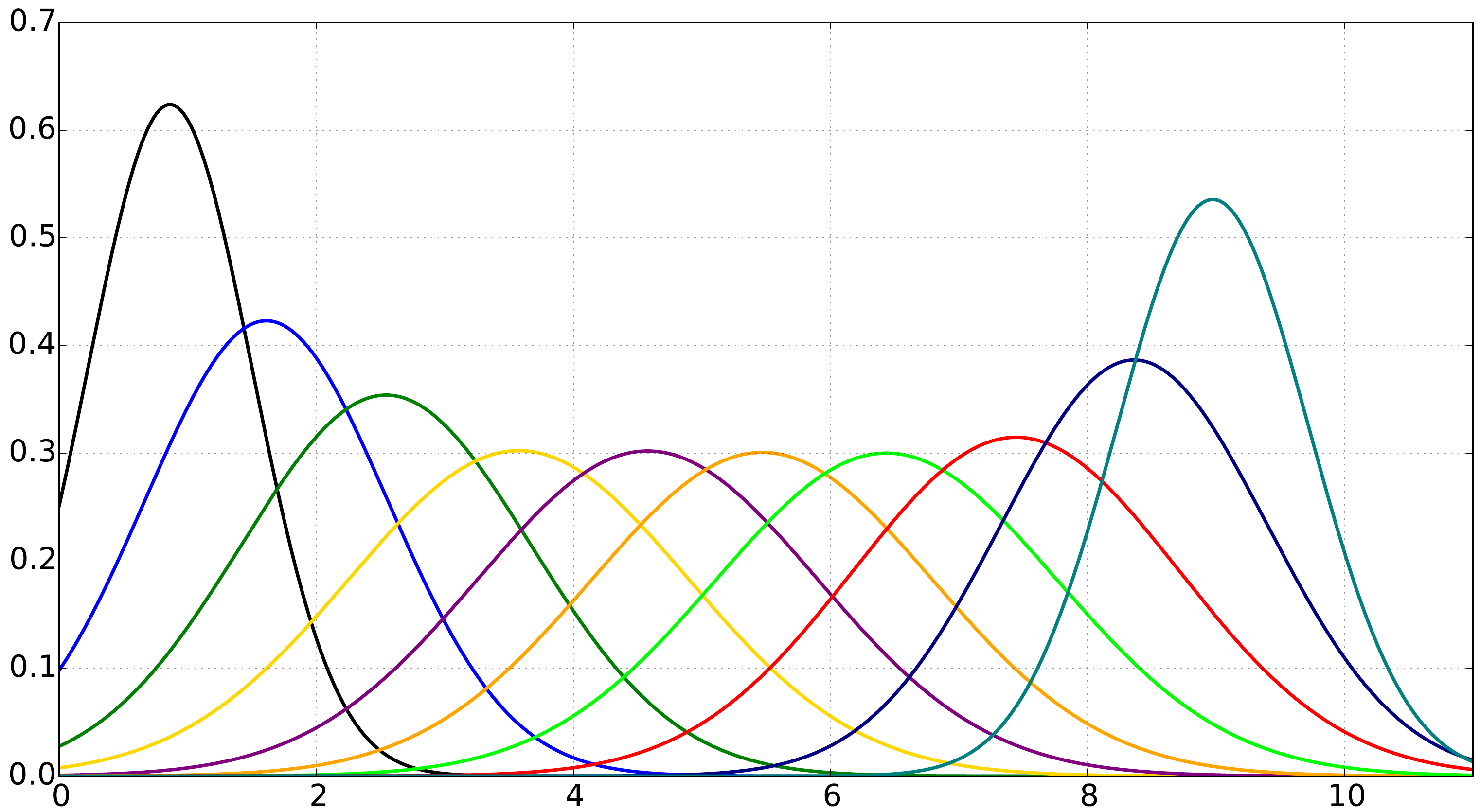}
\caption{German-English}
\end{subfigure}
\caption{Estimated distributions of the relative positions of the nearest neighbors of hidden states per bin.}
\label{fig:positionDistributions}
\end{figure}

Figures~\ref{fig:variance_recur_1} and \ref{fig:variance_recur_2} show the average variance per position for the recurrent model. One can see that the average variance close to the borders is lower than the variance in the middle. This means that the nearest neighbors of the words close to the borders of sentences are more concentrated in terms of similarity score in general. This could mean that the lexical meaning of those words plays less of a role compared to other information encoded in the corresponding hidden states, especially if we take the coverage per position into account.  Interestingly, this does not happen for the transformer model; see Figures~\ref{fig:variance_transfor} and \ref{fig:variance_transfor_de_en}.

\subsection{Positional Bias}

To test the hypothesis whether the nearest neighbors of the hidden states are sharing some positional bias, we estimate the position distribution of the nearest neighbors. This can also provide more evidence for the behavior of the hidden states close to the borders of sentences.

Figure~\ref{fig:positionDistributions_separate} and Figure~\ref{fig:positionDistributions_separate_de} show the distributions of relative position of nearest neighbors for the 10 bins of relative positions, as explained in Section~\ref{sec:positionalDist} for the English and German source sides, respectively. The closer it is to the borders, the stronger the positional bias becomes and the more nearest neighbors result from closer positions.

%
%
Figure~\ref{fig:positionDistributions} combines the distributions from Figure~\ref{fig:positionDistributions_separate} and Figure~\ref{fig:positionDistributions_separate_de} in two respective plots to make it easier to compare. Note the higher concentration (lower variance) of the distributions near the borders in both languages. For the sake of readability, we have removed the histograms in this figure. 

 %
This bias causes some words that do not share lexical or syntactic similarities with the words close to the borders to end up in their nearest neighbors list. This could be interpreted as an example of where corpus-level discourse dependency \citep{P16-1125} in recurrent neural networks may help improve language modelling performance.

\begin{table*}[thb]
\centering
\small
\caption{Average parse tree similarity (PARSEVAL scores) between word occurrences and their nearest neighbors. Note that the apparent identity of precision and recall values is due to rounding and the very close number of constituents in the candidate and gold parse trees.}
\label{tab:syntactic_similarity}
\begin{tabular}{|l|c|c|}
\multicolumn{3}{c}{English-German}\\
\hline
Model & Recurrent & Transformer \\  
\hline
Precision &  0.38   & 0.31\\
Recall & 0.38  & 0.31\\
Matched Brackets & 0.42  & 0.35\\
Cross Brackets & 0.31  & 0.28\\
Tag Accuracy & 0.46  & 0.40\\
\hline
\multicolumn{3}{c}{}\\
\multicolumn{3}{c}{German-English}\\
\hline
Model & Recurrent & Transformer\\
\hline
Precision & 0.12 & 0.11\\
Recall & 0.12 & 0.11 \\
Matched Brackets & 0.30 & 0.28\\
Cross Brackets & 0.80 & 0.77\\
Tag Accuracy & 0.32 & 0.31\\
\hline
\end{tabular}
\end{table*}

\subsection{Syntactic Similarity}
\label{subsec:syntactic_ch3}

The differences between recurrent and transformer models, as we have seen for the coverage of embeddings and WordNet relations, motivate a comparison of these models based on the syntactic similarities between the hidden states and their nearest neighbors. Especially because it has been shown on extrinsic tasks that recurrent models are superior in capturing syntactic structures~\citep{D18_Recurrent}.
To this end, we use the approach introduced in Section~\ref{sec:syntax}. 

\begin{figure}[tbh!]
\centering
\begin{subfigure}{\linewidth}
\centering
\includegraphics[scale=0.3]{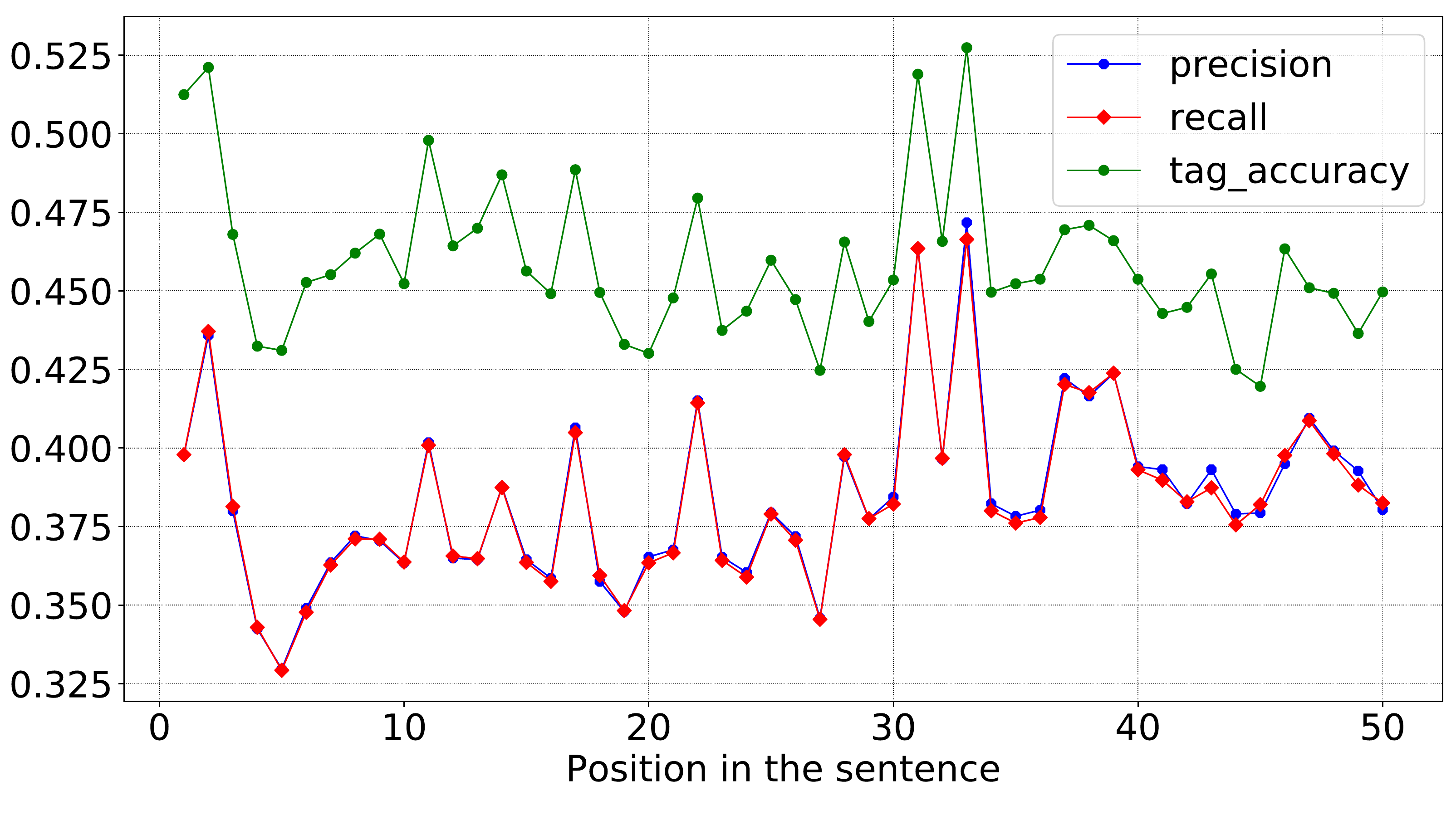}
\caption{Recurrent model.}
\label{fig:}
\end{subfigure} 

\begin{subfigure}{\linewidth}
\centering
\includegraphics[scale=0.3]{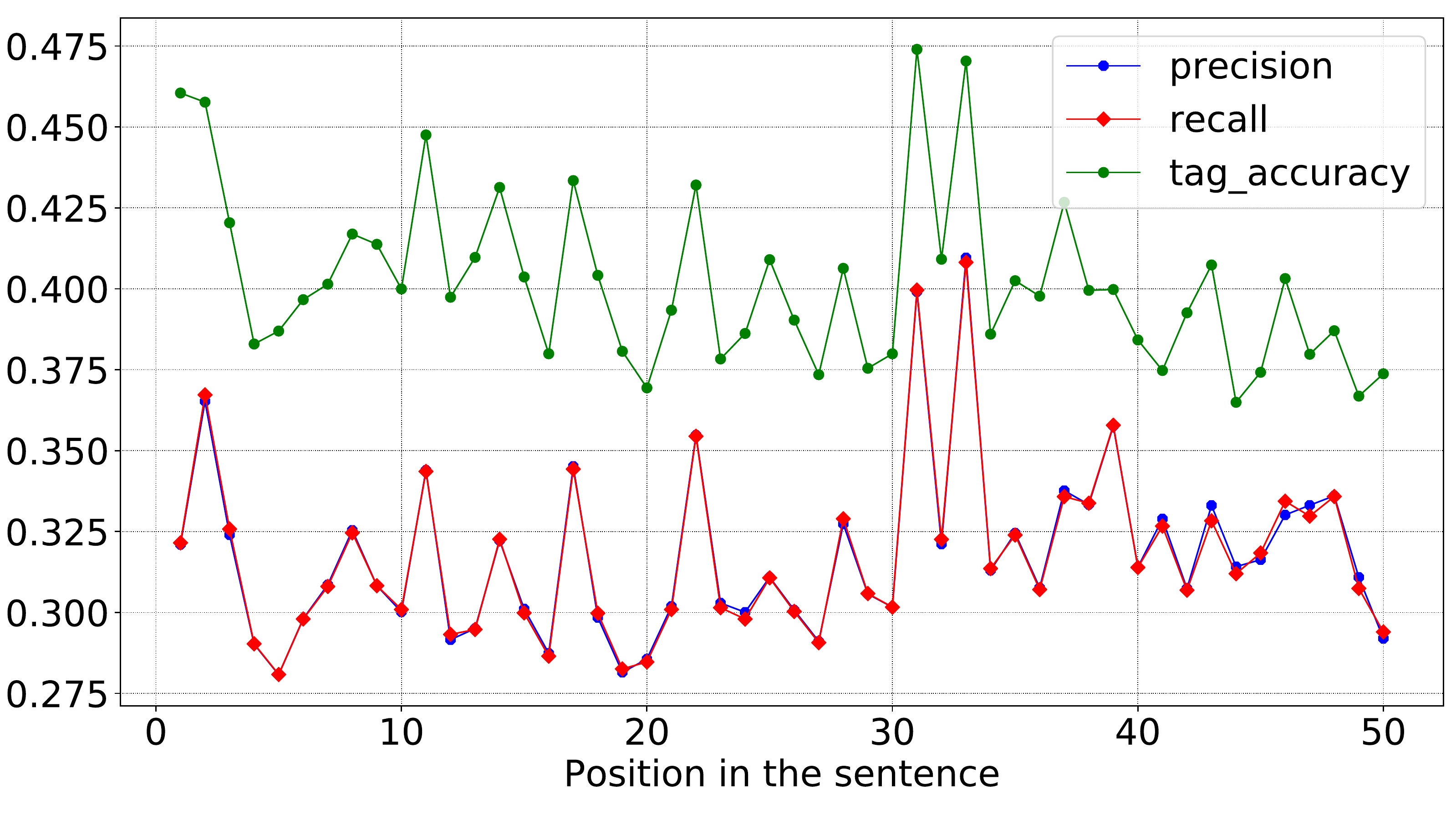}
\caption{Transformer.}
\label{fig:}
\end{subfigure}
\caption{Average syntactic similarity for each sentence position for sentences with the length up to 50 tokens for English-German.}
\label{fig:syntax_in_sentence_en_de}
\end{figure}

\begin{figure}[tbh!]
\centering
\begin{subfigure}{\linewidth}
\centering
\includegraphics[scale=0.3]{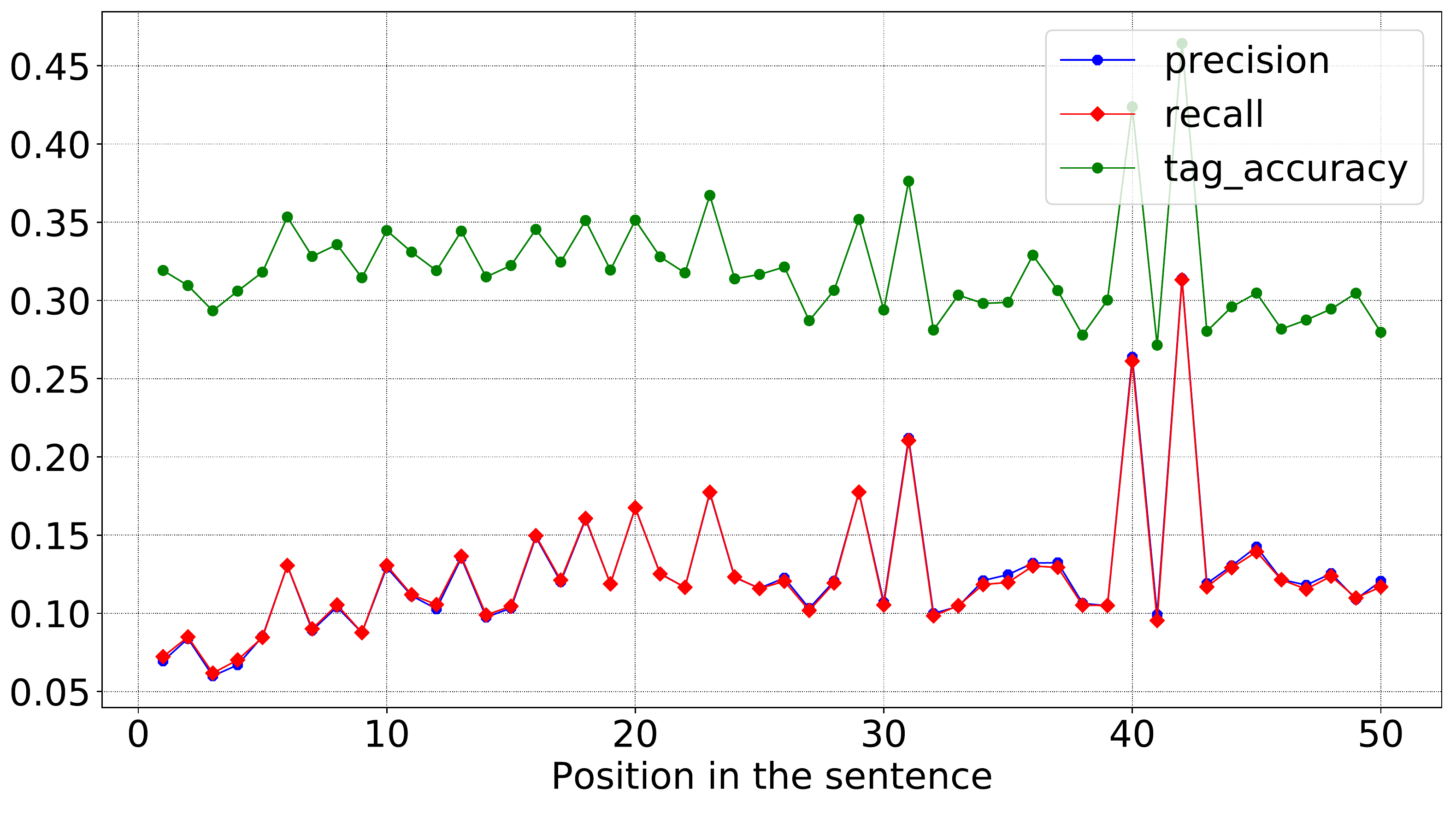}
\caption{Recurrent model.}
\label{fig:}
\end{subfigure}

\begin{subfigure}{\linewidth}
\centering
\includegraphics[scale=0.3]{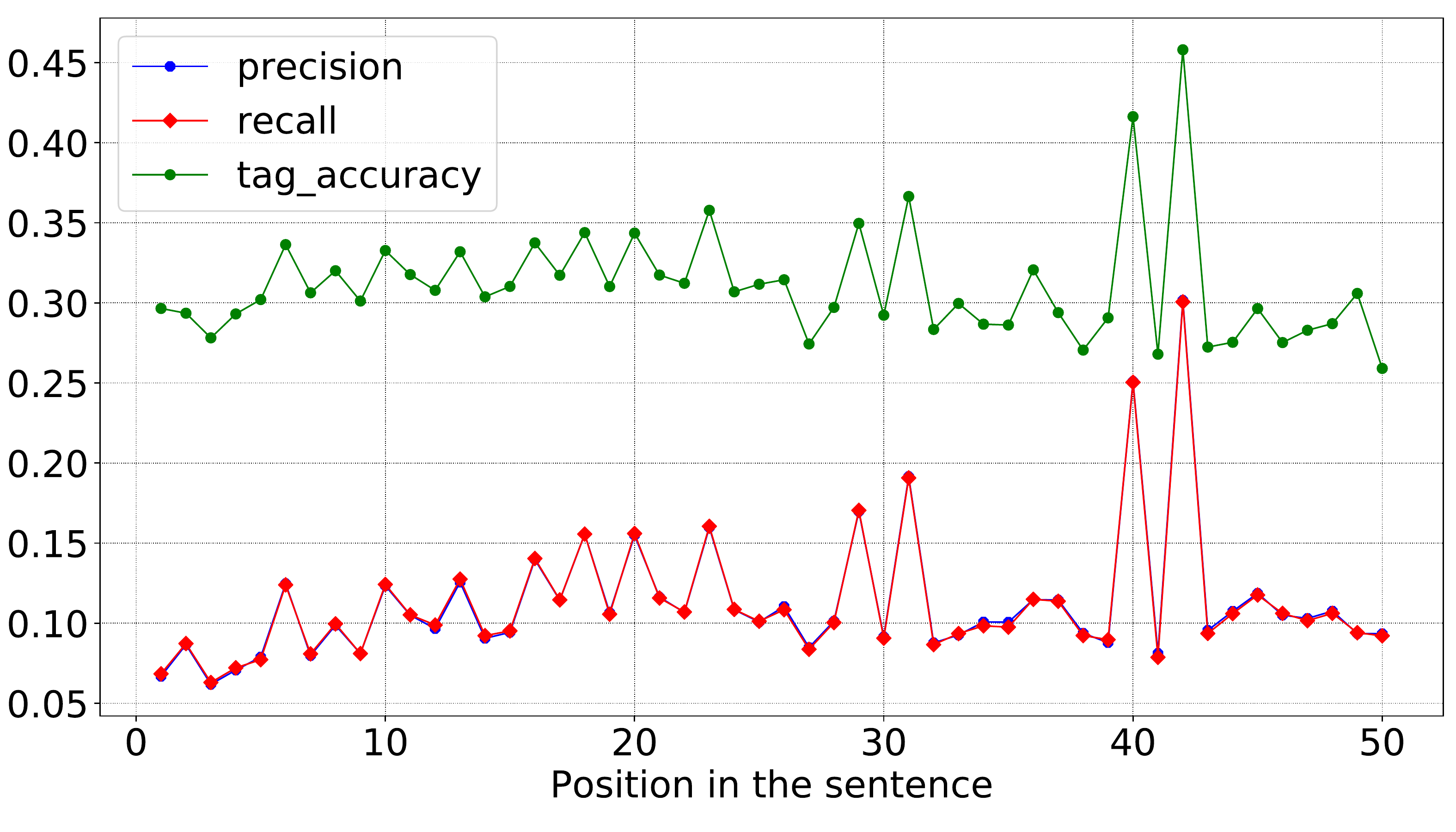}
\caption{Transformer.}
\label{fig:}
\end{subfigure}
\caption{Average syntactic similarity for each sentence position for sentences with the length up to 50 tokens for German-English.}
\label{fig:syntax_in_sentence_de_en}
\end{figure}

Table~\ref{tab:syntactic_similarity} shows the average similarity between corresponding constituent subtrees of hidden states and corresponding subtrees of their nearest neighbors, computed using PARSEVAL \citep{evalb}. Interestingly, the recurrent model has the highest average syntactic similarity. This confirms our hypothesis that the recurrent model dedicates more of the capacity of its hidden states, compared to transformer, to capturing syntactic structures. It is also in agreement with the results reported on implicit learning of syntactic structures using extrinsic tasks \citep{D18_Recurrent}.  We should add that our approach may not fully explain the degree to which syntax in general is captured by each model, but only to the extent to which this is measurable by comparing syntactic structures using PARSEVAL.

Figures~\ref{fig:syntax_in_sentence_en_de} and \ref{fig:syntax_in_sentence_de_en} show the average syntactic similarity scores per position in the sentences with length up to 50 for both German-English and English-German and for both recurrent and transformer models. We performed this analysis to check whether there is any trending increase or decrease similar to what we have observed for the mean coverage scores, see Section~\ref{subsec:WordNet_covrage}. 

Interestingly, there are slightly increasing trends in the precision and the recall scores for both the transformer and the recurrent model in both translation directions. However, the figures show that the change is more pronounced for the recurrent model. The recurrent model not only generally achieves higher syntactic similarity scores compared to the transformer model, but the score difference also increases by observing more context.

Another interesting observation in Figures~\ref{fig:syntax_in_sentence_en_de} and \ref{fig:syntax_in_sentence_de_en} is the similarity of the general patterns of changes in the diagrams for the recurrent and transformer model. This may mean that the difficulty of syntactic structures throughout the sentences is almost the same for both models. We attribute this difficulty to the depth of a word in the syntactic tree structure. However, future studies are required to confirm this suspicion.

Although the syntactic similarity scores are higher for the recurrent model, the changing pattern of the average similarity scores throughout sentences is nearly identical for both models.

%
%
%
\begin{figure}[tbh!]
\centering
\begin{subfigure}{\linewidth}
\centering
\includegraphics[scale=0.30]{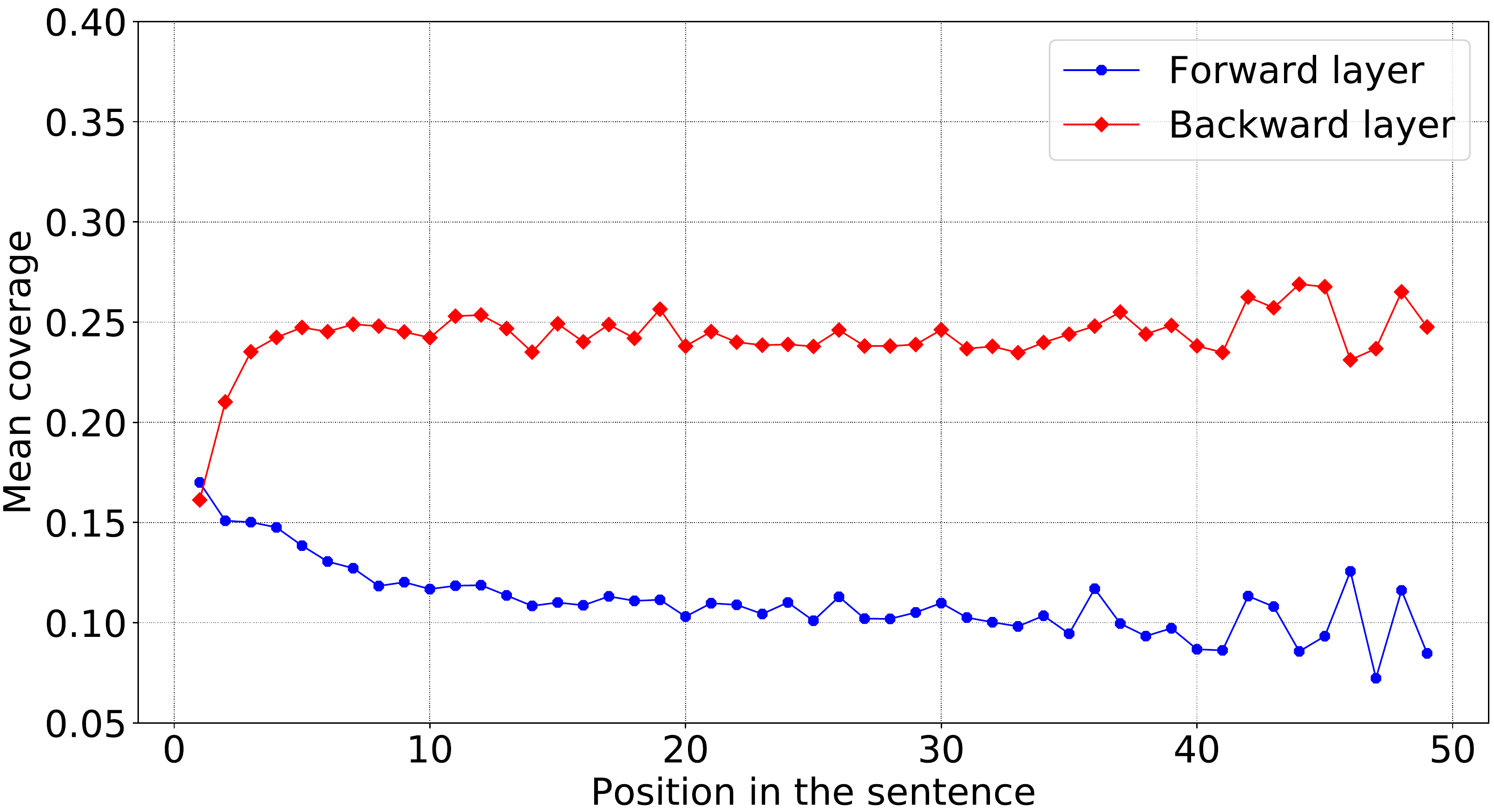}
\caption{Covered by the nearest neighbors of the embedding of the corresponding word of the hidden state.}
\label{fig:coverage_a}
\end{subfigure}

\begin{subfigure}{\linewidth}
\centering
\includegraphics[scale=0.30]{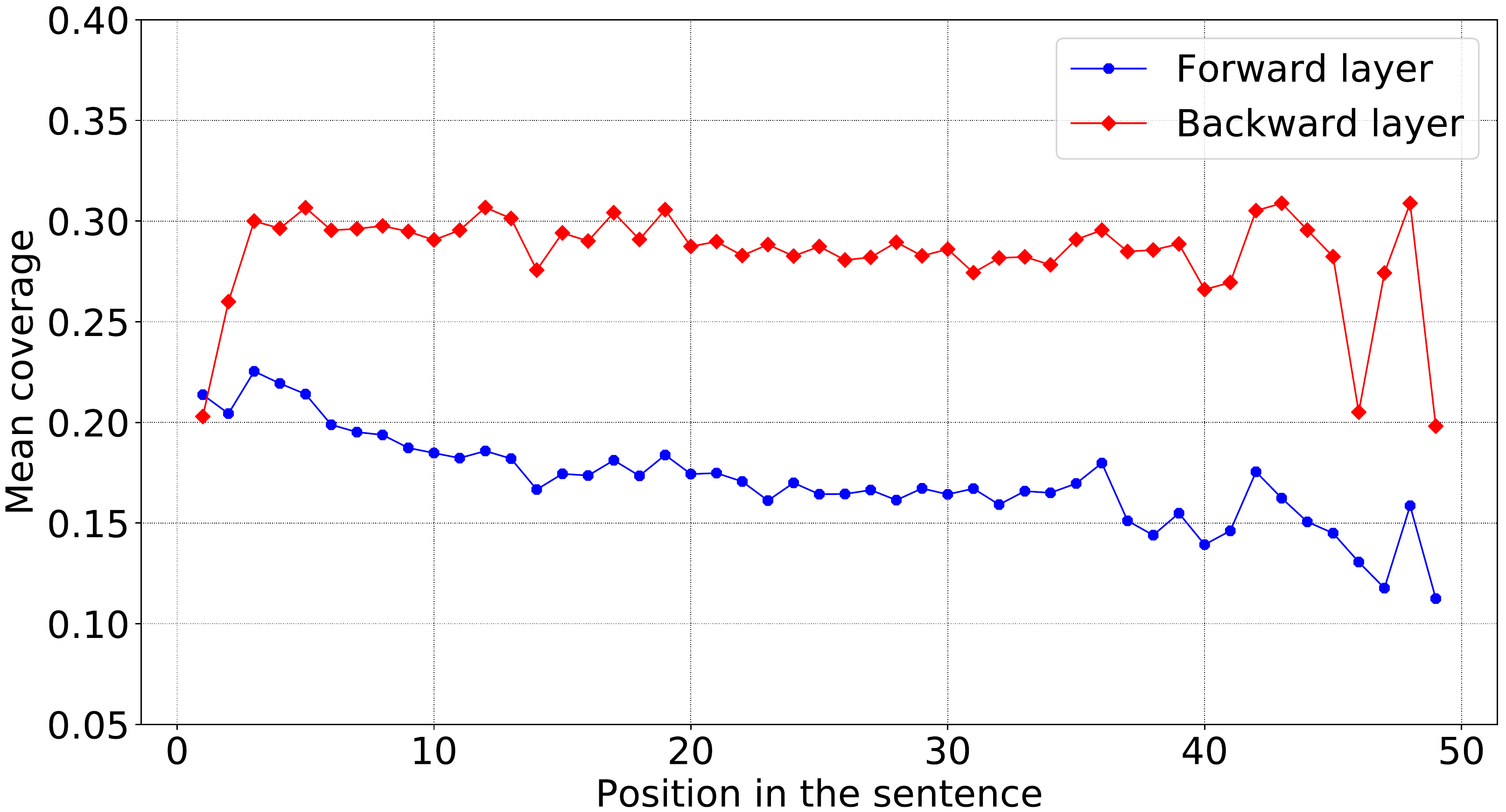}
\caption{Covered by the directly related words of the corresponding word of the hidden state in WordNet.}
\label{fig:coverage_b}
\end{subfigure}
\caption{The mean coverage per position of the nearest neighbors of hidden states from the forward and backward recurrent layers.}
\label{fig:positionCoverage_en_layer12}
\end{figure}

%
%
%
%
\subsection{Direction-Wise Analyses}
\label{subsec:directio_wise}

To gain a better understanding of the importance of directionality on the hidden states in the recurrent model, we repeat our experiments with hidden states from different directions. Note that so far the recurrent hidden states in our experiments were the concatenation of the hidden states from both directions of our encoder. In Section~\ref{section:dataset_models}, Figure~\ref{fig:rec_hidden_states} shows both direction-wise hidden states and their concatenations. So far we have used the hidden states $H_1, H_2, H_3, ..., H_T$ from both the recurrent and the transformer model. In the discussion below, we use the hidden states $\overrightarrow{h}_1, \overleftarrow{h}_1, \overrightarrow{h}_2, \overleftarrow{h}_2, \overrightarrow{h}_3, \overleftarrow{h}_3, ..., \overrightarrow{h}_T, \overleftarrow{h}_T$ from the recurrent model.

\begin{table}[tbh]
\centering
\small
\caption{Percentage of the nearest neighbors of hidden states, from the forward and backward layers, that are covered by the list of the nearest neighbors of embeddings and the list of the directly related words in WordNet.}
\label{table:EmbedCoverageLayers}
\begin{tabular}{|l|c|c|c|c|}
\hline
 \multirow{2}{*}{POS}& \multicolumn{2}{c|}{Embedding} & \multicolumn{2}{c|}{WordNet}\\
 \cline{2-5}
& Forward & Backward & Forward & Backward\\
\hline
All & 12\% & 24\% & 18\% & 29\% \\
VERB & 19\% & 36\% & 38\% & 52\%\\
NOUN & 9\% & 21\% & 14\% & 25\%\\
ADJ & 13\% & 22\%& 12\% & 17\%\\
ADV & 28\% & 34\%& 20\%& 23\%\\
\hline
\end{tabular}
\end{table}

Table~\ref{table:EmbedCoverageLayers} shows the statistics of embedding coverage and WordNet coverage from the forward and the backward layers. As shown, the coverage of the nearest neighbors of the hidden states from the backward recurrent layer is higher than the nearest neighbors based on those from the forward layer. 

Furthermore, Figure~\ref{fig:positionCoverage_en_layer12} shows the mean coverage per position of the nearest neighbors of hidden states from the forward and the backward recurrent layers. Figure~\ref{fig:coverage_a} shows to what extent the nearest neighbors of a hidden state are on average covered by the nearest neighbors of its corresponding word embedding, see Equation~\ref{eq:embeddingCov} in Section~\ref{sec:hiddenStatesVsEmbeddings}. As shown for the forward layer the coverage degrades towards the end of sentences. However, the coverage for the backward layer, except for the very beginning, almost stays constant. The coverage for the backward layer is much higher than the coverage for the forward layer indicating that it keeps more information from the embeddings compared to the forward layer. The decrease in the forward layer indicates that it captures more context information over time and ``forgets" more of the corresponding embeddings. 

Figure~\ref{fig:coverage_b} shows the extent to which the direct WordNet relations of a word are on average covered by the nearest neighbors of its corresponding hidden state, see Equation~\ref{eq:wordnetCov} in Section~\ref{subsec:WordNet_covrage}. The difference between the coverage of the nearest neighbors of hidden states from the backward layer compared to those from the forward layer confirms that the lexical semantics of words is better captured in the backward layer. This may be indicative of a division of responsibilities between the forward and the backward layer. To this end, we investigate syntactic information in the direction-wise hidden states. 

\subsubsection{Direction-Wise Syntactic Similarity}

In addition to investigating the hidden states from the backward and forward layers by computing and comparing the embedding and WordNet coverage, we also examine the syntactic similarities.

\begin{table*}[tbh]
\centering
\small
\caption{Average parse tree similarity (PARSEVAL scores) between word occurrences and their nearest neighbors for forward and backward layers.}
\label{tab:syntactic_similarity_layers}
\begin{tabular}{|l|c|c|}
\multicolumn{3}{c}{English-German}\\
\hline
Model & Forward Layer & Backward Layer \\
\hline
Precision & 0.39  &  0.30\\
Recall & 0.39 &  0.30\\
Matched Brackets & 0.43 & 0.35\\
Cross Brackets & 0.29 & 0.30\\
Tag Accuracy & 0.49 & 0.37\\
\hline
\end{tabular}
\end{table*}

\begin{figure}[tbh!]
\centering
\begin{subfigure}{\linewidth}
\centering
\includegraphics[scale=0.3]{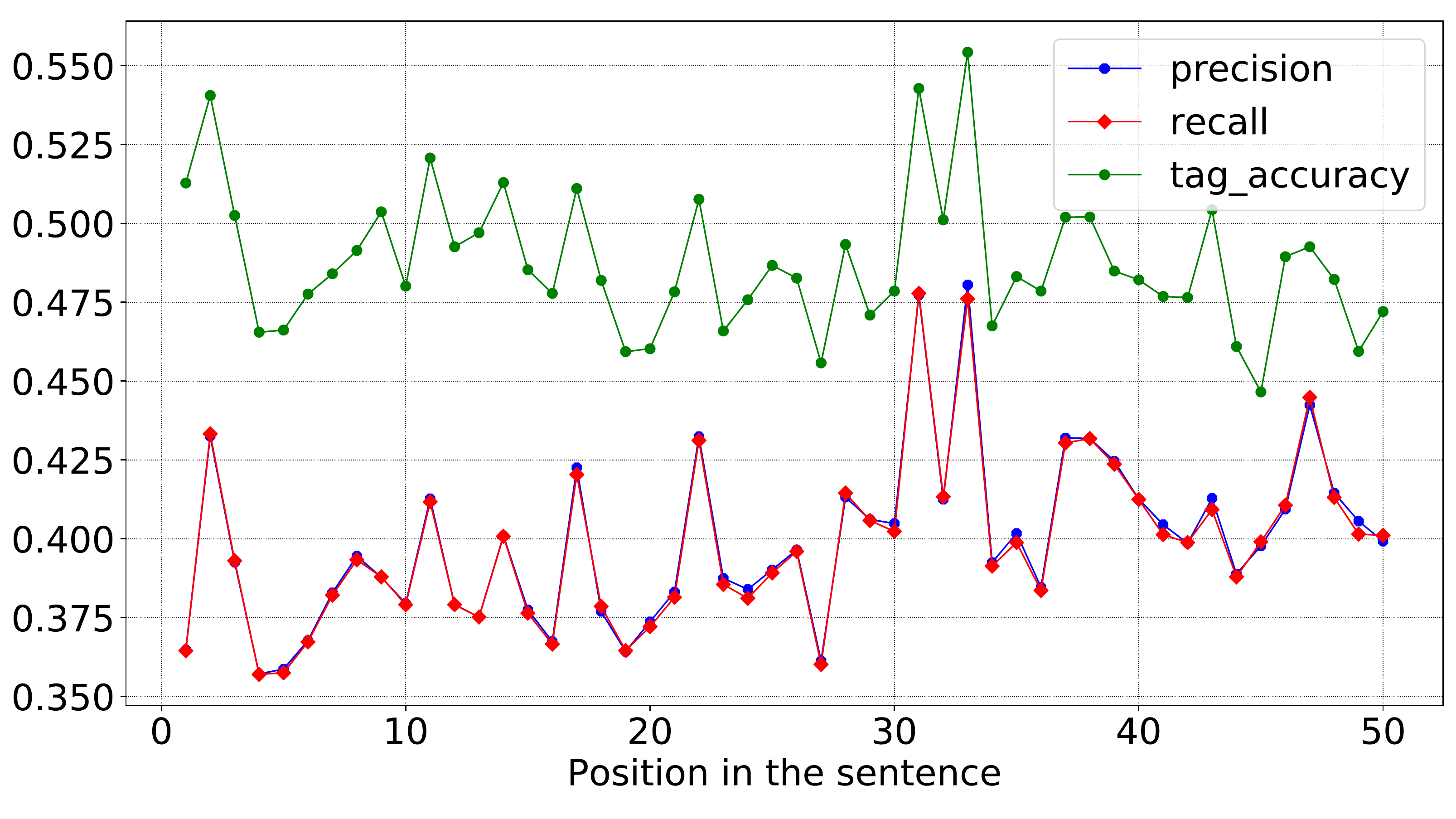}
\caption{Forward layer.}
\label{fig:syntax_in_sentence_layer1}
\end{subfigure}

\begin{subfigure}{\linewidth}
\centering
\includegraphics[scale=0.3]{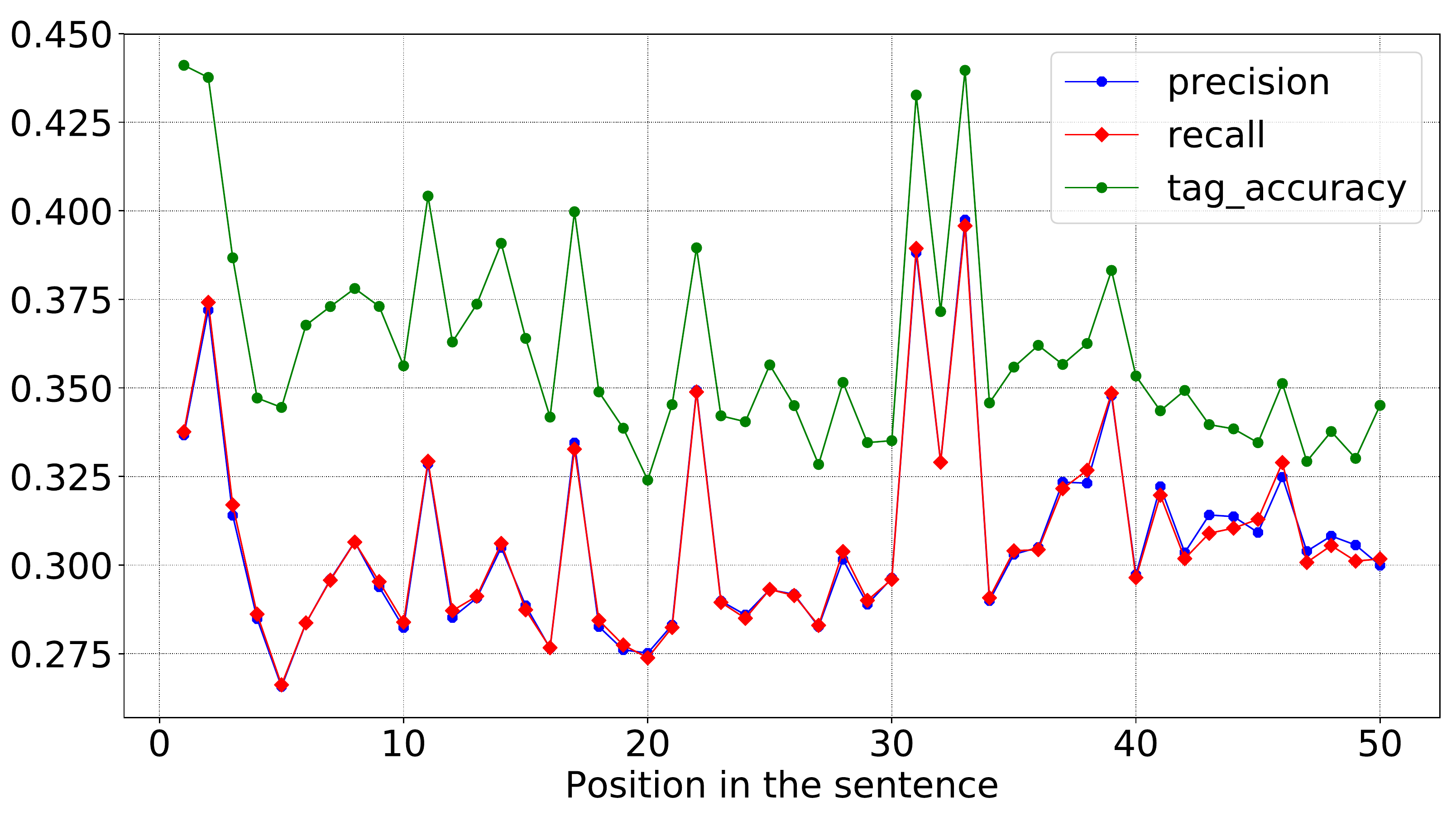}
\caption{Backward layer.}
\label{fig:syntax_in_sentence_layer2}
\end{subfigure}
\caption{Average syntactic similarity for each sentence position throughout sentences with lengths of up to 50 tokens for the forward and the backward layers from the recurrent model.}
\label{fig:syntax_in_sentence_layers}
\end{figure}

Table~\ref{tab:syntactic_similarity_layers} reports the similarity scores for the hidden states for both directions. Interestingly, the syntactic similarity for the hidden states from the forward layer is higher than the similarity for the hidden states from the backward layer. This difference is evidence for our hypothesis that the forward layer uses more of its capacity to capture context information compared to the backward layer. This also explain the drop in the embedding and WordNet coverages in the forward layer as the hidden states from this layer devote more capacity to context information.

Figures~\ref{fig:syntax_in_sentence_layer1} and \ref{fig:syntax_in_sentence_layer2} show the average syntactic similarity scores (precision, recall and tag accuracy) per position for sentences with length of up to 50 for the forward and the backward layers, respectively. Both figures show similarly increasing trends over time. However, the scores for the forward layer are generally higher than for the backward layer, see also Table~\ref{tab:syntactic_similarity_layers}. This means that the forward layer learns more syntactic information compared to the backward layer. This confirms that there is a devision of responsibilities between both layers.

\begin{figure}[tbh!]
\centering
\begin{subfigure}{\linewidth}
\centering
\includegraphics[scale=0.30]{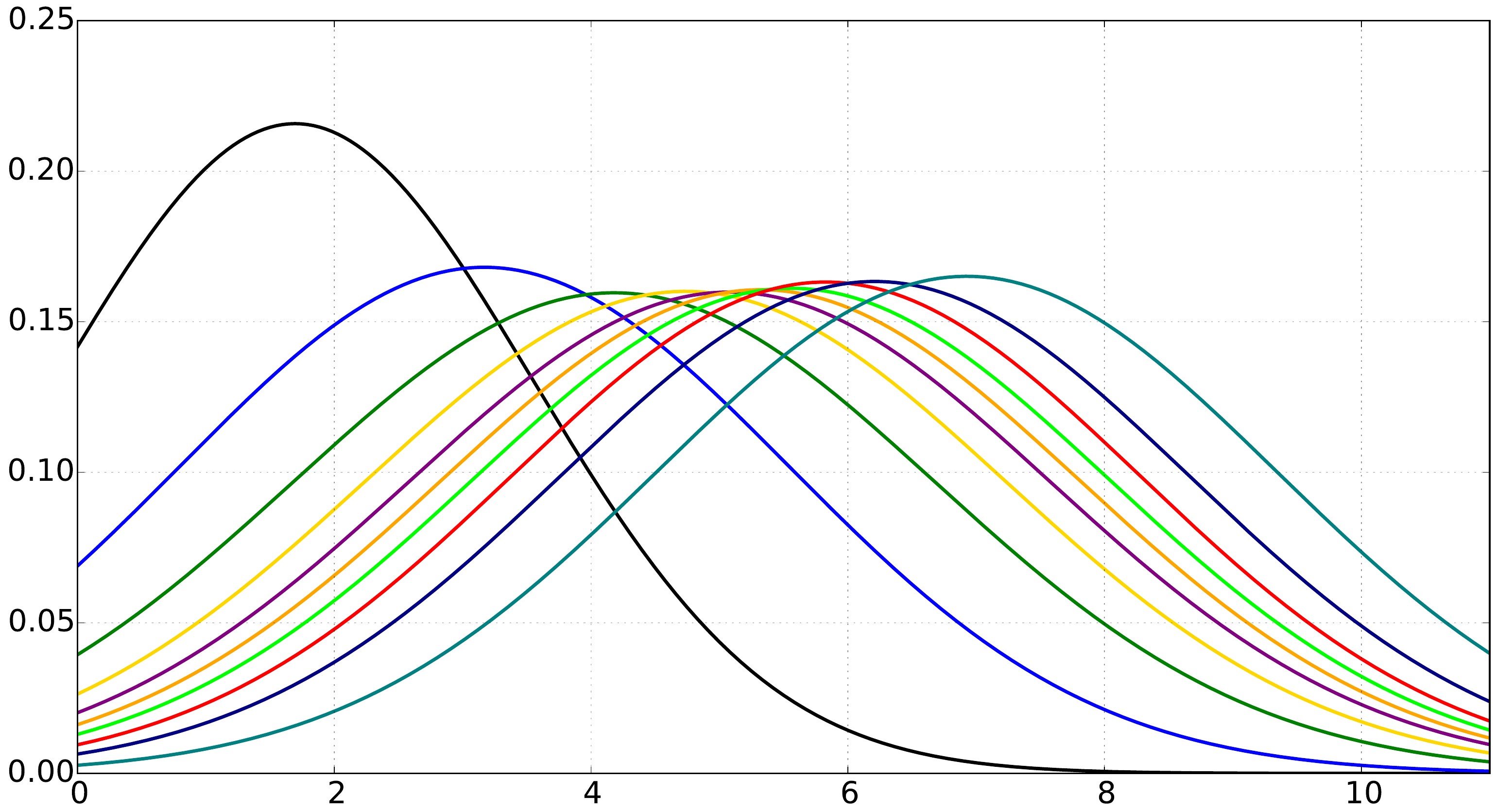}
\caption{Forward layer}
\label{fig:distFromLayer1}
\end{subfigure}

\begin{subfigure}{\linewidth}
\centering
\includegraphics[scale=0.30]{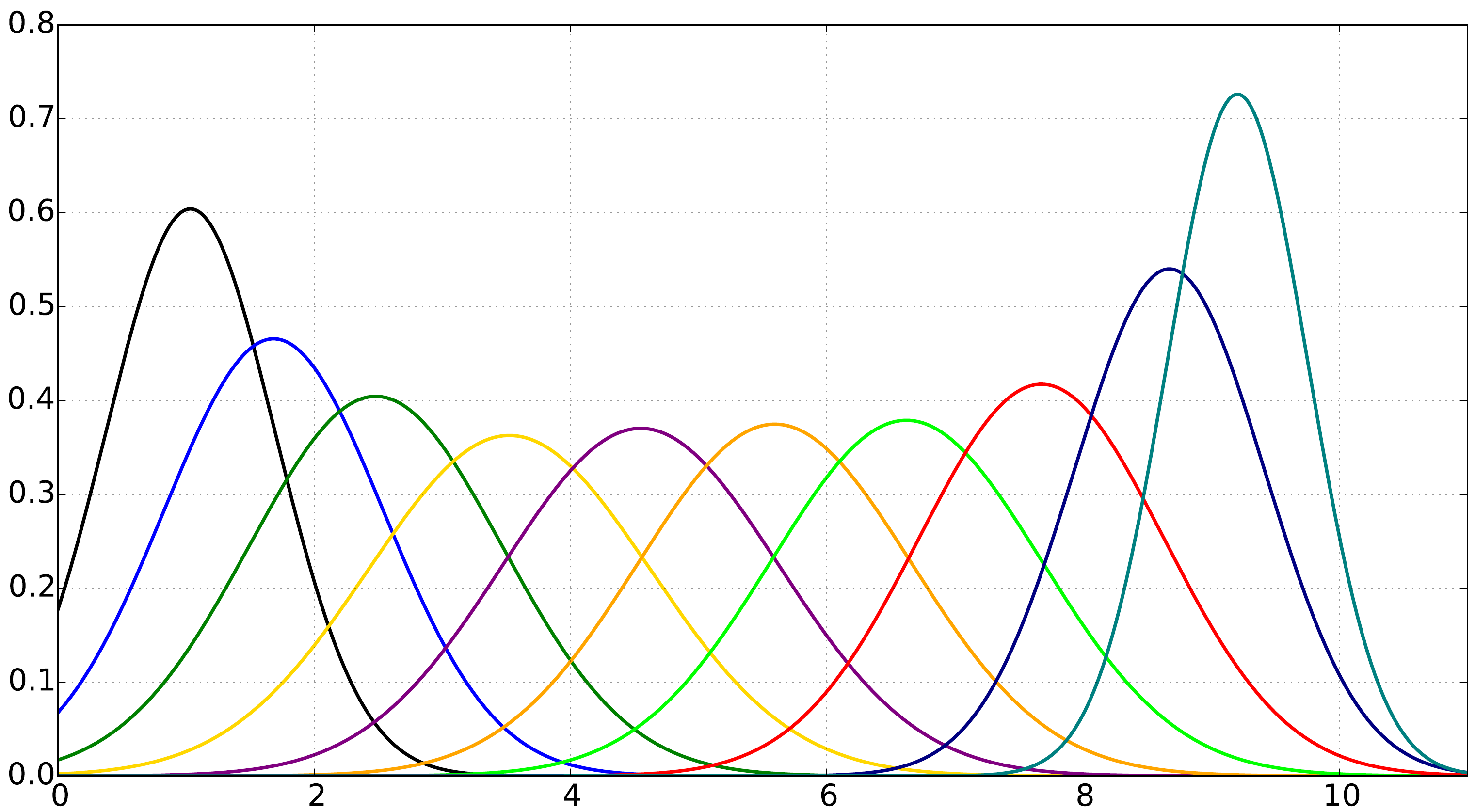}
\caption{Backward layer}
\label{fig:distFromLayer2}
\end{subfigure}

\caption{Estimated distributions of the relative positions of the nearest neighbors of hidden states, from the different layers, per bin of relative position.}
\label{fig:positionDistribution_en_layer12}
\end{figure}

\subsubsection{Direction-Wise Positional Bias}

Figure~\ref{fig:positionDistribution_en_layer12} visualizes the positional bias for the hidden states for the forward and backward directions. Interestingly, there is almost no positional bias for the hidden states in the forward layer, which entails positional biases are mostly coming from the backward layer hidden states. Putting the other observations from the direction-wise analysis together with these observations, it is clear that the positional bias is related to learning lexical semantics as it is observed in the layer that captures more lexical semantics. The forward layer, which learns more syntax, has almost no positional bias. The forward layer also has a higher variance, which means it takes more positions into account to capture wider contexts required for encoding syntactic structure. The positional bias distributions also support our hypothesis that there is a division of responsibilities between the forward and backward layer.

%
%
\section{Conclusion}

In this chapter, we introduced an intrinsic way of investigating the information encoded in the hidden states of neural machine translation models by examining the nearest neighbors of the encoder hidden states. We conducted our study by answering the following sub-research questions:
 
\begin{itemize}[wide, labelwidth=!, labelindent=0pt ]
\item[] \textbf{\ref{rq:hiddenStateSub1}} \textit{\acl{rq:hiddenStateSub1}}
\end{itemize}

\noindent Although some of the information carried by the word embeddings is transferred to the corresponding hidden states, we show that the hidden state representations capture information beyond the corresponding embeddings. We show that the hidden states capture more of the WordNet relations of the corresponding word than they capture from the nearest neighbors of the embeddings.

\begin{itemize}[wide, labelwidth=!, labelindent=0pt ]
\item[] \textbf{\ref{rq:hiddenStateSub2}} \textit{\acl{rq:hiddenStateSub2}}
\end{itemize}

\noindent We studied the lists of the nearest neighbors of the hidden states from various perspectives. We examined to what extent the capacity of the hidden states is devoted to capturing different types of information, including syntactic and lexical semantic information. By computing the shared portion of the nearest neighbors between the hidden states and the corresponding embeddings, we have an estimate of the capacity of the hidden states used to keep useful embedding information. Besides, our WordNet coverage score is an estimate of the extent to which the hidden states capture context-dependant lexical semantics. It is important to note that our definition of embedding coverage and WordNet coverage are not mutually exclusive. It means that there may be neighbors that are counted in computing both scores. However, WordNet coverage always trumps embedding coverage, which shows that the hidden states encode lexical information beyond what is captured by word embeddings. Additionally, the comparison of the scores from the two architectures provides additional insights into the extent to which each of the architectures encodes lexical semantics.

As mentioned before, to examine the syntactic information encoded by the hidden states, we compute the similarity of the corresponding local parse trees of the words of interest versus the parse trees of their nearest neighbors. The computed similarity scores give an estimate of the extent to which each of the studied architectures is capable of capturing syntax.  

\begin{itemize}[wide, labelwidth=!, labelindent=0pt ]
\item[] \textbf{\ref{rq:hiddenStateSub3}} \textit{\acl{rq:hiddenStateSub3}}
\end{itemize}

\noindent Using the proposed intrinsic methods, we compare the recurrent and transformer architectures in terms of capturing syntax and lexical semantics. For a detailed comparison we use the metrics of our intrinsic study which can be interpreted as an estimation of the amount of syntactic and lexical semantics information captured by the hidden states. We report embedding coverage, WordNet coverage, syntactic similarity, positional bias and concentration of nearest neighbors for the transformer and recurrent architectures and compare the models using these scores. We also study the changing trends of some of these measures throughout sentences to compare the behavior of the hidden states of the two architectures across time. We show that the transformer model is superior in terms of capturing lexical semantics, while the recurrent model better captures syntactic similarity.

Additionally, we investigated various types of linguistic information captured by the different directions of a bidirectional recurrent model.
We show that the backward recurrent layer captures more lexical and semantic information, whereas the forward recurrent layer captures more long-distance, contextual information. 
We also show that the forward recurrent layer has less positional bias compared to the backward recurrent layer.

Finally, our analyses indicate that the positional bias is more prevalent as it gets closer to the borders of sentences and that it dominates lexical and semantic similarities at these regions. This leads to nearest neighbors that are irrelevant in terms of lexical and semantic similarity and only share positional similarities with the word of interest.

Considering the answers for our sub-research questions, we can now go back to our main research question.

\begin{itemize}[wide, labelwidth=!, labelindent=0pt ]
\item[] \textbf{\ref{rq:hiddenStateMain}} \textit{\acl{rq:hiddenStateMain}}
\end{itemize}

\noindent By measuring the similarities of the nearest neighbors of the hidden states, we show that only part of the capacity of the hidden states is devoted to transferring the information already carried by the corresponding word embeddings. We also show that the hidden states capture WordNet-aligned lexical semantics beyond what is captured by the word embeddings. Furthermore, we demonstrate that the hidden states use some of their capacity to encode syntactic information and some positional biases, depending on the architecture of the network.
Our findings contribute to the interpretability of neural machine translation by showing what information is captured by the hidden states and to what extent this differs by architecture.

\bookmarksetup{startatroot} 
\addtocontents{toc}{\bigskip} 


\chapter{Conclusions}
\label{chapter:conclusions}

In this thesis we have investigated how phrase-based and neural machine translation capture different syntactic and semantic phenomena. Our goal was to contribute to the interpretability of both phrase-based and neural machine translation by studying the role of different words in phrase reordering in phrase-based machine translation and by analyzing the behavior of the attention model and the behavior of the hidden state representations in neural machine translation. 

Machine translation systems are among the most complex NLP systems. In phrase-based machine translation systems, phrase reordering is one of the most difficult tasks~\citep{bisazza2015survey, zhang-etal-2007-phrase} and as a result the models addressing this problem have often been the most complex components of the systems.
Deep neural machine translation models showed significant improvements in handling reordering~\citep{bentivogli-etal-2016-neural}, but at the cost of interpretability. The motivation for the work in this thesis has been to provide a deeper understanding of how phrase-based and neural machine translation systems capture different syntactic phenomena. To achieve this goal, we have focused on phrase reordering in phrase-based machine translation and the attention model plus the encoder hidden representations in neural machine translation.

In this thesis, we have addressed the following three aspects. First, although there have been many feature engineering attempts to propose stronger reordering models in phrase-based machine translation~\citep{Tillmann:2004:UOM:1613984.1614010,koehn05iwslt,Galley:2008:SEH:1613715.1613824,cherry:2013:NAACL-HLT,nagata2006clustered,durrani2014investigating}, none have looked into the phrase-internal similarities as proposed in this thesis. Second, the attention model in neural machine translation is often considered to behave like the traditional alignment model~\citep{liu-EtAl:2016:COLING,cohn-EtAl:2016:N16-1,chen2016guided,alkhouli-EtAl:2016:WMT} and is trained to follow alignment models from phrase-based machine translation without fully understanding the similarities and differences between the two models. Third, there are no intrinsic studies of what syntactic and semantic information is captured by the hidden state representations in neural machine translation and how much this differs between architectures.

In Chapter~\ref{chapter:research-01}, we have investigated the influence of the words in a phrase on the phrase reordering behavior in a phrase-based machine translation model. We have investigated different phrase generalization strategies to see which strategy sufficiently models the pattern by which the influence of the internal words on the reordering prediction of a phrase changes.

In Chapter~\ref{chapter:research-02}, we have studied to what extent words in the attention model in neural machine translation exhibit different attention distributions in different syntactic contexts. Additionally, we have compared an attention model to a traditional alignment model to show their similarities and differences. We have also shown under which conditions to train the attention model with explicit signals from a traditional alignment model.

In Chapter~\ref{chapter:research-03}, we have expanded our analysis to the hidden states of the encoder of neural machine translation models. We have investigated the information captured by the hidden states and have compared two prominent neural machine translation architectures based on the information they capture. Our investigation shows to what extent different syntactic and lexical semantic information is captured by the encoder hidden states. Additionally, our comparison shows which architecture is better in capturing structural syntactic information and which one is better in capturing lexical semantics. 

\section{Main Findings}

In this section, we revisit the research questions and summarize our findings regarding each research question. We start with the research questions on phrase reordering behavior in a phrase-based machine translation model:

\begin{itemize}[wide, labelwidth=!, labelindent=0pt ]
\item[]   \textbf{\ref{rq:reorderingMain}} \textit{\acl{rq:reorderingMain}}
\end{itemize}

\noindent To answer this question, we have introduced in Chapter~\ref{chapter:research-01} a novel method that builds on the established idea of backing off to shorter histories, commonly used in language model smoothing~\citep{chen1999empirical}. We have experimented with different backing-off strategies and shown that they can be successfully applied to smooth lexicalized and hierarchical reordering models in statistical machine translation. 
%
%
To investigate the influence of the internal words on reordering distributions, we experimented with different generalized forms of the phrase-pairs. We have kept the exposed heads in their original forms and have generalized the remaining words by using word classes or simply removing them. The experimental findings show that sub-phrase-pairs consisting of just the exposed heads of a phrase-pair tend to be the most important ones and most other words inside a phrase-pair have negligible influence on reordering behavior. 

Our results in Section~\ref{sec:exp_research_1} show that the full lexical form of phrase-pairs does not play a significant role in estimating reliable reordering distributions. However, the improvements achieved by keeping important words in their surface form show that the lexical forms of the exposed heads are important for estimating the reordering distributions more accurately.

\begin{itemize}[wide, labelwidth=!, labelindent=0pt ]
\item[]  \textbf{\ref{rq:reorderingSub1}} \textit{\acl{rq:reorderingSub1}}
\end{itemize}


\noindent Contradicting earlier approaches, which assume the last and the first words of a phrase-pair to be the most influential words for defining reordering behavior \citep{nagata2006clustered,cherry:2013:NAACL-HLT},  our experiments in Section~\ref{subsec:result_ch1} show that the exposed heads of phrase-pairs are stronger predictors of phrase reordering behavior. 
Here, we have experimented with both backing off towards the border words by assuming those as the important words, following \citet{nagata2006clustered} and \citet{cherry:2013:NAACL-HLT}, and generalizing towards exposed heads, assuming those as the most important words, following \citet{chelba2000structured} and~\citet{garmash2015bilingual}. We have shown that generalized representations of phrase-pairs based on exposed heads can help decrease sparsity and result in more reliable reordering distributions.


\begin{itemize}[wide, labelwidth=!, labelindent=0pt ]
\item[]  \textbf{\ref{rq:reorderingSub2}} \textit{\acl{rq:reorderingSub2}}
\end{itemize}

\noindent The results in Section~\ref{subsec:result_ch1} show that backing off is helpful. However, our back-off models are not our best models even though they perform better than the baselines. We have found that identifying the most influential internal words on the reordering is more important than how to back off towards them to estimate better general distribution. 



\begin{itemize}[wide, labelwidth=!, labelindent=0pt ]
\item[]  \textbf{\ref{rq:reorderingSub3}} \textit{\acl{rq:reorderingSub3}}
\end{itemize}

\noindent Our generalized representations of phrase-pairs based on exposed heads have achieved the best results in our experiments. As mentioned above, this is mostly due to the better choice of the most influential internal words rather than the approach we took for the generalization of less important words. 
Interestingly, simply removing the less influential words yields the best performance improvements in our experiments, see Section~\ref{sec:exp_research_1}. 

Note that we cannot simply conclude that the exposed heads or the border words are always the important words that define the reordering behavior of phrases. However, these words arguably provide good signals on the reordering behavior of the phrase-pairs and exposed heads are more influential compared to border words on average.

Considering the analysis of the length of infrequent phrase-pairs used during translation in Section~\ref{sec:analysis}, we also conclude that a smoothing model that would be able to further improve the reordering distributions of single-word phrase-pairs is crucial for achieving bigger improvements.
 

In Chapter~\ref{chapter:research-02}, we have continued our interpretability analysis of how machine translation systems capture different syntactic phenomena by studying attention in neural machine translation. We have compared traditional alignments, which are an essential component in phrase-based machine translation, with attention in neural machine translation. 


For this part of our research, we asked the following questions:

\begin{itemize}[wide, labelwidth=!, labelindent=0pt ]
\item[] \textbf{\ref{rq:attnMain}} \textit{\acl{rq:attnMain}}
\end{itemize}

\noindent Attention models in neural machine translation are often considered similar to the traditional alignment model of phrase-based systems~\citep{alkhouli-EtAl:2016:WMT, cohn-EtAl:2016:N16-1, liu-EtAl:2016:COLING, chen2016guided}. In that sense, an attention model should pay attention to the translational equivalent of the generated target word as defined by traditional alignments. To verify this, our first step has been to compare the attention model with traditional alignments and ask the following question: 


\begin{itemize}[wide, labelwidth=!, labelindent=0pt ]
\item[] \textbf{\ref{rq:attnSub1}} \textit{\acl{rq:attnSub1}}
\end{itemize}

 
\noindent We have shown in Section~\ref{sec:emp_analysis_ch2} that attention agrees with traditional alignments to a certain extent. However, this differs substantially by attention mechanism and the type of the word being generated. When generating nouns, attention mostly focuses on the translational equivalent of the target word on the source side and behaves indeed very similar to traditional alignments. However, its behavior diverges from that of the traditional alignment model when it comes to generating words that need more contextual information to be generated correctly. This is especially the case for verbs that need to be in agreement with their subject or are modified by an auxiliary verb.


Next, we have investigated whether the divergence from traditional alignment is due to errors in the attention model or is evidence for attention capturing information beyond traditional alignments. Here, we have asked the following question:

\begin{itemize}[wide, labelwidth=!, labelindent=0pt ]
\item[]  \textbf{\ref{rq:attnSub2}} \textit{\acl{rq:attnSub2}}
\end{itemize}

\noindent In Section~\ref{subse:attention_con_ch2}, we have examined the correlation of the translation quality with attention loss and attention concentration. The low correlations between these quantities, especially in the case of generating verbs, show that the difference is not a result of an error in attention, but a side effect of capturing additional information compared to traditional alignments.

The higher concentration of attention for nouns and adjectives shows that for these syntactic phenomena, attention behaves like traditional alignment and captures mostly translational equivalence. However, by distributing its weight to other relevant context words, in particular while translating verbs, attention integrates other context words that influence the generation of the target word. 


\begin{itemize}[wide, labelwidth=!, labelindent=0pt ]
\item[]  \textbf{\ref{rq:attnSub3}} \textit{\acl{rq:attnSub3}}
\end{itemize}

\noindent We have shown in Section~\ref{sec:emp_analysis_ch2} that the attention behavior changes based on the POS tag of the target word. The concentrated pattern of attention and the relatively high correlations for nouns show that training attention with explicit alignment labels is useful for generating nouns. However, this is not the case for verbs, as a large portion of attention is devoted to words other than the alignment points. In these cases, training attention with alignments will force the attention model to ignore this useful information. 

\begin{itemize}[wide, labelwidth=!, labelindent=0pt ]
\item[] \textbf{\ref{rq:attnSub4}} \textit{\acl{rq:attnSub4}}
\end{itemize}

\noindent We have shown in Section~\ref{subsec:Attn_dist_ch2} that the attention model learns to assign some attention weights to dependency relations of a word while translating it. This is especially true in the case of words that require context information in order to be translated properly. For example, verbs, adverbs, conjunctions and particles only receive half of the attention weight on average, while the rest of the attention is paid to their dependent words.

In Chapter~\ref{chapter:research-03}, we have focused on the encoder hidden states in different neural machine translation architectures by looking at their nearest neighbors. This is a natural follow up of our study of attention on the encoder side in Chapter~\ref{chapter:research-02}. We ask the following research questions:

\begin{itemize}[wide, labelwidth=!, labelindent=0pt ]
\item[] \textbf{\ref{rq:hiddenStateMain}} \textit{\acl{rq:hiddenStateMain}}
\end{itemize}

\noindent By detecting the similarities and the differences of the nearest neighbors of the hidden states, we have shown to what extent the hidden states capture syntactic and lexical semantic information. We have shown that part of the capacity of the hidden states is devoted to transferring information already carried by the corresponding word embeddings. We have also shown that the hidden states capture some WordNet-aligned lexical semantics beyond what is captured by the word embeddings. Furthermore, we have demonstrated that the hidden states use some of their capacity to encode syntactic information and some positional biases, depending on the architecture of the network.

\begin{itemize}[wide, labelwidth=!, labelindent=0pt ]
\item[] \textbf{\ref{rq:hiddenStateSub1}} \textit{\acl{rq:hiddenStateSub1}}
\end{itemize}

\noindent Although some of the information carried by the word embeddings is transferred to the corresponding hidden states, we have shown in Section~\ref{sec:emp_analysis_ch3} that hidden state representations capture quite different information from what is captured by the corresponding embeddings. We have also shown that hidden states capture more of the WordNet relations of the corresponding word than they capture from the nearest neighbors of the embeddings.

\begin{itemize}[wide, labelwidth=!, labelindent=0pt ]
\item[] \textbf{\ref{rq:hiddenStateSub2}} \textit{\acl{rq:hiddenStateSub2}}
\end{itemize}

\noindent We studied the list of the nearest neighbors of the hidden states from various perspectives. We have examined to what extent the capacity of the hidden states is devoted to capturing different types of information, including syntactic and lexical semantic information. By computing the shared portion of the nearest neighbors between the hidden states and the corresponding embeddings, we have given an estimate of the capacity of the hidden states used to keep embedding information. In addition, our WordNet coverage score is an estimate of the extent to which the hidden states capture context-dependent lexical semantics. It is important to note that our definition of embedding coverage and WordNet coverage are not mutually exclusive since there may be neighbors that are counted in computing both scores. However, WordNet coverage always trumps embedding coverage, which shows that hidden states encode lexical information beyond what is captured by word embeddings. 

To examine the syntactic information encoded by the hidden states, we have computed the similarity between the local parse trees of the words of interest and the parse trees of their neighbors (see Section~\ref{subsec:syntactic_ch3}). The computed similarity scores give an estimate of the extent to which each of the studied architectures is capable of capturing syntax.  

Our results show that attention model and hidden state representations play a complementary role in capturing contextual information. Hidden state representations are more capable of capturing local structural information, while more global contextual information is captured by the attention model. 

\begin{itemize}[wide, labelwidth=!, labelindent=0pt ]
\item[] \textbf{\ref{rq:hiddenStateSub3}} \textit{\acl{rq:hiddenStateSub3}}
\end{itemize}

\noindent Here, we have compared recurrent and transformer architectures in terms of capturing syntax and lexical semantics in Section~\ref{sec:emp_analysis_ch3}. To perform a detailed comparison we have used metrics in our intrinsic study that estimate the amount of syntactic and lexical semantic information captured by the hidden states. In Section~\ref{sec:emp_analysis_ch3}, we have reported embedding coverage, WordNet coverage, syntactic similarity, positional bias and concentration of nearest neighbors for transformer and recurrent architectures and have compared the models using these scores. We have also studied the changing trends of some of these measures throughout sentences to compare the behavior of the hidden states of the two architectures. We have shown that there are differences in the capacity that the hidden states from each architecture devote to capturing a specific type of information. We show that the transformer model is superior in terms of capturing lexical semantics, while the recurrent model better captures syntactic similarity.

Additionally, we provide a detailed analysis of the behavior of the hidden states, both direction-wise and for the concatenations. 
We have investigated various types of linguistic information captured by the different directions of hidden states in a bidirectional recurrent model.
We have shown that the reverse recurrent layer captures more lexical information, whereas the forward recurrent layer captures more long-distance, contextual information. 
One can also see that the forward recurrent layer has less positional bias compared to the backward recurrent layer.

Finally, we have provided analyses on how the behavior of the hidden states change through sentences. Our analyses indicate that positional bias is more prevalent close to the border of sentences and it dominates lexical and semantic similarities at these regions. This leads to nearest neighbors that are irrelevant in terms of lexical similarity and only share positional similarities with the word of interest.

\section{Future Work}
In this thesis, we have contributed to the interpretability of both phrase-based and neural machine translation systems. We shed light on how these system capture some syntactic and semantic phenomena and how various architectures differ in capturing those phenomena. However, current neural machine translation models are so complex that more analytical studies are required to fully interpret their behavior. Below are future directions motivated by our work.

Our work motivates more complex attention models. We have shown that an attention model learns to focus its weight for translating nouns which mostly have a single translational equivalent in the source language. We have also shown that it learns to pay attention to dependent context words when translating verbs. However, improvements achieved by explicitly training an attention model using signals from traditional alignments in the e-commerce domain show that a simple attention model by itself is not sufficient~\citep{chen2016guided}. This encourages more complex attention models, like those of~\citet{feng-etal-2016-improving} and~\citet{lin-etal-2018-learning}, that can differentiate more easily between different syntactic phenomena which require different attention distribution without explicit signals from traditional alignment models. 
  
A natural follow up to this work could be an analysis of how more complex attention models or more layers of attention could capture part of the information that is captured by the hidden state representations. This may be helpful to understand how to devote hidden state capacity to similarities in the context that are not already captured by the models. For example, our comparison of the recurrent model and transformer model in Chapter~\ref{chapter:research-03} shows that transformers are capable of capturing syntactic structure information to a large extent, similar to recurrent LSTM models. This has been made possible by the multi-headed attention models, including self-attention, in transformers. 

Finally, it is still not fully clear what information is captured in the hidden state representation of neural machine translation systems in general. \citet{shi-padhi-knight:2016:EMNLP2016} and~\citet{belinkov2017neural, I17-1001} examined the information captured by the hidden state representations by using extrinsic classification tasks. In Chapter~\ref{chapter:research-03}, we have taken an intrinsic approach to show what is captured and how. The fact that hidden state representations in different architectures capture different types of information needs to be analyzed more deeply.


\backmatter



\renewcommand{\bibsection}{\chapter{Bibliography}}
\renewcommand{\bibname}{Bibliography}
\markboth{Bibliography}{Bibliography}
\renewcommand{\bibfont}{\footnotesize}
\setlength{\bibsep}{0pt}

\bibliographystyle{abbrvnat}
\bibliography{./bib/hghader}


\chapter{Summary}

Machine translation systems are large machine learning systems that learn to translate from one natural language (for example Chinese) to another (for example English). These systems scan the input text and generate part of the translation step by step. However, due to their complexity, it is difficult to interpret the decisions they make at each time step. For example, it is not possible to explain every word choice or word order that a machine translation system uses during the generation of the translation. Therefore, it is difficult to explain why a system makes certain mistakes during translation or how it is capable of regenerating complex syntactic or semantic phenomena in the target language.

Two popular types of machine translation are phrase-based and neural machine translation systems. Both of these types of machine translation systems are composed of multiple complex models or layers. Each of these models and layers learns different linguistic aspects of the source language. However, for some of these models and layers, it is not clear which linguistic phenomena are learned or how this information is learned. For phrase-based machine translation systems, it is often clear what information is learned by each model, and the question is rather how this information is learned, especially for its phrase reordering model. For neural machine translation systems, the situation is even more complex, since for many cases it is not exactly clear what information is learned and how it is learned. 

To shed light on what linguistic phenomena are captured by machine translation systems, we analyze the behavior of important models in both phrase-based and neural machine translation systems. We consider phrase reordering models from phrase-based machine translation systems to investigate which words from inside of a phrase have the biggest impact on defining the phrase reordering behavior. Additionally, to contribute to the interpretability of neural machine translation systems we study the behavior of the attention model, which is a key component in neural machine translation systems and the closest model in functionality to phrase reordering models in phrase-based systems. The attention model together with the encoder hidden state representations form the main components to encode source side linguistic information in neural machine translation. To this end, we also analyze the information captured in the encoder hidden state representations of a neural machine translation system. We investigate the extent to which syntactic and lexical-semantic information from the source side is captured by hidden state representations of different neural machine translation architectures.

\chapter{Samenvatting}

Computervertaalsystemen zijn complexe lerende algoritmes die vertalingen van een natuurlijke taal (zoals Chinees) naar een andere natuurlijke taal (zoals Engels) kunnen maken. Deze systemen werken door een ingevoerde tekst stapsgewijs te doorlopen en daarbij telkens een deel van de vertaling op te bouwen. Door de complexiteit in deze algoritmes is het moeilijk om te interpreteren welke beslissingen in elke stap van dit proces door het algoritme worden gemaakt. Het is bijvoorbeeld onmogelijk om elke woordkeuze of woordvolgorde in de opbouw van de vertaling te verklaren. Het is daardoor ook moeilijk om te begrijpen waarom bepaalde fouten zich in de vertaling voordoen, of in te schatten hoe goed een vertaalsysteem complexe syntactische of semantische structuren in de vertaling kan doorvoeren.

Twee populaire vormen van computervertaalsystemen zijn phrase-gebaseerde (Engels: \emph{phrase-based}) en neurale vertaalsystemen. Beide soorten vertaalsystemen zijn opgebouwd uit meerdere complexe modellen of lagen. Elk van deze modellen of lagen leert andere linguïstische aspecten van de taal van waaruit vertaald moet worden. Voor sommige modellen en lagen is het echter onduidelijk welke linguïstische aspecten worden geleerd en hoe deze informatie wordt geëxtraheerd. Voor phrase-gebaseerde vertaalsystemen is het weliswaar duidelijk welke informatie door elk model wordt geleerd, maar niet hoe deze informatie wordt geleerd, vooral voor herordeningsmodellen (Engels: \emph{reordering models}). Voor neurale vertalingssystemen is de situatie nog complexer, omdat voor veel gevallen zowel onduidelijk is welke informatie wordt geleerd als hoe deze informatie wordt geleerd.

Om inzicht te verschaffen in welke linguïstische aspecten worden opgevangen in computervertaalsystemen, analyseren we het gedrag van belangrijke modellen in zowel phrase-gebaseerde als neurale vertalingssystemen. We nemen herordeningsmodellen van phrase-gebaseerde modellen om te onderzoeken welke woorden van een zin de grootste impact op het gedrag van het herordeningsmodel hebben. Ook dragen we bij aan de interpreteerbaarheid van neurale vertaalmodellen, door de aandachtsmodellen die een sleutelrol in deze systemen vormen én het meest vergelijkbaar zijn met herordeningsmodellen in phrase-gebaseerde modellen, te bestuderen. Het aandachtsmodel vervult samen met de verborgen coderingslaag (Engels: \emph{encoder hidden state}) de sleutelrol voor het opnemen van de linguïstische informatie in de invoertaal. Om dit te begrijpen analyseren we de informatie die wordt vastgelegd in de verborgen coderingslaag van het neurale vertaalsysteem. We kijken hierbij naar de mate waarin syntactische en lexicaal-semantische informatie van de invoer-tekst wordt vastgelegd in de verborgen coderingslaag van verschillende neurale vertalings-architecturen.

\end{document}